\pdfoutput=1
\documentclass[10pt,twocolumn,letterpaper]{article}
\usepackage{paperstyle}

\usepackage{booktabs}
\usepackage{pifont}
\newcommand{\cmark}{\ding{51}}
\newcommand{\xmark}{\ding{55}}
\newcommand{\pmark}{$\triangle$}

\usepackage{booktabs}
\usepackage{colortbl}
\usepackage{pifont}
\definecolor{okgreen}{RGB}{46,139,87}
\definecolor{nored}{RGB}{190,60,60}
\definecolor{oursbg}{RGB}{232,241,250}
\newcommand{\nap}{\textcolor{gray}{--}}

\usepackage{siunitx}

\usepackage{siunitx}
\usepackage{multirow}
\definecolor{rankA}{HTML}{F4C2C2}
\definecolor{rankB}{HTML}{FADBC6}
\definecolor{rankC}{HTML}{FDF1CC}
\definecolor{refgray}{HTML}{8C8C8C}
\newcommand{\first}[1]{\cellcolor{rankA}\textbf{#1}}
\newcommand{\second}[1]{\cellcolor{rankB}#1}
\newcommand{\third}[1]{\cellcolor{rankC}#1}
\newcommand{\rankkey}{{\setlength{\fboxsep}{1.5pt}\colorbox{rankA}{\textbf{1st}}\,\colorbox{rankB}{2nd}\,\colorbox{rankC}{3rd}}}
\newcommand{\ours}{\textbf{AETHER (Ours)}}

\newcommand{\reff}[1]{\textcolor{refgray}{#1}}
\newcommand{\grp}[2]{\multirow{#1}{*}{\rotatebox[origin=c]{90}{\scriptsize #2}}}

\usepackage{algorithm}
\usepackage{algpseudocode}
\algrenewcommand\algorithmicrequire{\textbf{Input:}}
\algrenewcommand\algorithmicensure{\textbf{Output:}}
\algrenewcommand\alglinenumber[1]{\scriptsize #1:}
\newcommand{\seqlabel}[2][0.0536\textwidth]{\raisebox{#1}{\rotatebox[origin=c]{90}{\scriptsize #2}}}
\algnewcommand\algorithmicswitch{\textbf{switch}}
\algnewcommand\algorithmiccase{\textbf{case}}
\algdef{SE}[SWITCH]{Switch}{EndSwitch}[1]{\algorithmicswitch\ #1}{\algorithmicend\ \algorithmicswitch}
\algdef{SE}[CASE]{Case}{EndCase}[1]{\algorithmiccase\ #1:}{\algorithmicend\ \algorithmiccase}
\algtext*{EndSwitch}\algtext*{EndCase}\algtext*{EndIf}\algtext*{EndFor}\algtext*{EndWhile}\algtext*{EndFunction}

\graphicspath{{figures/}}
\definecolor{linkblue}{rgb}{0.21,0.49,0.74}
\usepackage[pagebackref,breaklinks,colorlinks,allcolors=linkblue]{hyperref}
\title{AETHER: An Autonomous-driving Evaluation Testbed with High-fidelity Environment Reconstruction}
\author{Siying Li \quad Yujia Zhao \quad Yuanbiao Cao \quad An Xie \quad Yuhang Chen\\
Jie Sun \quad Jian Sun \quad Ying Ni$^{*}$ \quad Haotian Shi$^{*}$\\[0.4em]
College of Transportation, Tongji University}
\begin{document}
\maketitle
{\renewcommand{\thefootnote}{*}\footnotetext{Corresponding authors.}}
\begin{abstract}
Closed-loop evaluation of end-to-end driving requires sensor observations that remain faithful once the policy departs from the recorded trajectory, and road users that respond to the ego rather than replay the log.
We present AETHER, a testbed that turns real driving logs into interactive Gaussian environments for camera--LiDAR closed-loop evaluation of off-the-shelf end-to-end planners.
AETHER renders both sensors from shared Gaussian geometry, restores novel views with a learned restorer, and flags each step by whether perception fidelity is preserved at the current deviation from the log.
Vehicles move on a route-anchored ground with consistent attitude, while pedestrians and riders are driven by an online model that couples crossing intent, root motion, full-body pose, and ground contact.
Across nuScenes and PandaSet with four representative planners, AETHER sustains high-fidelity camera and LiDAR observations under policy-induced viewpoint shifts, outperforming existing view restoration and matching dedicated LiDAR reconstruction at a fraction of its training cost, while its pedestrians and riders are behaviorally realistic and physically consistent.
Our closed-loop study further reveals that replaying logged traffic inflates collision rates while under-counting collisions caused by the ego itself.
Code is available at \url{https://github.com/siyingli99/Aether}.
\end{abstract}
\section{Introduction}
\label{sec:introduction}

\begin{table}[t]
\centering
\caption{\textbf{Comparison of closed-loop driving simulators.}
RT: real-time; Rest.: off-log view restoration; Flag: fidelity flag; Traffic: interactive traffic flow; Gnd.: grounded vehicles; VRU: interactive pedestrians and riders.
\cmark~yes, \pmark~partial, \xmark~no, \nap~n/a.}
\label{tab:sim_compare}
\footnotesize
\setlength{\tabcolsep}{1.75pt}
\renewcommand{\arraystretch}{1.02}
\begin{tabular}{@{}l cccc cc ccc@{}}
\toprule
& \multicolumn{4}{c}{Sensing} & \multicolumn{2}{c}{Off-log} & \multicolumn{3}{c}{Road users} \\
\cmidrule(lr){2-5}\cmidrule(lr){6-7}\cmidrule(l){8-10}
Simulator & Real & RT & Cam & LiDAR & Rest. & Flag & Traffic & Gnd. & VRU \\
\midrule
CARLA~\cite{dosovitskiy2017carla}         & \xmark & \cmark & \cmark & \cmark & \nap   & \xmark & \cmark & \cmark & \pmark \\
DriveArena~\cite{yang2025drivearena}      & \cmark & \xmark & \cmark & \xmark & \nap   & \xmark & \cmark & \nap   & \xmark \\
B2D-R~\cite{you2024bench2driver}          & \cmark & \xmark & \cmark & \xmark & \nap   & \xmark & \cmark & \nap   & \xmark \\
\midrule
UniSim~\cite{yang2023unisim}              & \cmark & \xmark & \cmark & \cmark & \pmark & \xmark & \xmark & \xmark & \xmark \\
NeuroNCAP~\cite{ljungbergh2024neuroncap}  & \cmark & \xmark & \cmark & \xmark & \xmark & \xmark & \xmark & \xmark & \xmark \\
HUGSIM~\cite{zhou2026hugsim}              & \cmark & \cmark & \cmark & \xmark & \cmark & \xmark & \xmark & \pmark & \xmark \\
SceneCrafter~\cite{ge2025scenecrafter}    & \cmark & \cmark & \cmark & \xmark & \xmark & \xmark & \cmark & \cmark & \xmark \\
DecoupleGS~\cite{li2026decouplegs}        & \cmark & \cmark & \cmark & \xmark & \xmark & \xmark & \cmark & \pmark & \xmark \\
\midrule
\textbf{AETHER (Ours)}                    & \cmark & \cmark & \cmark & \cmark & \cmark & \cmark & \cmark & \cmark & \cmark \\
\bottomrule
\end{tabular}
\end{table}

\begin{figure*}[t]
  \centering
  \includegraphics[width=\textwidth]{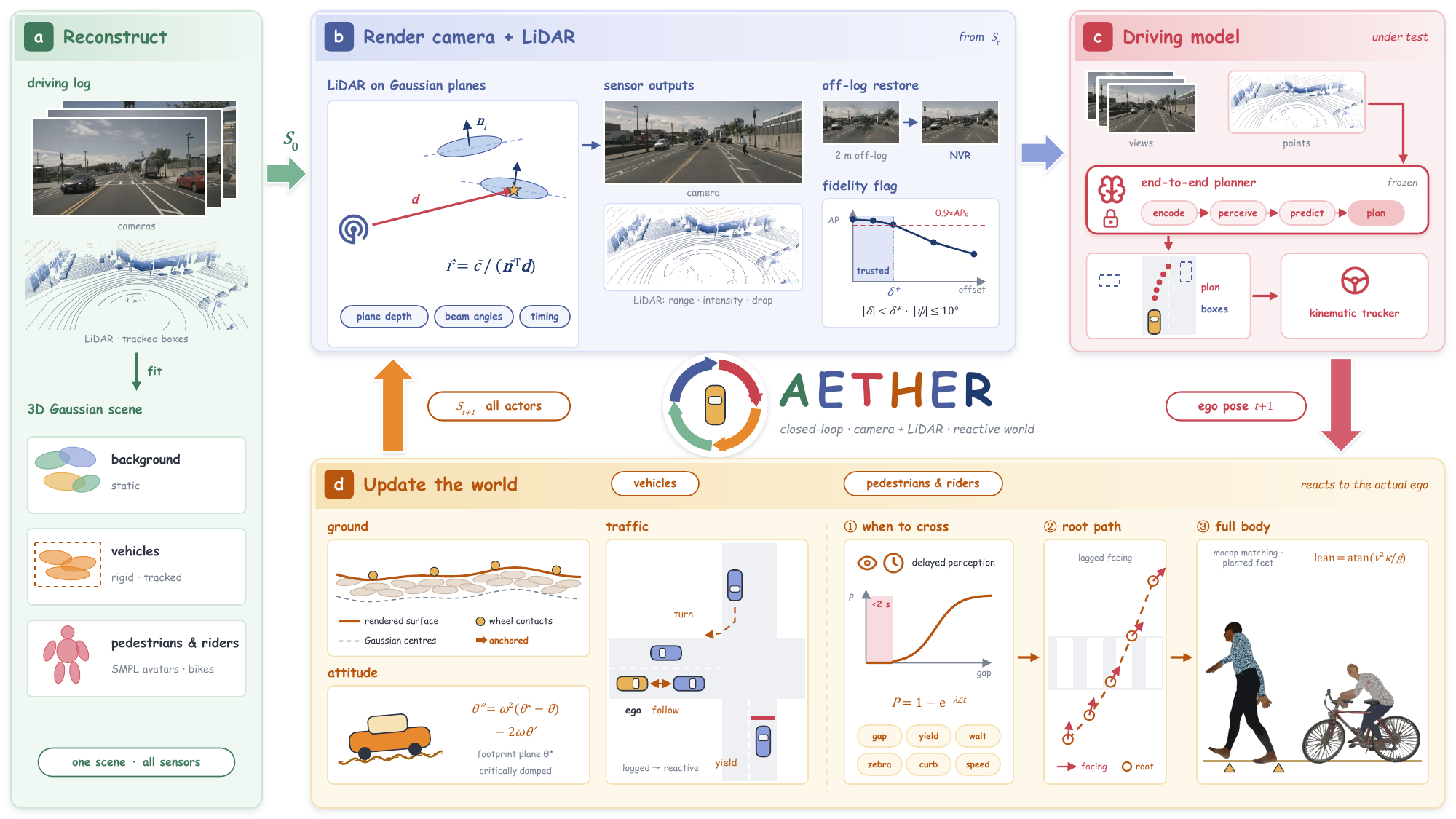}
  \caption{\textbf{Overview of AETHER.}
  \textbf{(a)}~A driving log is reconstructed into one 3D Gaussian scene (static background, rigid vehicles, and inserted SMPL pedestrians and riders) that every sensor shares.
  \textbf{(b)}~From the current state $S_t$, AETHER renders camera images and LiDAR returns whose range comes from the blended local Gaussian planes. NVR restores off-log views, and a fidelity flag marks a step as trusted only while the lateral offset stays below the calibrated $\delta^*$ and the heading error is at most $10^\circ$.
  \textbf{(c)}~The driving model under test receives the rendered sensors unchanged, and a kinematic tracker executes its plan.
  \textbf{(d)}~Around the new ego pose, vehicles stand on a route-anchored ground with damped attitude and drive as reactive traffic, while pedestrians and riders decide when to cross, steer their root, and animate their full body. This yields $S_{t+1}$ and closes the loop.}
  \label{fig:overview}
\end{figure*}

End-to-end (E2E) autonomous driving systems learn to generate driving trajectories directly from raw sensor observations, including camera images and LiDAR point clouds~\cite{hu2023uniad,jiang2023vad,chitta2023transfuser,liu2025gaussianfusion}.
However, their evaluation still relies heavily on open-loop comparisons against logged trajectories, where the environment remains unaffected by the model's actions.
Such evaluations fail to capture the accumulation of prediction and planning errors and their consequences for interactions with surrounding traffic, limiting their ability to reflect actual driving performance~\cite{codevilla2018offline,dauner2023parting,li2024egostatus}.
Closed-loop evaluation is therefore essential, yet real-world testing is costly, potentially unsafe, and difficult to scale, making simulation an increasingly important alternative.
A reliable closed-loop simulator must provide realistic and geometrically consistent multi-sensor observations as the ego vehicle deviates from recorded trajectories, while enabling surrounding road users to respond plausibly to its actions, all at interactive rates.

Existing simulators follow three paradigms (\cref{tab:sim_compare}).
Game-engine simulators~\cite{dosovitskiy2017carla} are fully controllable, but their synthetic rendering leaves a large domain gap to real sensor data, and their handcrafted assets are costly to scale.
Generative simulators~\cite{yang2025drivearena,you2024bench2driver,leng2026dreamstream} are photorealistic; yet, without explicit 3D structure, they suffer from spatio-temporal inconsistency and hallucination at a high computational cost.
Reconstruction-based simulators~\cite{yang2023unisim,ljungbergh2024neuroncap,zhou2026hugsim,ge2025scenecrafter,li2026decouplegs} recover real scenes with neural fields or Gaussian splatting, striking a favorable balance between realism, geometric consistency and real-time rendering.
Their fidelity, however, remains bounded by the log: observations degrade once the ego leaves the recorded trajectory, most systems render cameras only, and road users are replayed or scripted, with articulated pedestrians and riders largely absent.

These shortcomings share one origin: reconstruction is constrained only by what the log observed, whereas closed-loop evaluation continually drives the scene into configurations the log never covered.
Two difficulties follow.
\emph{(i)~Observation extrapolation.}
Off-log viewpoints are under-constrained, so renderings deteriorate progressively rather than fail abruptly.
Restoring them in image space~\cite{wu2025difix3d} may change content that the LiDAR, rendered from the same scene, does not contain. As quality also decays without a clear limit, a planner failure cannot be told apart from a rendering artifact.
\emph{(ii)~Non-rigid foreground synthesis.}
New pedestrian and rider motion must at once respond to the ego, keep the articulated pose consistent with the root trajectory, and stay in contact with a reconstructed ground that is itself noisy and biased.
Modeling these coupled levels in isolation, as scripted triggers, social forces~\cite{helbing1995social} or clip playback do, yields sliding feet, floating bodies, and implausible reactions; the same ground also makes vehicles float or jitter, which the ego's cameras pass on to the planner.

Our key insight is that a log should serve as an \emph{anchor} rather than a \emph{script}: whatever the ego's departure invalidates is regenerated by a model tied to the real data that still constrains it, and how far this tie holds is itself measured.
Building on this insight, we propose AETHER, a reconstruction-based testbed for sensor-level closed-loop evaluation of end-to-end driving systems.
Camera and LiDAR are rendered from shared Gaussian geometry; off-log views are repaired by a novel-view restorer that leaves this geometry intact, and a fidelity radius calibrated on the planner's own perception marks each step as trusted or not.
Vehicles move in reactive traffic on a route-anchored ground, while pedestrians and riders follow a hierarchical behavior--root--body model whose decisions respond to the ego and whose contacts are enforced on the ground.
Experiments show that AETHER provides faithful multi-sensor observations and plausible interactive road users, enabling reliable closed-loop evaluation.
Our contributions are as follows:
\begin{itemize}
    \item A reconstruction-based closed-loop testbed that unifies camera--LiDAR sensing, reactive traffic and interactive articulated pedestrians and riders for sensor-level evaluation of end-to-end driving systems beyond the recorded log.
    \item A log-anchored formulation of closed-loop simulation that resolves its two core difficulties through cross-modally consistent observation extrapolation with a perception-calibrated trust region, and contact-coupled hierarchical synthesis of responsive non-rigid road users.
    \item A fidelity-aware evaluation protocol that separates planner failures from simulation artifacts, revealing that log-replay evaluation systematically misattributes collisions to the planner under test.
\end{itemize}
\section{Related Work}
\label{sec:related}

\paragraph{Closed-loop driving simulation.}
Beyond game engines~\cite{dosovitskiy2017carla} and generative world models~\cite{yang2025drivearena,you2024bench2driver,leng2026dreamstream}, reconstruction-based simulators re-render real scenes along the ego's new trajectory.
UniSim~\cite{yang2023unisim} renders camera and LiDAR for closed-loop testing, and NeuroNCAP~\cite{ljungbergh2024neuroncap} builds safety-critical scenarios on NeuRAD~\cite{tonderski2024neurad}.
HUGSIM~\cite{zhou2026hugsim}, SceneCrafter~\cite{ge2025scenecrafter} and DecoupleGS~\cite{li2026decouplegs} bring 3DGS-based evaluation to real-time rates, and RAD~\cite{gao2025rad} uses such environments for reinforcement learning.
Most render cameras only, and none report whether its observations remain reliable off the recorded trajectory.

\paragraph{Camera and LiDAR novel-view synthesis.}
Driving scenes are commonly reconstructed as compositional Gaussians with separate background, vehicles and humans~\cite{yan2024streetgaussians,chen2025omnire}.
LiDAR simulation has evolved from ray casting on meshes~\cite{manivasagam2020lidarsim} to neural fields~\cite{zheng2024lidar4d}, and some methods share one representation between camera and LiDAR~\cite{yang2023unisim,tonderski2024neurad,hess2025splatad}.
Their quality drops away from the recorded trajectory.
Off-trajectory views are recovered by distilling generative priors into the reconstruction~\cite{zhao2025drivedreamer4d,ni2025recondreamer,yan2025streetcrafter} or by single-step image restoration~\cite{wu2025difix3d}, both of which repair cameras alone and leave the remaining fidelity unquantified.

\paragraph{Simulating traffic participants.}
Reactive vehicles are driven by car-following models~\cite{treiber2000idm} or microscopic traffic simulators~\cite{lopez2018sumo,wen2023limsim}, as in nuPlan~\cite{caesar2021nuplan}, whereas reconstructed scenes mostly replay logged vehicles with poses from logged boxes.
Pedestrian behavior is classically modeled by social forces~\cite{helbing1995social} or reciprocal collision avoidance~\cite{vandenberg2011orca}, and character animation synthesizes full-body motion from motion capture~\cite{mahmood2019amass,loper2015smpl} through motion matching~\cite{holden2020learned}.
Driving simulators rarely combine the two: crossing decisions, root motion, body pose, and ground contact are not modeled jointly, and the response to the ego is limited or absent.
\section{Method}
\label{sec:method}

\subsection{Overview}
\label{sec:method_overview}

AETHER turns a driving log into a closed-loop environment in which an end-to-end planner $\pi$ drives the ego (\cref{fig:overview}).
At step $t$, the scene state $\mathcal{S}_t$ comprises the poses of the ego and all vehicles and the root states and body poses of all pedestrians and riders.
The simulator renders camera images $\mathbf{I}_t$ and a LiDAR sweep $\mathbf{P}_t$ from $\mathcal{S}_t$, the ego tracks the planned trajectory $\pi(\mathbf{I}_t,\mathbf{P}_t)$, and all other road users respond to the new ego state, yielding $\mathcal{S}_{t+1}$.
Following the anchoring principle, every component regenerates what the ego's departure invalidates while remaining tied to a real-data source: the sensors to the reconstructed geometry and to real images (\cref{sec:method_sensor}), vehicles to the measured road contact (\cref{sec:method_vehicle}), and pedestrians and riders to recorded crossings and motion capture (\cref{sec:method_vru}).

\subsection{Sensor Fidelity off the Log}
\label{sec:method_sensor}

\paragraph{Shared-geometry camera and LiDAR rendering.}
The scene is represented by Gaussians $\{\mathcal{G}_i\}$ with means $\boldsymbol{\mu}_i$, covariances $\boldsymbol{\Sigma}_i$, opacities $o_i$ and colors $\mathbf{c}_i$, decomposed into a static background, rigid vehicles and deformable humans~\cite{chen2025omnire}.
Instead of fitting a separate LiDAR representation, we render LiDAR from the same Gaussians, each augmented with a feature $\mathbf{f}_i$ and a normal $\mathbf{n}_i$ given by its shortest axis.
For a ray $\mathbf{r}=(\mathbf{o},\mathbf{d})$ with compositing weights $w_i=\alpha_i\prod_{j<i}(1-\alpha_j)$, the range is the intersection of the ray with the blended local plane,
\begin{equation}
  \hat{r}(\mathbf{r}) = \frac{\bar{c}}{\bar{\mathbf{n}}^{\top}\mathbf{d}}, \quad
  \bar{c}=\sum\nolimits_i w_i\, \mathbf{n}_i^{\top}(\boldsymbol{\mu}_i-\mathbf{o}),\quad \bar{\mathbf{n}}=\sum\nolimits_i w_i \mathbf{n}_i .
  \label{eq:range}
\end{equation}
\Cref{eq:range} carries the planar depth of PGSR~\cite{chen2024pgsr} from camera depth to the rays of a spinning LiDAR. It is exact on locally planar surfaces, whereas blending the ranges of Gaussian centers, as in SplatAD~\cite{hess2025splatad}, is biased at grazing incidence, the dominant regime on roads and walls.
Intensity and ray drop are decoded from the composited feature and the sensing geometry,
\begin{equation}
  \big(\hat{e},\, \hat{\ell}_{\text{drop}}\big) = \Phi\Big(\textstyle\sum_i w_i \mathbf{f}_i,\; \cos\theta,\; \hat{r},\; \mathbf{b}\Big),
  \label{eq:lidar_head}
\end{equation}
where $\theta$ is the incidence angle, $\mathbf{b}$ embeds the laser index and $\Phi$ is a small MLP.
Both sensors are optimized jointly, so LiDAR supervision refines the geometry shared with the cameras rather than a separate copy, and any change of a surface appears consistently in both modalities.

\paragraph{Novel-view restoration.}
Off the log, renderings degrade into blur, floaters and holes where the reconstruction is under-constrained.
We learn a novel-view restorer (NVR) $\mathcal{R}$ that maps such renderings back to the distribution of real images.
Since off-log views have no ground truth, we synthesize the degradation on logged views: renderings $\tilde{\mathbf{I}}$ from under-converged reconstructions, whose artifacts resemble those of under-constrained regions, are paired with the real images $\mathbf{I}$, and $\mathcal{R}$ is trained to minimize a reconstruction and perceptual loss $\mathcal{L}\big(\mathcal{R}(\tilde{\mathbf{I}}),\mathbf{I}\big)$ on scenes held out from evaluation.
At test time, $\mathcal{R}$ is applied in a single step without per-scene optimization.
Because it edits appearance only, the shared geometry, and thus the LiDAR, stays untouched; whether the restored image still agrees with this geometry is measured in \cref{sec:exp_cam}.

\paragraph{Deviation-aware fidelity flag.}
Restoration enlarges, but does not remove, the region in which renderings can be trusted.
We delimit this region in terms of what the planner consumes, namely its own perception.
For a planner with detection head $h$, frames are grouped by the ego's lateral deviation $\delta$ from the log, and the retention of each group is
\begin{equation}
  \rho(\delta) = \frac{\mathrm{AP}\big(h(\mathbf{I}),\mathcal{B}\big)\big|_{\delta}}{\mathrm{AP}\big(h(\mathbf{I}),\mathcal{B}\big)\big|_{\delta\approx 0}},
  \label{eq:retention}
\end{equation}
where $\mathcal{B}$ are the ground-truth boxes of the reconstructed actors.
The fidelity radius is $\delta^{\star}=\sup\{\delta : \rho(\delta')\ge\tau,\ \forall\,\delta'\le\delta\}$, calibrated once per planner and restorer.
$\rho$ measures the loss caused by leaving the log relative to rendering on it; the remaining gap to real images on the log is reported separately (\cref{sec:exp_cam}).
A simulation step is flagged \emph{trusted} if the ego's lateral and heading deviations stay below $\delta^{\star}$ and $\psi_{\max}$, so that every closed-loop result is reported with the fraction of trusted steps and the flag state at each collision.

\subsection{Grounded Vehicles and Reactive Traffic}
\label{sec:method_vehicle}

\paragraph{Route-anchored ground.}
Because Gaussian road surfaces form a slab below the visible road, we estimate the ground $z_0(\mathbf{x})$ from rendered depth rather than from Gaussian centers.
This surface still carries decimeter-scale ripples, whereas the logged ego trajectory $\boldsymbol{\gamma}(s)$ directly measures where the wheels touched the road.
We therefore anchor the surface to this contact profile,
\begin{equation}
  z(\mathbf{x}) = z_0(\mathbf{x}) + \eta\big(d(\mathbf{x})\big)\,\big(g_\sigma * \epsilon\big)\big(s(\mathbf{x})\big),
  \label{eq:ground}
\end{equation}
where $\epsilon(s) = z_{\boldsymbol{\gamma}}(s) - z_0(\boldsymbol{\gamma}(s))$ is the contact residual along the route, $s(\mathbf{x})$ and $d(\mathbf{x})$ are the arc length and distance of the nearest route point, $g_\sigma$ smooths $\epsilon$ along the arc length, and $\eta$ decays from one near the route to zero a few meters away.
The surface thus follows the measured road along the route while keeping its reconstructed cross-fall and off-route shape.

\paragraph{Consistent attitude.}
Evaluating $z$ under each wheel would transfer reconstruction noise to the body.
Instead, a plane fitted to $z$ over the vehicle footprint yields a target height, pitch $\bar{\vartheta}$ and roll $\bar{\varphi}$, which the body follows through a critically damped response,
\begin{equation}
  \ddot{\vartheta} = \omega^2(\bar{\vartheta}-\vartheta) - 2\omega\,\dot{\vartheta},
  \label{eq:suspension}
\end{equation}
and likewise for $\varphi$, with $\omega$ in the range of passenger-car suspensions.
The response converges to the road grade without bias while filtering frame-to-frame fluctuations, which matters most for the ego's body-mounted cameras.

\paragraph{Reactive traffic.}
Rather than replaying logged vehicles blindly, each follows its recorded path and speed until the ego or another vehicle blocks it; an IDM~\cite{treiber2000idm} car-follower then brakes for it before it returns to its logged timeline.
Added vehicles keep at least five moving vehicles near the ego, so the traffic keeps the density of the recorded scene while responding to the planner.

\begin{figure*}[t]
  \centering
  \includegraphics[width=\textwidth]{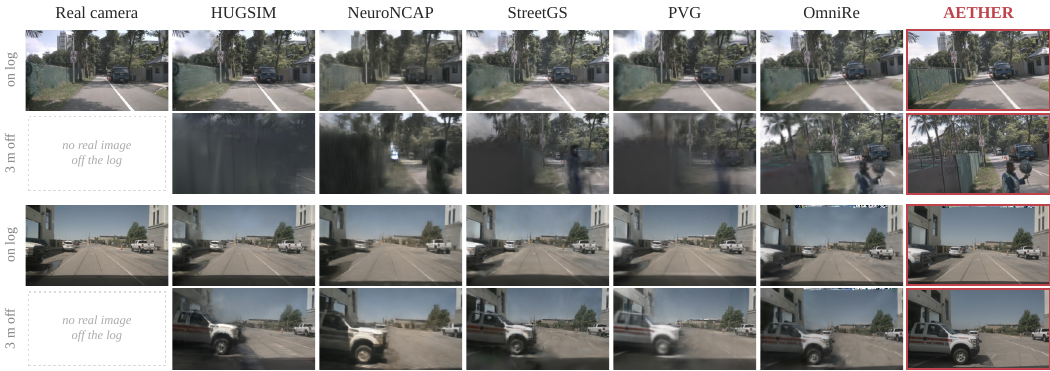}
  \caption{\textbf{Front camera on the log and \SI{3}{m} off it.}}
  \label{fig:sensor_qual}
\end{figure*}
\subsection{Interactive Pedestrians and Riders}
\label{sec:method_vru}

A pedestrian is driven by three coupled layers, each anchored to its own real-data source: a \emph{behavior} layer decides where and when to move, a \emph{root} layer turns the decision into locomotion, and a \emph{body} layer produces a full-body pose consistent with that locomotion and the ground.
The agent observes the ego and other vehicles only through their current states, delayed by a reaction time; it never sees their future, so changes in the ego's driving affect it only through what it can perceive.

\paragraph{Behavior.}
A state machine moves through approach, wait, start and cross, with hesitate, run or retreat responses to threats during crossing.
The central decision, when to step onto the road, is modeled as a discrete-time hazard: near the curb, the probability of entering within $\Delta t$ is $1-\exp(-\lambda\Delta t)$, with
\begin{equation}
  \log\lambda = \beta_0 + \beta_g\, g + \beta_y\, y + \beta_w\, t_w + \beta_c\, c + \beta_d\, d + \beta_v\, v,
  \label{eq:hazard}
\end{equation}
where $g$ is the perceived gap to the most critical approaching vehicle, $y$ indicates whether it yields, $t_w$ is the waiting time, $c$ indicates a marked crosswalk, $d$ is the distance to the curb and $v$ the walking speed.
The coefficients are fitted by maximum likelihood on recorded pedestrian road entries in nuScenes~\cite{caesar2020nuscenes}, with the rarely observed yielding term set as a prior.
Gap acceptance thus responds to the ego's actual speed and distance, unlike scripted triggers that fire at a fixed time or distance.

\begin{figure*}[t]
  \centering
  \includegraphics[width=\textwidth]{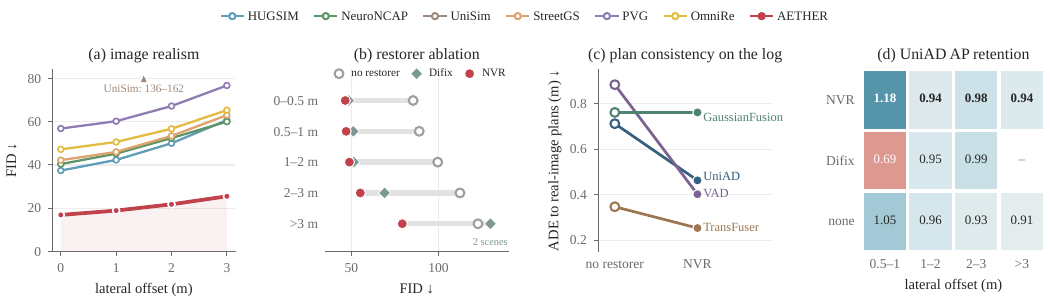}
  \caption{\textbf{Camera fidelity.}
  (a) FID vs.\ lateral offset.
  (b) FID of restorers by offset.
  (c) Plan consistency of four planners on the log.
  (d) UniAD AP relative to the same frame on the log.}
  \label{fig:sensor_quant}
\end{figure*}
\paragraph{Root.}
Speed, walking direction and body facing are coupled states that follow their commands through bounded first- and second-order responses fitted to the start-up profiles of real pedestrians.
A pedestrian starting from rest while facing away from the road therefore turns as it starts walking, which a prescribed trajectory cannot reproduce.

\paragraph{Body and contact.}
Full-body poses are selected from motion capture~\cite{mahmood2019amass} by motion matching~\cite{holden2020learned}, querying with the root's own prediction of its path and facing together with the current foot states.
Contacts labeled in the database are enforced against the reconstructed ground: a planted foot is locked on $z$ and the legs are adjusted by inverse kinematics, so the body stays consistent with both its root motion and the terrain.

\begin{figure}[t]
  \centering
  \includegraphics[width=\columnwidth]{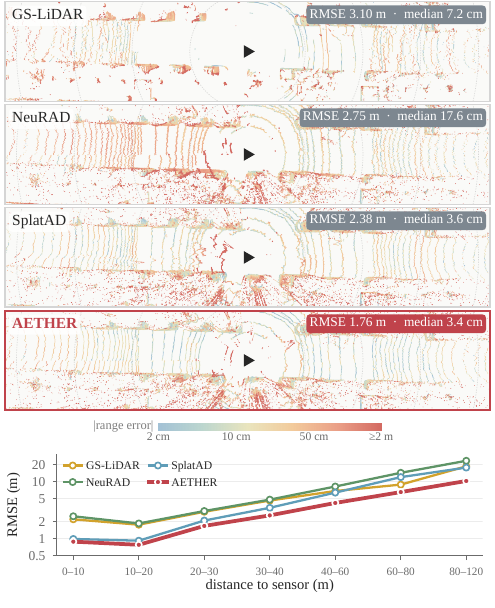}
  \caption{\textbf{LiDAR range error.} Top: a held-out sweep in BEV. Bottom: RMSE vs.\ distance.}
  \label{fig:lidar_qual}
\end{figure}

\paragraph{Riders.}
Riders follow a kinematic bicycle model whose vehicle leans by $\arctan(v^2\kappa/g_0)$ at speed $v$, path curvature $\kappa$ and gravitational acceleration $g_0$.
Hands, pelvis and feet are attached to the grips, saddle and pedals, and pedaling is coupled to wheel speed.
The behavior layer uses the same perception and hazard form, with yield, stop and brake decisions in place of a curb crossing.

\section{Experiments}
\label{sec:exp}

We evaluate sensor fidelity once the ego leaves the log, including where the fidelity flag places its limit (\cref{sec:exp_cam,sec:exp_lidar}), the grounding of inserted vehicles (\cref{sec:exp_ground}), the motion and responsiveness of pedestrians and riders (\cref{sec:exp_vru}), and the effect of reactive traffic on closed-loop evaluation (\cref{sec:exp_loop}).

\begin{table}[t]
  \centering
  \caption{\textbf{Camera fidelity on and off the log} on 17 HUGSIM scenes. FID and each planner's detection mAP on the log (0\,m) and \SI{3}{m} off it; ADE (m) between plans from rendered and real images on the log. $^\dagger$Re-implementation. \rankkey}
  \label{tab:camera}
  \footnotesize
  \setlength{\tabcolsep}{1.15pt}
  \renewcommand{\arraystretch}{1.08}
  \begin{tabular}{@{}lcccccccc@{}}
    \toprule
    & \multicolumn{2}{c}{FID $\downarrow$} & \multicolumn{2}{c}{UniAD mAP $\uparrow$} & \multicolumn{2}{c}{VAD mAP $\uparrow$} & \multicolumn{2}{c}{ADE $\downarrow$} \\
    \cmidrule(lr){2-3}\cmidrule(lr){4-5}\cmidrule(lr){6-7}\cmidrule(l){8-9}
    Method & 0\,m & 3\,m & 0\,m & 3\,m & 0\,m & 3\,m & UniAD & VAD \\
    \midrule
    \reff{Real images} & \reff{--} & \reff{--} & \reff{.380} & \reff{--} & \reff{.202} & \reff{--} & \reff{--} & \reff{--} \\
    \midrule
    UniSim$^\dagger$~\cite{yang2023unisim}   & 136.4 & 162.1 & .107 & .091 & .043 & .034 & 1.45 & .93 \\
    NeuroNCAP~\cite{ljungbergh2024neuroncap} & \third{40.4} & \second{60.0} & .125 & .104 & .056 & .043 & .80 & .89 \\
    HUGSIM~\cite{zhou2026hugsim}             & \second{37.4} & \third{60.7} & .241 & .197 & \second{.111} & .068 & \third{.52} & \second{.51} \\
    \midrule
    PVG~\cite{chen2023pvg}                   & 56.8 & 76.7 & .232 & .161 & .064 & .044 & .61 & .86 \\
    StreetGS~\cite{yan2024streetgaussians}   & 42.2 & 63.0 & \third{.253} & \third{.215} & \third{.107} & \third{.080} & \second{.46} & \third{.70} \\
    OmniRe~\cite{chen2025omnire}             & 47.2 & 65.3 & \second{.263} & \second{.235} & \third{.107} & \second{.081} & .56 & .85 \\
    \midrule
    \ours & \first{16.9} & \first{25.5} & \first{.265} & \first{.242} & \first{.170} & \first{.149} & \first{.41} & \first{.38} \\
    \bottomrule
  \end{tabular}
\end{table}

\subsection{Setup}
\label{sec:exp_setup}

\noindent\textbf{Data and planners.}
Closed-loop and camera experiments use reconstructed nuScenes~\cite{caesar2020nuscenes} scenes, and LiDAR is evaluated on held-out PandaSet~\cite{xiao2021pandaset} sweeps.
Scenes used to train NVR are excluded from all evaluations.
We test UniAD~\cite{hu2023uniad}, VAD~\cite{jiang2023vad}, TransFuser~\cite{chitta2023transfuser} in its camera-only NAVSIM variant~\cite{dauner2024navsim} and GaussianFusion~\cite{liu2025gaussianfusion} with their public weights, unmodified.
Scene lists and run settings are given in the supplementary material.

\noindent\textbf{Metrics.}
Image realism is measured by FID~\cite{heusel2017gans}, perception by the BEV detection mAP of each planner's own head, and plan consistency by the ADE between plans from rendered and real images.
Closed-loop driving uses the PDM score~\cite{dauner2024navsim}; a collision counts as the ego's fault unless the ego is stationary or is rear-ended by a faster vehicle.

\begin{table}[t]
  \centering
  \caption{\textbf{LiDAR novel-view synthesis} on held-out PandaSet sweeps. CD in m, MedSE in m$^2$. \rankkey}
  \label{tab:lidar}
  \footnotesize
  \setlength{\tabcolsep}{2.4pt}
  \renewcommand{\arraystretch}{1.08}
  \begin{tabular}{@{}lcccccc@{}}
    \toprule
    & \multicolumn{4}{c}{Geometry} & \multicolumn{2}{c}{Appearance} \\
    \cmidrule(lr){2-5}\cmidrule(l){6-7}
    Method & CD $\downarrow$ & F $\uparrow$ & RMSE $\downarrow$ & MedSE $\downarrow$ & Int. $\downarrow$ & Drop $\uparrow$ \\
    \midrule
    LiDAR-GS~\cite{chen2024lidargs} & \third{.357} & .461 & 18.34 & .4234 & .0957 & .772 \\
    GS-LiDAR~\cite{jiang2025gslidar}  & .508 & \first{.495} & 9.44 & .0742 & .0940 & .825 \\
    NeuRAD~\cite{tonderski2024neurad} & .414 & .183 & \third{7.53} & \third{.0388} & \third{.0887} & \third{.957} \\
    SplatAD~\cite{hess2025splatad}    & \second{.337} & \third{.470} & \second{3.75} & \second{.0082} & \first{.0588} & \first{.967} \\
    \midrule
    \ours & \first{.314} & \second{.474} & \first{2.85} & \first{.0066} & \second{.0626} & \second{.966} \\
    \bottomrule
  \end{tabular}
\end{table}

\subsection{Camera}
\label{sec:exp_cam}

\noindent\textbf{Realism.}
We compare with the simulators HUGSIM~\cite{zhou2026hugsim}, NeuroNCAP~\cite{ljungbergh2024neuroncap} and UniSim~\cite{yang2023unisim} (re-implemented), and with the reconstruction methods Street Gaussians~\cite{yan2024streetgaussians}, PVG~\cite{chen2023pvg} and OmniRe~\cite{chen2025omnire}, which is also our backbone without NVR, on the scenes released by HUGSIM (\cref{fig:sensor_qual,tab:camera}).
AETHER more than halves the FID of every baseline both on and off the log, and \SI{3}{m} off the log it is still lower than any baseline on the log (\cref{fig:sensor_quant}a).
The planners see the same difference: AETHER gives the highest detection mAP for both planners at both offsets, VAD keeps nearly twice the mAP of the best baseline \SI{3}{m} off the log, and both planners plan closest to their real-image behavior.

\begin{table}[t]
  \centering
  \caption{\textbf{Pedestrians and riders.} Root motion against 1{,}083 recorded pedestrian tracks (ADE in m, facing error in degrees, Wasserstein distances of acceleration and yaw rate); body motion (skate in cm/s, penetration in cm, jump in degrees); interaction with the ego (negative-control deviation in m); riders (mean hand, foot and pelvis gap in cm; share of riding frames that pedal and of braking frames that still pedal; foot height at a stop in cm). Best in \textbf{bold}.}
  \label{tab:vru}
  \footnotesize
  \setlength{\tabcolsep}{3.2pt}
  \renewcommand{\arraystretch}{1.06}
  \begin{tabular}{@{}clcccc@{}}
    \toprule
    & & ADE $\downarrow$ & Facing $\downarrow$ & $W_1^{a}$ $\downarrow$ & $W_1^{\omega}$ $\downarrow$ \\
    \cmidrule(l){3-6}
    \grp{3}{Root} & Scripted path   & .287 & 9.1 & .088 & .073 \\
    & Social force~\cite{helbing1995social} & .294 & 8.3 & .086 & .022 \\
    & \ours           & \textbf{.267} & \textbf{6.9} & \textbf{.083} & \textbf{.015} \\
    \midrule
    & & Skate $\downarrow$ & Penetr.\ $\downarrow$ & Jerk $\downarrow$ & Jump $\downarrow$ \\
    \cmidrule(l){3-6}
    \grp{4}{Body} & \reff{Mocap (ref.)} & \reff{7.7} & \reff{.08} & \reff{28} & \reff{11.5} \\
    & Clip playback   & 43.5 & .08 & 450 & 27.3 \\
    & Motion matching & 18.1 & .17 & \textbf{83} & \textbf{14.1} \\
    & \ours           & \textbf{10.3} & \textbf{.00} & 88 & 19.0 \\
    \midrule
    & & Coll.\ $\downarrow$ & Reacts $\uparrow$ & Earlier $\uparrow$ & Neg.\ ctrl $\downarrow$ \\
    \cmidrule(l){3-6}
    \grp{4}{Interact.} & Scripted trigger      & \textbf{.000} & \textbf{1.00} & .00 & 14.3 \\
    & Social force~\cite{helbing1995social} & .225 & .20 & .00 & \textbf{0.0} \\
    & SF + ORCA~\cite{vandenberg2011orca}   & .100 & .45 & .17 & \textbf{0.0} \\
    & \ours                                 & .008 & \textbf{1.00} & \textbf{.92} & \textbf{0.0} \\
    \midrule
    & & Contact $\downarrow$ & Pedal $\uparrow$ & Brake $\downarrow$ & Stop $\downarrow$ \\
    \cmidrule(l){3-6}
    \grp{4}{Rider} & Clip playback & 23.3 & \textbf{1.00} & .59 & 11.1 \\
    & Rigid rider   & \textbf{0.0} & .00 & \textbf{.00} & 29.0 \\
    & Procedural IK & 0.3 & .98 & .79 & 15.7 \\
    & \ours         & 0.3 & \textbf{1.00} & \textbf{.00} & \textbf{1.5} \\
    \bottomrule
  \end{tabular}
\end{table}

\noindent\textbf{Restorer ablation.}
Without a restorer, FID nearly doubles up to \SI{3}{m}, while the released Difix~\cite{wu2025difix3d} is on par with NVR near the log but degrades beyond \SI{2}{m} (\cref{fig:sensor_quant}b).
NVR also brings the plans of all camera-driven planners closer to their real-image plans, reducing ADE by 27--55\%, whereas GaussianFusion is insensitive to the restorer (\cref{fig:sensor_quant}c).
In closed loop, VAD barely drives without NVR (route completion 0.10 vs.\ 0.42).
Restoration does not move image content: on actors found by Mask R-CNN~\cite{he2017maskrcnn}, the optical flow~\cite{teed2020raft} from rendered to restored image is sub-pixel at every offset (median \SI{0.05}{px}), so the image stays aligned with the LiDAR, while 15\% more actors become detectable.

\begin{figure*}[t]
  \centering
  \includegraphics[width=\textwidth]{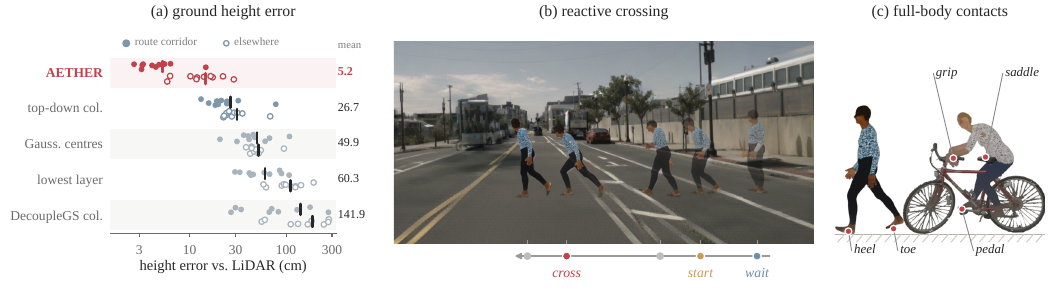}
  \caption{\textbf{Grounded vehicles and interactive pedestrians.}
  (a) Ground height error (dots: scenes, bars: means).
  (b) A pedestrian waits, starts and crosses in front of the ego.
  (c) Enforced foot and rider contacts.}
  \label{fig:actors}
\end{figure*}
\noindent\textbf{Where to trust.}
We compare each offset frame with the frame at the same step on the log.
With NVR, UniAD keeps at least 94\% of its on-log AP in every bin up to more than \SI{3}{m}, so its trusted offset is $\delta^*{=}\SI{3}{m}$, whereas released Difix already falls to 69\% between 0.5 and \SI{1}{m} (\cref{fig:sensor_quant}d).
VAD keeps 96\% of its mAP at \SI{2}{m} but 88\% at \SI{3}{m} (supp.), giving $\delta^*{=}\SI{2}{m}$ with $\tau{=}0.9$; we use $\psi_{\max}{=}\ang{10}$.
In closed loop, NVR raises the UniAD score at every start offset (0.52 vs.\ 0.47 on the log, 0.25 vs.\ 0.23 at \SI{3}{m}), yet the score still halves from \SI{0}{m} to \SI{3}{m}. As UniAD's detection is retained over this range, lost detection does not explain the drop, although other perception outputs may still degrade.

\begin{figure*}[t]
  \centering
  \includegraphics[width=\textwidth]{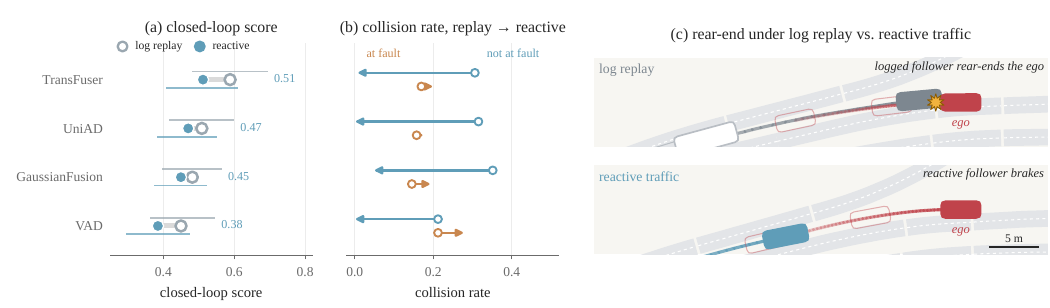}
  \caption{\textbf{Closed-loop evaluation.}
  (a) Score with 95\% CIs. (b) Collision rate from log replay to reactive traffic.
  (c) The same episode under both traffic modes.}
  \label{fig:closedloop}
\end{figure*}
\subsection{LiDAR}
\label{sec:exp_lidar}

We compare with SplatAD~\cite{hess2025splatad}, NeuRAD~\cite{tonderski2024neurad}, GS-LiDAR~\cite{jiang2025gslidar} and LiDAR-GS~\cite{chen2024lidargs} on the same rays (\cref{tab:lidar}).
Sharing geometry with the camera branch gives the most accurate range, with a depth RMSE 24\% lower than SplatAD's and 62\% lower than NeuRAD's; GS-LiDAR reaches a higher F-score but more than three times our RMSE.
Errors drop most on the road surface and building facades, and the gap widens with distance (\cref{fig:lidar_qual}).
Intensity and ray drop remain slightly behind SplatAD.
Sharing the camera reconstruction, the LiDAR needs only 4k extra iterations (\SI{14}{min}) against 30k for SplatAD.
At equal budget, center ranges instead of \cref{eq:range} raise CD to 0.319 and MedSE to .0074 with RMSE unchanged, while freezing the shared geometry raises CD to 0.427 and RMSE to 3.54; joint optimization brings most of the gain (supp.).

\subsection{Grounded vehicles}
\label{sec:exp_ground}

We compare the ground surface with accumulated LiDAR ground (\cref{fig:actors}a).
Surfaces read from Gaussian centres or vertical Gaussian columns err by 0.3 to \SI{1.9}{m}, whereas ours errs by \SI{5.2}{cm} in the driven corridor and \SI{14.5}{cm} elsewhere.
Anchoring it to the logged wheel contacts keeps this accuracy and lowers the rotation error of the simulated camera against the logged one from \ang{0.75} to \ang{0.64}, on par with a flat plane on these mostly flat roads.

\begin{table}[t]
  \centering
  \caption{\textbf{Component ablation in closed loop.} Fraction of episodes whose collision outcome flips relative to the full model, and the UniAD score. The first row is the run-to-run noise of the full model.}
  \label{tab:ablation}
  \footnotesize
  \setlength{\tabcolsep}{4pt}
  \renewcommand{\arraystretch}{1.08}
  \begin{tabular}{@{}lccc@{}}
    \toprule
    & \multicolumn{2}{c}{Outcome flip} & \\
    \cmidrule(lr){2-3}
    Variant & UniAD & TransFuser & UniAD score \\
    \midrule
    \reff{Full model} & \reff{.05--.07} & \reff{.06--.10} & \reff{.439} \\
    \midrule
    Sparse IDM traffic       & \textbf{.52} & \textbf{.45} & .590 \\
    w/o NVR                  & .10 & .15 & .372 \\
    Coarse ground            & .03 & .02 & .409 \\
    w/o suspension           & .03 & .00 & .417 \\
    Nominal extrinsics       & .03 & .00 & .410 \\
    \bottomrule
  \end{tabular}
\end{table}

\subsection{Pedestrians and riders}
\label{sec:exp_vru}

\noindent\textbf{Motion.}
Against recorded pedestrian tracks, our root model reproduces facing and turning dynamics more closely than scripted paths or social forces (\cref{tab:vru}).
With the same root trajectories, clip playback skates at \SI{43.5}{cm/s}, while our contact-aware motion matching reduces this to \SI{10.3}{cm/s}, close to mocap, and removes ground penetration.
A rigid rider and procedural IK also keep riders on the bike, but ignore the riding state: the former never pedals and hangs its foot \SI{29}{cm} above the ground at a stop, the latter pedals while braking; ours does neither and alone leans into turns.
Inserted actors also stay visible to the planners: with NVR, UniAD detects 69\% of inserted pedestrians, against 54\% without it.
Plain motion matching has slightly lower jerk and pose jumps, which our contact constraints trade for less skating.

\noindent\textbf{Behavior.}
Our crossing model halves the held-out negative log-likelihood of a constant hazard (0.13 vs.\ 0.30).
In closed loop, our pedestrians react to every ego intervention: they enter earlier in 92\% of cases when the ego yields, collide only when the ego holds a constant speed without reacting, and stay still when the ego is irrelevant.
A scripted trigger fires even for an ego on a parallel road, and social-force models~\cite{helbing1995social,vandenberg2011orca} collide in 10 to 23\% of interventions.
\cref{fig:actors}b shows a pedestrian waiting at the kerb and crossing in front of the ego, and \cref{fig:actors}c the resulting contacts.

\begin{table}[t]
  \centering
  \caption{\textbf{Collisions and fidelity flag in closed loop} (55 held-out scenes). Episodes with collisions under log replay (Rep.) and reactive traffic (Rea.), trusted steps, and reactive at-fault collisions in a trusted step. $^*$Uncalibrated; UniAD's radius.}
  \label{tab:trust}
  \footnotesize
  \setlength{\tabcolsep}{2.6pt}
  \renewcommand{\arraystretch}{1.08}
  \begin{tabular}{@{}lccccccc@{}}
    \toprule
    & & \multicolumn{2}{c}{At fault} & \multicolumn{2}{c}{Not at fault} & Trusted & At fault, \\
    \cmidrule(lr){3-4}\cmidrule(lr){5-6}
    Planner & $\delta^\star$ & Rep. & Rea. & Rep. & Rea. & steps & trusted \\
    \midrule
    TransFuser     & \SI{3}{m}$^*$ & 28 & 32 & 50 & 1 & 82\% & 79\% \\
    UniAD          & \SI{3}{m}     & 26 & 29 & 52 & 0 & 87\% & 76\% \\
    GaussianFusion & \SI{3}{m}$^*$ & 24 & 32 & 58 & 8 & 61\% & 59\% \\
    VAD            & \SI{2}{m}     & 35 & 46 & 35 & 0 & 82\% & 51\% \\
    \midrule
    All            &               & 113 & 139 & 195 & 9 & 78\% & 64\% \\
    \bottomrule
  \end{tabular}
\end{table}

\subsection{Closed-loop evaluation}
\label{sec:exp_loop}

\cref{fig:closedloop} evaluates the four planners on held-out scenes under log replay and under our reactive traffic, which drives the same logged vehicles from the same states along their logged routes.
The ranking is the same in both modes, although adjacent planners are not separable.
The modes disagree on why planners crash. Under log replay, 39 to 47\% of episodes end in a collision, mostly not by the planner's fault: logged vehicles drive into an ego that left its recorded path.
98\% of them are rear-end collisions, and only 6\% follow ego braking above \SI{4}{m/s^2}.
Reactive traffic removes nearly all of them (195 $\rightarrow$ 9 of 656 paired episodes), while at-fault collisions do not change significantly (113 $\rightarrow$ 139, paired difference $+0.04$, 95\% CI $[-0.02, 0.10]$; \cref{fig:closedloop}b and \cref{tab:trust}).
\cref{fig:closedloop}c shows a typical case: a replayed vehicle rear-ends the slowing ego, whereas the reactive follower brakes.
Replay thus mixes planner errors with log artifacts; whether reactive traffic matches real driving is untested.

\noindent\textbf{Trusted steps.}
Over all episodes, 78\% of steps are trusted, limited mostly by heading rather than lateral deviation.
Under reactive traffic, 64\% of at-fault collisions occur in a trusted step and are attributed to the planner; the rest are reported separately (\cref{tab:trust}).

\noindent\textbf{Ablation.}
We swap one component at a time and count the episodes whose collision outcome flips (\cref{tab:ablation}).
Replacing our traffic with a few IDM~\cite{treiber2000idm} vehicles flips about half of the episodes and inflates the score; as it changes both density and behavior, it bounds rather than isolates the effect of the traffic model.
Removing NVR flips 10 to 15\% and lowers the UniAD score from 0.44 to 0.37.
A coarse ground, no suspension or nominal extrinsics stay within run-to-run noise; they serve sensor and pose fidelity (\cref{sec:exp_ground}).

\section{Conclusion}
\label{sec:conclusion}

We presented AETHER, a closed-loop testbed that treats driving logs as anchors rather than scripts: camera and LiDAR are rendered from shared Gaussian geometry and restored off the log under a perception-calibrated fidelity flag, vehicles move in reactive traffic on route-anchored ground, and pedestrians and riders respond to the ego with contact-consistent motion.
Experiments show more faithful sensing than existing simulators, plausible and responsive road users, and a closed-loop evaluation that separates planner failures from artifacts of log replay.
Fidelity still decays beyond a few meters of lateral offset, and traffic lights and weather are not yet modeled; we plan to address these and to use AETHER for closed-loop training.
\section*{LLM Usage Statement}
Large language models were used to assist with language editing and drafting of the text. All methods, experiments, results, and claims were designed, verified, and are taken responsibility for by the authors.

{
    \small
    \bibliographystyle{refstyle}
    \bibliography{main}

@String{CVPR = {Proceedings of the IEEE/CVF Conference on Computer Vision and Pattern Recognition (CVPR)}}

@String{ICCV = {Proceedings of the IEEE/CVF International Conference on Computer Vision (ICCV)}}

@String{ICLR = {International Conference on Learning Representations (ICLR)}}

@String{NIPS = {Advances in Neural Information Processing Systems}}

@String{PAMI = {IEEE Transactions on Pattern Analysis and Machine Intelligence}}

@String{TOG = {ACM Transactions on Graphics}}

@inproceedings{dosovitskiy2017carla,
  title     = {{CARLA}: An Open Urban Driving Simulator},
  author    = {Alexey Dosovitskiy and German Ros and Felipe Codevilla and Antonio Lopez and Vladlen Koltun},
  booktitle = {Proceedings of the 1st Annual Conference on Robot Learning},
  series    = {Proceedings of Machine Learning Research},
  publisher = {PMLR},
  volume    = {78},
  pages     = {1--16},
  year      = {2017},
  url       = {https://proceedings.mlr.press/v78/dosovitskiy17a.html}
}

@inproceedings{yang2023unisim,
  title     = {{UniSim}: A Neural Closed-Loop Sensor Simulator},
  author    = {Ze Yang and Yun Chen and Jingkang Wang and Sivabalan Manivasagam and Wei-Chiu Ma and Anqi Joyce Yang and Raquel Urtasun},
  booktitle = CVPR,
  pages     = {1389--1399},
  year      = {2023},
  doi       = {10.1109/CVPR52729.2023.00140},
  url       = {https://openaccess.thecvf.com/content/CVPR2023/html/Yang_UniSim_A_Neural_Closed-Loop_Sensor_Simulator_CVPR_2023_paper.html}
}

@inproceedings{tonderski2024neurad,
  title     = {{NeuRAD}: Neural Rendering for Autonomous Driving},
  author    = {Adam Tonderski and Carl Lindstr{\"o}m and Georg Hess and William Ljungbergh and Lennart Svensson and Christoffer Petersson},
  booktitle = CVPR,
  pages     = {14895--14904},
  year      = {2024},
  doi       = {10.1109/CVPR52733.2024.01411},
  url       = {https://openaccess.thecvf.com/content/CVPR2024/html/Tonderski_NeuRAD_Neural_Rendering_for_Autonomous_Driving_CVPR_2024_paper.html}
}

@inproceedings{hess2025splatad,
  title     = {{SplatAD}: Real-Time {Lidar} and Camera Rendering with {3D Gaussian Splatting} for Autonomous Driving},
  author    = {Georg Hess and Carl Lindstr{\"o}m and Maryam Fatemi and Christoffer Petersson and Lennart Svensson},
  booktitle = CVPR,
  pages     = {11982--11992},
  year      = {2025},
  doi       = {10.1109/CVPR52734.2025.01119},
  url       = {https://openaccess.thecvf.com/content/CVPR2025/html/Hess_SplatAD_Real-Time_Lidar_and_Camera_Rendering_with_3D_Gaussian_Splatting_CVPR_2025_paper.html}
}

@article{zhou2026hugsim,
  title     = {{HUGSIM}: A Real-Time, Photo-Realistic and Closed-Loop Simulator for Autonomous Driving},
  author    = {Hongyu Zhou and Longzhong Lin and Jiabao Wang and Yichong Lu and Dongfeng Bai and Bingbing Liu and Yue Wang and Andreas Geiger and Yiyi Liao},
  journal   = {IEEE Transactions on Pattern Analysis and Machine Intelligence},
  volume    = {48},
  number    = {4},
  pages     = {4673--4691},
  year      = {2026},
  doi       = {10.1109/TPAMI.2025.3647952},
  url       = {https://doi.org/10.1109/TPAMI.2025.3647952}
}

@inproceedings{yang2025drivearena,
  title     = {{DriveArena}: A Closed-loop Generative Simulation Platform for Autonomous Driving},
  author    = {Xuemeng Yang and Licheng Wen and Tiantian Wei and Yukai Ma and Jianbiao Mei and Xin Li and Wenjie Lei and Daocheng Fu and Pinlong Cai and Min Dou and Liang He and Yong Liu and Botian Shi and Yu Qiao},
  booktitle = ICCV,
  pages     = {26933--26943},
  year      = {2025},
  doi       = {10.1109/ICCV51701.2025.02500},
  url       = {https://openaccess.thecvf.com/content/ICCV2025/html/Yang_DriveArena_A_Closed-loop_Generative_Simulation_Platform_for_Autonomous_Driving_ICCV_2025_paper.html}
}

@inproceedings{ge2025scenecrafter,
  title     = {Unraveling the Effects of Synthetic Data on End-to-End Autonomous Driving},
  author    = {Junhao Ge and Zuhong Liu and Longteng Fan and Yifan Jiang and Jiaqi Su and Yiming Li and Zhejun Zhang and Siheng Chen},
  booktitle = ICCV,
  pages     = {28859--28869},
  year      = {2025},
  doi       = {10.1109/ICCV51701.2025.02680},
  url       = {https://openaccess.thecvf.com/content/ICCV2025/html/Ge_Unraveling_the_Effects_of_Synthetic_Data_on_End-to-End_Autonomous_Driving_ICCV_2025_paper.html}
}

@inproceedings{li2026decouplegs,
  title     = {{DecoupleGS}: Interactive {3D Gaussian Splatting} for End-to-End Autonomous Driving Testing},
  author    = {Siying Li and Ying Ni and Jie Sun and Jian Sun and Haotian Shi},
  booktitle = {Computer Vision -- ECCV 2026},
  series    = {Lecture Notes in Computer Science},
  publisher = {Springer},
  volume    = {17044},
  pages     = {512--530},
  year      = {2026},
  doi       = {10.1007/978-3-032-37577-3_28},
  url       = {https://link.springer.com/chapter/10.1007/978-3-032-37577-3_28}
}

@inproceedings{leng2026dreamstream,
  title     = {{DreamStream}: Towards Policy-Oriented Generative Simulation for End-to-End Driving},
  author    = {Ziyang Leng and Sicheng Mo and Seth Z. Zhao and Haoyuan Cai and Yu Zeng and Rowan McAllister and Bolei Zhou},
  booktitle = {10th Annual Conference on Robot Learning (CoRL)},
  year      = {2026},
  url       = {https://vail.cs.ucla.edu/DreamStream/}
}

@inproceedings{wu2025difix3d,
  title     = {{DIFIX3D+}: Improving {3D} Reconstructions with Single-Step Diffusion Models},
  author    = {Jay Zhangjie Wu and Yuxuan Zhang and Haithem Turki and Xuanchi Ren and Jun Gao and Mike Zheng Shou and Sanja Fidler and Zan Gojcic and Huan Ling},
  booktitle = CVPR,
  pages     = {26024--26035},
  year      = {2025},
  doi       = {10.1109/CVPR52734.2025.02424},
  url       = {https://openaccess.thecvf.com/content/CVPR2025/html/Wu_DIFIX3D_Improving_3D_Reconstructions_with_Single-Step_Diffusion_Models_CVPR_2025_paper.html}
}

@inproceedings{hu2023uniad,
  title     = {Planning-Oriented Autonomous Driving},
  author    = {Hu, Yihan and Yang, Jiazhi and Chen, Li and Li, Keyu and Sima, Chonghao and Zhu, Xizhou and Chai, Siqi and Du, Senyao and Lin, Tianwei and Wang, Wenhai and Lu, Lewei and Jia, Xiaosong and Liu, Qiang and Dai, Jifeng and Qiao, Yu and Li, Hongyang},
  booktitle = CVPR,
  pages     = {17853--17862},
  year      = {2023},
  doi       = {10.1109/CVPR52729.2023.01712},
  url       = {https://openaccess.thecvf.com/content/CVPR2023/html/Hu_Planning-Oriented_Autonomous_Driving_CVPR_2023_paper.html}
}

@inproceedings{jiang2023vad,
  title     = {{VAD}: Vectorized Scene Representation for Efficient Autonomous Driving},
  author    = {Jiang, Bo and Chen, Shaoyu and Xu, Qing and Liao, Bencheng and Chen, Jiajie and Zhou, Helong and Zhang, Qian and Liu, Wenyu and Huang, Chang and Wang, Xinggang},
  booktitle = ICCV,
  pages     = {8340--8350},
  year      = {2023},
  doi       = {10.1109/ICCV51070.2023.00766},
  url       = {https://openaccess.thecvf.com/content/ICCV2023/html/Jiang_VAD_Vectorized_Scene_Representation_for_Efficient_Autonomous_Driving_ICCV_2023_paper.html}
}

@article{chitta2023transfuser,
  title     = {{TransFuser}: Imitation with Transformer-Based Sensor Fusion for Autonomous Driving},
  author    = {Chitta, Kashyap and Prakash, Aditya and Jaeger, Bernhard and Yu, Zehao and Renz, Katrin and Geiger, Andreas},
  journal   = PAMI,
  volume    = {45},
  number    = {11},
  pages     = {12878--12895},
  year      = {2023},
  doi       = {10.1109/TPAMI.2022.3200245},
  url       = {https://pubmed.ncbi.nlm.nih.gov/35984797/}
}

@inproceedings{liu2025gaussianfusion,
  title     = {{GaussianFusion}: {Gaussian}-Based Multi-Sensor Fusion for End-to-End Autonomous Driving},
  author    = {Liu, Shuai and Liang, Quanmin and Li, Zefeng and Li, Boyang and Huang, Kai},
  booktitle = NIPS,
  volume    = {38},
  pages     = {53594--53616},
  year      = {2025},
  doi       = {10.52202/085713-1603},
  url       = {https://proceedings.neurips.cc/paper_files/paper/2025/hash/44a76997ae5874529e49f2c2bd47b41c-Abstract-Conference.html}
}

@inproceedings{li2024egostatus,
  title     = {Is Ego Status All You Need for Open-Loop End-to-End Autonomous Driving?},
  author    = {Li, Zhiqi and Yu, Zhiding and Lan, Shiyi and Li, Jiahan and Kautz, Jan and Lu, Tong and Alvarez, Jose M.},
  booktitle = CVPR,
  pages     = {14864--14873},
  year      = {2024},
  doi       = {10.1109/CVPR52733.2024.01408},
  url       = {https://openaccess.thecvf.com/content/CVPR2024/html/Li_Is_Ego_Status_All_You_Need_for_Open-Loop_End-to-End_Autonomous_CVPR_2024_paper.html}
}

@inproceedings{ljungbergh2024neuroncap,
  title     = {{NeuroNCAP}: Photorealistic Closed-Loop Safety Testing for Autonomous Driving},
  author    = {Ljungbergh, William and Tonderski, Adam and Johnander, Joakim and Caesar, Holger and {\AA}str{\"o}m, Kalle and Felsberg, Michael and Petersson, Christoffer},
  booktitle = {Computer Vision -- ECCV 2024},
  series    = {Lecture Notes in Computer Science},
  publisher = {Springer},
  volume    = {15088},
  pages     = {161--177},
  year      = {2025},
  doi       = {10.1007/978-3-031-73404-5_10},
  url       = {https://link.springer.com/chapter/10.1007/978-3-031-73404-5_10}
}

@article{you2024bench2driver,
  title     = {{Bench2Drive-R}: Turning Real World Data into Reactive Closed-Loop Autonomous Driving Benchmark by Generative Model},
  author    = {You, Junqi and Jia, Xiaosong and Zhang, Zhiyuan and Zhu, Yutao and Yan, Junchi},
  journal   = {arXiv preprint arXiv:2412.09647},
  year      = {2024},
  doi       = {10.48550/arXiv.2412.09647},
  url       = {https://arxiv.org/abs/2412.09647}
}

@inproceedings{gao2025rad,
  title     = {{RAD}: Training an End-to-End Driving Policy via Large-Scale {3DGS}-based Reinforcement Learning},
  author    = {Gao, Hao and Chen, Shaoyu and Jiang, Bo and Liao, Bencheng and Shi, Yiang and Guo, Xiaoyang and Pu, Yuechuan and Yin, Haoran and Li, Xiangyu and Zhang, Xinbang and Zhang, Ying and Liu, Wenyu and Zhang, Qian and Wang, Xinggang},
  booktitle = NIPS,
  volume    = {38},
  pages     = {36642--36667},
  year      = {2025},
  doi       = {10.52202/085713-1092},
  url       = {https://proceedings.nips.cc/paper_files/paper/2025/hash/2ed3a566a0af6dcec424b988f1880ecc-Abstract-Conference.html}
}

@inproceedings{caesar2020nuscenes,
  title     = {{nuScenes}: A Multimodal Dataset for Autonomous Driving},
  author    = {Caesar, Holger and Bankiti, Varun and Lang, Alex H. and Vora, Sourabh and Liong, Venice Erin and Xu, Qiang and Krishnan, Anush and Pan, Yu and Baldan, Giancarlo and Beijbom, Oscar},
  booktitle = CVPR,
  pages     = {11621--11631},
  year      = {2020},
  doi       = {10.1109/CVPR42600.2020.01164},
  url       = {https://openaccess.thecvf.com/content_CVPR_2020/html/Caesar_nuScenes_A_Multimodal_Dataset_for_Autonomous_Driving_CVPR_2020_paper.html}
}

@article{caesar2021nuplan,
  title     = {{nuPlan}: A Closed-Loop {ML}-Based Planning Benchmark for Autonomous Vehicles},
  author    = {Caesar, Holger and Kabzan, Juraj and Tan, Kok Seang and Fong, Whye Kit and Wolff, Eric and Lang, Alex and Fletcher, Luke and Beijbom, Oscar and Omari, Sammy},
  journal   = {arXiv preprint arXiv:2106.11810},
  year      = {2021},
  doi       = {10.48550/arXiv.2106.11810},
  url       = {https://arxiv.org/abs/2106.11810}
}

@inproceedings{xiao2021pandaset,
  title     = {{PandaSet}: Advanced Sensor Suite Dataset for Autonomous Driving},
  author    = {Xiao, Pengchuan and Shao, Zhenlei and Hao, Steven and Zhang, Zishuo and Chai, Xiaolin and Jiao, Judy and Li, Zesong and Wu, Jian and Sun, Kai and Jiang, Kun and Wang, Yunlong and Yang, Diange},
  booktitle = {IEEE International Intelligent Transportation Systems Conference},
  pages     = {3095--3101},
  year      = {2021},
  doi       = {10.1109/ITSC48978.2021.9565009},
  url       = {https://ieeexplore.ieee.org/document/9565009/}
}

@article{helbing1995social,
  title     = {Social Force Model for Pedestrian Dynamics},
  author    = {Helbing, Dirk and Moln{\'a}r, P{\'e}ter},
  journal   = {Physical Review E},
  volume    = {51},
  number    = {5},
  pages     = {4282--4286},
  year      = {1995},
  doi       = {10.1103/PhysRevE.51.4282},
  url       = {https://journals.aps.org/pre/abstract/10.1103/PhysRevE.51.4282}
}

@inproceedings{wen2023limsim,
  title     = {{LimSim}: A Long-Term Interactive Multi-Scenario Traffic Simulator},
  author    = {Wen, Licheng and Fu, Daocheng and Mao, Song and Cai, Pinlong and Dou, Min and Li, Yikang and Qiao, Yu},
  booktitle = {IEEE International Conference on Intelligent Transportation Systems},
  pages     = {1255--1262},
  year      = {2023},
  doi       = {10.1109/ITSC57777.2023.10422219},
  url       = {https://doi.org/10.1109/ITSC57777.2023.10422219}
}

@inproceedings{yan2024streetgaussians,
  title     = {{Street Gaussians}: Modeling Dynamic Urban Scenes with {Gaussian} Splatting},
  author    = {Yan, Yunzhi and Lin, Haotong and Zhou, Chenxu and Wang, Weijie and Sun, Haiyang and Zhan, Kun and Lang, Xianpeng and Zhou, Xiaowei and Peng, Sida},
  booktitle = {Computer Vision -- ECCV 2024},
  series    = {Lecture Notes in Computer Science},
  publisher = {Springer},
  volume    = {15131},
  pages     = {156--173},
  year      = {2025},
  doi       = {10.1007/978-3-031-73464-9_10},
  url       = {https://link.springer.com/chapter/10.1007/978-3-031-73464-9_10}
}

@inproceedings{chen2025omnire,
  title     = {{OmniRe}: Omni Urban Scene Reconstruction},
  author    = {Chen, Ziyu and Yang, Jiawei and Huang, Jiahui and de Lutio, Riccardo and Martinez Esturo, Janick and Ivanovic, Boris and Litany, Or and Gojcic, Zan and Fidler, Sanja and Pavone, Marco and Song, Li and Wang, Yue},
  booktitle = ICLR,
  year      = {2025},
  url       = {https://proceedings.iclr.cc/paper_files/paper/2025/hash/d4938496368e1c5806138bea8837595c-Abstract-Conference.html}
}

@inproceedings{manivasagam2020lidarsim,
  title     = {{LiDARsim}: Realistic {LiDAR} Simulation by Leveraging the Real World},
  author    = {Manivasagam, Sivabalan and Wang, Shenlong and Wong, Kelvin and Zeng, Wenyuan and Sazanovich, Mikita and Tan, Shuhan and Yang, Bin and Ma, Wei-Chiu and Urtasun, Raquel},
  booktitle = CVPR,
  pages     = {11167--11176},
  year      = {2020},
  doi       = {10.1109/CVPR42600.2020.01118},
  url       = {https://openaccess.thecvf.com/content_CVPR_2020/html/Manivasagam_LiDARsim_Realistic_LiDAR_Simulation_by_Leveraging_the_Real_World_CVPR_2020_paper.html}
}

@inproceedings{zheng2024lidar4d,
  title     = {{LiDAR4D}: Dynamic Neural Fields for Novel Space-Time View {LiDAR} Synthesis},
  author    = {Zheng, Zehan and Lu, Fan and Xue, Weiyi and Chen, Guang and Jiang, Changjun},
  booktitle = CVPR,
  pages     = {5145--5154},
  year      = {2024},
  doi       = {10.1109/CVPR52733.2024.00492},
  url       = {https://openaccess.thecvf.com/content/CVPR2024/html/Zheng_LiDAR4D_Dynamic_Neural_Fields_for_Novel_Space-time_View_LiDAR_Synthesis_CVPR_2024_paper.html}
}

@inproceedings{zhao2025drivedreamer4d,
  title     = {{DriveDreamer4D}: World Models Are Effective Data Machines for {4D} Driving Scene Representation},
  author    = {Zhao, Guosheng and Ni, Chaojun and Wang, Xiaofeng and Zhu, Zheng and Zhang, Xueyang and Wang, Yida and Huang, Guan and Chen, Xinze and Wang, Boyuan and Zhang, Youyi and Mei, Wenjun and Wang, Xingang},
  booktitle = CVPR,
  pages     = {12015--12026},
  year      = {2025},
  doi       = {10.1109/CVPR52734.2025.01122},
  url       = {https://openaccess.thecvf.com/content/CVPR2025/html/Zhao_DriveDreamer4D_World_Models_Are_Effective_Data_Machines_for_4D_Driving_CVPR_2025_paper.html}
}

@inproceedings{ni2025recondreamer,
  title     = {{ReconDreamer}: Crafting World Models for Driving Scene Reconstruction via Online Restoration},
  author    = {Ni, Chaojun and Zhao, Guosheng and Wang, Xiaofeng and Zhu, Zheng and Qin, Wenkang and Huang, Guan and Liu, Chen and Chen, Yuyin and Wang, Yida and Zhang, Xueyang and Zhan, Yifei and Zhan, Kun and Jia, Peng and Lang, Xianpeng and Wang, Xingang and Mei, Wenjun},
  booktitle = CVPR,
  pages     = {1559--1569},
  year      = {2025},
  doi       = {10.1109/CVPR52734.2025.00153},
  url       = {https://openaccess.thecvf.com/content/CVPR2025/html/Ni_ReconDreamer_Crafting_World_Models_for_Driving_Scene_Reconstruction_via_Online_CVPR_2025_paper.html}
}

@inproceedings{yan2025streetcrafter,
  title     = {{StreetCrafter}: Street View Synthesis with Controllable Video Diffusion Models},
  author    = {Yan, Yunzhi and Xu, Zhen and Lin, Haotong and Jin, Haian and Guo, Haoyu and Wang, Yida and Zhan, Kun and Lang, Xianpeng and Bao, Hujun and Zhou, Xiaowei and Peng, Sida},
  booktitle = CVPR,
  pages     = {822--832},
  year      = {2025},
  doi       = {10.1109/CVPR52734.2025.00085},
  url       = {https://openaccess.thecvf.com/content/CVPR2025/html/Yan_StreetCrafter_Street_View_Synthesis_with_Controllable_Video_Diffusion_Models_CVPR_2025_paper.html}
}

@article{treiber2000idm,
  title     = {Congested Traffic States in Empirical Observations and Microscopic Simulations},
  author    = {Treiber, Martin and Hennecke, Ansgar and Helbing, Dirk},
  journal   = {Physical Review E},
  volume    = {62},
  number    = {2},
  pages     = {1805--1824},
  year      = {2000},
  doi       = {10.1103/PhysRevE.62.1805},
  url       = {https://journals.aps.org/pre/abstract/10.1103/PhysRevE.62.1805}
}

@inproceedings{lopez2018sumo,
  title     = {Microscopic Traffic Simulation using {SUMO}},
  author    = {Alvarez Lopez, Pablo and Behrisch, Michael and Bieker-Walz, Laura and Erdmann, Jakob and Fl{\"o}tter{\"o}d, Yun-Pang and Hilbrich, Robert and L{\"u}cken, Leonhard and Rummel, Johannes and Wagner, Peter and Wie{\ss}ner, Evamarie},
  booktitle = {IEEE International Conference on Intelligent Transportation Systems},
  pages     = {2575--2582},
  year      = {2018},
  doi       = {10.1109/ITSC.2018.8569938},
  url       = {https://doi.org/10.1109/ITSC.2018.8569938}
}

@inproceedings{vandenberg2011orca,
  title     = {Reciprocal n-Body Collision Avoidance},
  author    = {van den Berg, Jur and Guy, Stephen J. and Lin, Ming and Manocha, Dinesh},
  booktitle = {Robotics Research: The 14th International Symposium ISRR},
  series    = {Springer Tracts in Advanced Robotics},
  publisher = {Springer},
  volume    = {70},
  pages     = {3--19},
  year      = {2011},
  doi       = {10.1007/978-3-642-19457-3_1},
  url       = {https://link.springer.com/chapter/10.1007/978-3-642-19457-3_1}
}

@inproceedings{mahmood2019amass,
  title     = {{AMASS}: Archive of Motion Capture as Surface Shapes},
  author    = {Mahmood, Naureen and Ghorbani, Nima and Troje, Nikolaus F. and Pons-Moll, Gerard and Black, Michael J.},
  booktitle = ICCV,
  pages     = {5442--5451},
  year      = {2019},
  doi       = {10.1109/ICCV.2019.00554},
  url       = {https://openaccess.thecvf.com/content_ICCV_2019/html/Mahmood_AMASS_Archive_of_Motion_Capture_As_Surface_Shapes_ICCV_2019_paper.html}
}

@article{loper2015smpl,
  title     = {{SMPL}: A Skinned Multi-Person Linear Model},
  author    = {Loper, Matthew and Mahmood, Naureen and Romero, Javier and Pons-Moll, Gerard and Black, Michael J.},
  journal   = TOG,
  volume    = {34},
  number    = {6},
  pages     = {248:1--248:16},
  year      = {2015},
  doi       = {10.1145/2816795.2818013},
  url       = {https://doi.org/10.1145/2816795.2818013}
}

@article{holden2020learned,
  title     = {Learned Motion Matching},
  author    = {Holden, Daniel and Kanoun, Oussama and Perepichka, Maksym and Popa, Tiberiu},
  journal   = TOG,
  volume    = {39},
  number    = {4},
  pages     = {53:1--53:13},
  year      = {2020},
  doi       = {10.1145/3386569.3392440},
  url       = {https://doi.org/10.1145/3386569.3392440}
}

@article{chen2023pvg,
  title     = {Periodic Vibration {Gaussian}: Dynamic Urban Scene Reconstruction and Real-time Rendering},
  author    = {Chen, Yurui and Gu, Chun and Jiang, Junzhe and Zhu, Xiatian and Zhang, Li},
  journal   = {International Journal of Computer Vision},
  volume    = {134},
  year      = {2026},
  note      = {Article 83},
  doi       = {10.1007/s11263-026-02740-3},
  url       = {https://link.springer.com/article/10.1007/s11263-026-02740-3}
}

@inproceedings{chen2024lidargs,
  title     = {{LiDAR-GS}: Real-time {LiDAR} Re-Simulation using {Gaussian} Splatting},
  author    = {Chen, Qifeng and Du, Sicong and Tang, Tao and Xie, Rengan and Chen, Peng and Huo, Yuchi and Yang, Sheng},
  booktitle = {14th International Conference on Computational Visual Media (CVM 2026)},
  year      = {2026},
  note      = {Poster P31},
  url       = {https://iccvm.org/2026/files/papers/640.pdf}
}

@inproceedings{jiang2025gslidar,
  title     = {{GS-LiDAR}: Generating Realistic {LiDAR} Point Clouds with Panoramic {G}aussian Splatting},
  author    = {Jiang, Junzhe and Gu, Chun and Chen, Yurui and Zhang, Li},
  booktitle = {International Conference on Learning Representations (ICLR)},
  year      = {2025},
  url       = {https://proceedings.iclr.cc/paper_files/paper/2025/hash/8e86fd925dda476f574689f4ce748193-Abstract-Conference.html}
}

@inproceedings{heusel2017gans,
  title     = {{GANs} Trained by a Two Time-Scale Update Rule Converge to a Local {N}ash Equilibrium},
  author    = {Heusel, Martin and Ramsauer, Hubert and Unterthiner, Thomas and Nessler, Bernhard and Hochreiter, Sepp},
  booktitle = NIPS,
  publisher = {Curran Associates, Inc.},
  volume    = {30},
  pages     = {6626--6637},
  year      = {2017},
  url       = {https://proceedings.neurips.cc/paper/2017/hash/8a1d694707eb0fefe65871369074926d-Abstract.html}
}

@inproceedings{codevilla2018offline,
  title     = {On Offline Evaluation of Vision-Based Driving Models},
  author    = {Codevilla, Felipe and L{\'o}pez, Antonio M. and Koltun, Vladlen and Dosovitskiy, Alexey},
  booktitle = {Computer Vision -- ECCV 2018},
  series    = {Lecture Notes in Computer Science},
  publisher = {Springer},
  volume    = {11219},
  pages     = {246--262},
  year      = {2018},
  doi       = {10.1007/978-3-030-01267-0_15},
  url       = {https://link.springer.com/chapter/10.1007/978-3-030-01267-0_15}
}

@inproceedings{dauner2023parting,
  title     = {Parting with Misconceptions about Learning-based Vehicle Motion Planning},
  author    = {Dauner, Daniel and Hallgarten, Marcel and Geiger, Andreas and Chitta, Kashyap},
  booktitle = {Proceedings of the 7th Conference on Robot Learning},
  series    = {Proceedings of Machine Learning Research},
  publisher = {PMLR},
  volume    = {229},
  pages     = {1268--1281},
  year      = {2023},
  url       = {https://proceedings.mlr.press/v229/dauner23a.html}
}

@article{chen2024pgsr,
  title     = {{PGSR}: Planar-Based {Gaussian} Splatting for Efficient and High-Fidelity Surface Reconstruction},
  author    = {Chen, Danpeng and Li, Hai and Ye, Weicai and Wang, Yifan and Xie, Weijian and Zhai, Shangjin and Wang, Nan and Liu, Haomin and Bao, Hujun and Zhang, Guofeng},
  journal   = {IEEE Transactions on Visualization and Computer Graphics},
  volume    = {31},
  number    = {9},
  pages     = {6100--6111},
  year      = {2025},
  doi       = {10.1109/TVCG.2024.3494046},
  url       = {https://doi.org/10.1109/TVCG.2024.3494046}
}

@inproceedings{dauner2024navsim,
  title     = {{NAVSIM}: Data-Driven Non-Reactive Autonomous Vehicle Simulation and Benchmarking},
  author    = {Dauner, Daniel and Hallgarten, Marcel and Li, Tianyu and Weng, Xinshuo and Huang, Zhiyu and Yang, Zetong and Li, Hongyang and Gilitschenski, Igor and Ivanovic, Boris and Pavone, Marco and Geiger, Andreas and Chitta, Kashyap},
  booktitle = {Advances in Neural Information Processing Systems (Datasets and Benchmarks Track)},
  volume    = {37},
  pages     = {28706--28719},
  year      = {2024},
  doi       = {10.52202/079017-0902},
  url       = {https://proceedings.neurips.cc/paper_files/paper/2024/hash/32768f7faf1995026ef9821c696f3404-Abstract-Datasets_and_Benchmarks_Track.html}
}

@inproceedings{he2017maskrcnn,
  title     = {Mask {R-CNN}},
  author    = {He, Kaiming and Gkioxari, Georgia and Doll{\'a}r, Piotr and Girshick, Ross},
  booktitle = {Proceedings of the IEEE International Conference on Computer Vision (ICCV)},
  pages     = {2961--2969},
  year      = {2017},
  doi       = {10.1109/ICCV.2017.322},
  url       = {https://openaccess.thecvf.com/content_iccv_2017/html/He_Mask_R-CNN_ICCV_2017_paper.html}
}

@inproceedings{teed2020raft,
  title     = {{RAFT}: Recurrent All-Pairs Field Transforms for Optical Flow},
  author    = {Teed, Zachary and Deng, Jia},
  booktitle = {Computer Vision -- ECCV 2020},
  series    = {Lecture Notes in Computer Science},
  publisher = {Springer},
  volume    = {12347},
  pages     = {402--419},
  year      = {2020},
  doi       = {10.1007/978-3-030-58536-5_24},
  url       = {https://link.springer.com/chapter/10.1007/978-3-030-58536-5_24}
}

@inproceedings{du20243drealcar,
  title     = {{3DRealCar}: An In-the-wild {RGB-D} Car Dataset with 360-degree Views},
  author    = {Du, Xiaobiao and Wang, Yida and Sun, Haiyang and Wu, Zhuojie and Sheng, Hongwei and Wang, Shuyun and Ying, Jiaying and Lu, Ming and Zhu, Tianqing and Zhan, Kun and Yu, Xin},
  booktitle = ICCV,
  pages     = {26488--26498},
  year      = {2025},
  doi       = {10.1109/ICCV51701.2025.02458},
  url       = {https://openaccess.thecvf.com/content/ICCV2025/html/Du_3DRealCar_An_In-the-wild_RGB-D_Car_Dataset_with_360-degree_Views_ICCV_2025_paper.html}
}

@inproceedings{black2023bedlam,
  title     = {{BEDLAM}: A Synthetic Dataset of Bodies Exhibiting Detailed Lifelike Animated Motion},
  author    = {Black, Michael J. and Patel, Priyanka and Tesch, Joachim and Yang, Jinlong},
  booktitle = CVPR,
  pages     = {8726--8737},
  year      = {2023},
  doi       = {10.1109/CVPR52729.2023.00843},
  url       = {https://openaccess.thecvf.com/content/CVPR2023/html/Black_BEDLAM_A_Synthetic_Dataset_of_Bodies_Exhibiting_Detailed_Lifelike_Animated_CVPR_2023_paper.html}
}

@inproceedings{liu2025uco3d,
  title     = {{UnCommon} Objects in {3D}},
  author    = {Liu, Xingchen and Tayal, Piyush and Wang, Jianyuan and Zarzar, Jesus and Monnier, Tom and Tertikas, Konstantinos and Duan, Jiali and Toisoul, Antoine and Zhang, Jason Y. and Neverova, Natalia and Vedaldi, Andrea and Shapovalov, Roman and Novotny, David},
  booktitle = CVPR,
  pages     = {14102--14113},
  year      = {2025},
  doi       = {10.1109/CVPR52734.2025.01316},
  url       = {https://openaccess.thecvf.com/content/CVPR2025/html/Liu_UnCommon_Objects_in_3D_CVPR_2025_paper.html}
}
}
\clearpage
\setcounter{page}{1}
\maketitlesupplementary
\appendix
\renewcommand{\thesection}{\Alph{section}}
\renewcommand{\thetable}{S\arabic{table}}
\renewcommand{\thefigure}{S\arabic{figure}}
\renewcommand{\thealgorithm}{S\arabic{algorithm}}
\setcounter{table}{0}
\setcounter{figure}{0}
\setcounter{algorithm}{0}

This supplement gives the implementation details (\cref{sec:supp_impl}), including the system architecture, the planner interface and pseudocode of the traffic and road-user models, the evaluation protocol (\cref{sec:supp_protocol}) and the additional results (\cref{sec:supp_results}) that the main paper refers to, followed by closed-loop visualizations (\cref{sec:supp_vis}) and limitations (\cref{sec:supp_limits}).
Unless stated otherwise, all numbers come from the same runs as the main paper. Citations refer to the reference list of the main paper.

\section{Implementation details}
\label{sec:supp_impl}

\subsection{System architecture}
\label{sec:supp_arch}
\noindent\textbf{Processes.}
A closed-loop run involves three kinds of processes.
The \emph{simulator} process holds the reconstruction on the GPU and runs everything that defines the scene state: ego dynamics, traffic, pedestrians and riders, camera and LiDAR rendering, navigation, collision checks and metrics.
The \emph{restorer} (NVR) runs as a persistent daemon in its own Python environment.
Each \emph{planner} runs unmodified in its own process and environment, started by a wrapper script from the adapter file of \cref{sec:supp_transfer}; the simulator waits until the planner's log prints the ready message of the adapter, i.e., until the weights are loaded.
Separate processes let planners with incompatible dependencies (for example, the mmdetection3d versions of UniAD and VAD and the NAVSIM stack of TransFuser and GaussianFusion) be evaluated in the same simulator, and a planner crash cannot corrupt the simulator state.
\Cref{alg:supp_loop} lists one episode.

\begin{algorithm}[t]
  \caption{\textbf{Closed-loop episode.} One simulator process; the planner runs in its own process and the restorer in a daemon.}
  \label{alg:supp_loop}
  \footnotesize
  \begin{algorithmic}[1]
    \Require scene bundle $\mathcal{B}$, scenario $\mathcal{C}$, planner adapter $\pi$
    \State load reconstruction, ground model, HD map and route of $\mathcal{B}$; install traffic and VRU models of $\mathcal{C}$
    \State start the NVR daemon and the planner process; wait for the planner's ready message
    \State $\mathcal{S}_0 \gets$ logged ego state $+$ start perturbation of the seed; $t \gets 0$
    \While{no terminal event \textbf{and} $t < T_{\max}$}
      \State pose every Gaussian class at $\mathcal{S}_t$ \Comment{ego camera rig on ground and body attitude}
      \State $\mathbf{I}_t \gets \textsc{Render}(\mathcal{S}_t)$;\; $\mathbf{I}_t \gets \textsc{NVR}(\mathbf{I}_t)$
      \If{$\pi$ needs LiDAR} $\mathbf{P}_t \gets \textsc{LidarSweep}(\mathcal{S}_t)$ \EndIf
      \State $c_t \gets \textsc{Command}(\text{route}, \mathcal{S}_t)$ \Comment{\SI{3}{s} look-ahead, \SI{2}{m} threshold}
      \State send $(\mathbf{I}_t, \mathbf{P}_t, \text{ego state}, \text{calibration}, c_t)$ on the observation pipe
      \State receive waypoints $\mathbf{W}_t$ on the plan pipe; express them in the rear-axle frame
      \State $(a_t, \dot\delta_t) \gets \textsc{Track}(\mathbf{W}_t, v_t, \delta_t)$
      \State ego $\gets \textsc{Bicycle}(\text{ego}, a_t, \dot\delta_t, \Delta t)$;\; attitude $\gets \textsc{Suspension}(\Delta t)$
      \State vehicles $\gets \textsc{ReactiveTraffic}(t{+}\Delta t, \text{ego})$ \Comment{\cref{alg:supp_traffic}}
      \State pedestrians, riders $\gets$ \cref{alg:supp_ped,alg:supp_rider} in \SI{30}{Hz} substeps over $\Delta t$
      \State check swept-footprint, static and off-scene events; $t \gets t+\Delta t$
    \EndWhile
    \State send the end message to the planner; compute the episode metrics
  \end{algorithmic}
\end{algorithm}

\noindent\textbf{Order within a step.}
The sensors are rendered from the state $\mathcal{S}_t$, the planner acts on them, and only then do the ego and the other road users move to $\mathcal{S}_{t+1}$.
All road users therefore react to the ego's previous action, as in reality, and no model sees the planner's output before the ego executes it.
Vehicles see the state at the start of the step; pedestrians and riders see vehicle states delayed by their reaction time.

\noindent\textbf{Coordinate frames.}
All simulation state lives in a \emph{road frame} per scene: $x$ along the initial ego heading, $y$ to the left, $z$ up, with the origin at the anchor sensor of the first frame (nuScenes: front camera) projected onto the road.
Vehicle states, sensor mounts and planner outputs use the \emph{vehicle frame} with its origin at the center of the rear axle on the ground ($x$ forward, $y$ left), the nuScenes ego convention.
The renderer's \emph{scene frame} differs from the road frame by a fixed rigid transform per dataset, which only the rendering and scene-loading code uses.
The camera mounts are calibrated so that the ego placed at its logged first pose on the modeled ground reproduces the logged camera poses.

\noindent\textbf{Timing.}
The simulation step is $\Delta t{=}\SI{0.25}{s}$, and the planner is queried at every step (\SI{4}{Hz}) and replans from scratch.
Pedestrians and riders are integrated at \SI{30}{Hz}, the frame rate of the motion database.
Time-varying content of the reconstruction, such as the sky and recorded pedestrians, is taken from the nearest reconstruction frame; recorded vehicles in reactive mode are posed continuously along their logged paths.
The navigation command is updated before each observation: the ego is projected along the route by the distance it covers in \SI{3}{s} at its current speed, a lateral offset of more than \SI{2}{m} in the ego frame gives a turn command, and a turn command is held for at least \SI{1}{s}.
A stopped ego therefore receives \emph{straight}, as in the nuScenes labels the planners are trained on.

\subsection{Planner interface and data transfer}
\label{sec:supp_transfer}
\noindent\textbf{Adapters.}
A planner is registered by a short adapter file that gives its launch command, ready message, cameras, image color order, whether it needs LiDAR, and the time step, frame and axes of its waypoints (\cref{tab:supp_interface}).
AETHER converts the waypoints from the declared frame (for example the LiDAR frame with axes right and forward for UniAD and VAD) to the rear-axle frame using the same sensor mounts that it sends to the planner.
No planner code is changed; the adapters only translate between the message below and the planner's own data loader.

\begin{table*}[t]
  \centering
  \caption{\textbf{Planner interface.} Every planner receives rendered images at $800{\times}450$ (its adapter resizes them to the model's input), the camera intrinsics and extrinsics of the recording rig, the ego state and a navigation command, every simulation step. Waypoints are spaced \SI{0.5}{s}; their frame and axes are declared in the adapter and converted to the rear-axle frame by AETHER.}
  \label{tab:supp_interface}
  \footnotesize
  \setlength{\tabcolsep}{3.2pt}
  \begin{tabular}{@{}lllllll@{}}
    \toprule
    Planner & Training data & Cameras & Model input & LiDAR & Waypoints & Output frame \\
    \midrule
    UniAD~\cite{hu2023uniad}            & nuScenes  & all six                  & $1600{\times}900$ & --                 & 6 (\SI{3}{s}) & LiDAR, (right, forward) \\
    VAD~\cite{jiang2023vad}             & nuScenes  & all six                  & $1600{\times}900$ & --                 & 6 (\SI{3}{s}) & LiDAR, (right, forward) \\
    TransFuser (LTF)~\cite{dauner2024navsim} & OpenScene & front, front left/right & stitched front panorama & -- (latent BEV) & 8 (\SI{4}{s}) & rear axle \\
    GaussianFusion~\cite{liu2025gaussianfusion} & OpenScene & front, front left/right, back & per model & rendered sweep & 8 (\SI{4}{s}) & rear axle \\
    \bottomrule
  \end{tabular}
\end{table*}

\noindent\textbf{Transport.}
Simulator and planner exchange messages over two named pipes (FIFOs), one for observations and one for plans, in a working directory per run.
Each message is one pickled object (protocol 4) written by opening the pipe, writing and closing it, and read until end of file.
While waiting, each side polls the other process, so a crashed planner raises an error with the tail of its log instead of blocking the loop; startup and per-step timeouts are \SI{900}{s}.
An end message closes the planner, and a planner that does not exit is terminated with its process group.

\noindent\textbf{Observation message.}
The observation is a pair \texttt{(sensors, info)}.
\texttt{sensors} holds the restored images as \texttt{uint8} arrays per camera and, if requested, the LiDAR sweep as an $N{\times}4$ \texttt{float32} array (position and intensity) in the LiDAR frame of the rig.
\texttt{info} holds the time stamp, the navigation command, the ego position, heading, speed, acceleration, steering angle and rate, the ego box, and for every camera the intrinsics and the vehicle-to-camera and LiDAR-to-camera transforms of the recording rig.
It also carries the boxes of all road users and the road-frame state for analysis and privileged baselines; the planners evaluated here do not use them.
For six cameras this is about \SI{6.5}{MB} per step.
The plan message is either the waypoint array or a dictionary with waypoints and time step.

\noindent\textbf{Restorer daemon.}
The NVR daemon receives all cameras of a step in one message over its standard input, a JSON header followed by the raw RGB bytes, and returns the restored images the same way.
Images are processed as one batch at their native $800{\times}450$ (longer side below the \SI{1024}{px} limit) in bfloat16, with a fixed noise seed, so a run is reproducible.
Only the images go to the restorer; LiDAR, depth and the scene state are not changed by it.

\noindent\textbf{Scenario files.}
An experiment is a list of scenario files (\cref{fig:supp_yaml}), each pointing to a scene bundle.
A bundle contains the reconstruction checkpoint and its configuration, the rig (intrinsics and sensor mounts), the logged ego track and first-frame camera poses, the recorded actor trajectories, the HD map (lanes and drivable area) in the road frame, the LiDAR sidecar of \cref{sec:supp_lidar_impl}, and a cache of the ground model.
Replay and reactive variants of a scene differ only in their \texttt{traffic.model} entry, and crossing tests only in their \texttt{pedestrians} entry.

\begin{figure}[t]
  \centering
  \begin{minipage}{\columnwidth}
  \scriptsize
\begin{verbatim}
scene: bundles/nuscenes-0106   # bundle
sim: {dt: 0.25, max_steps: 48, seed: 0,
      tracker: pure_pursuit}
ego: {start: logged, speed: logged,
      start_noise: {lateral: 0.3, yaw: 0.03,
                    speed_mps: 1.0}}
refine: {enabled: true, model: nvr}
lidar: {required_source: v2}   # joint head
navigation: {mode: aligned}
traffic:
  assets: [3drealcar_..., 3drealcar_...]
  max_vehicles: 6
  model: {type: reactive, min_vehicles: 5,
          radius: 40.0, max_extra: 6}
pedestrians:
  model: aether
  assets: [bedlam_03, bedlam_05]
  vehicles: [uco3d_bicycle_02]
  gait_db: gaitdb_cmu_v4.npz
  walkers:
    - {id: ped, asset: 0, speed: 1.3, seed: 0,
       path: [[29.8,-7.1], [28.3,1.8],
              [27.9,3.8], [26.4,12.7]],
       kerb: [-4.5, 4.5]}
    - {id: cyc, asset: 1, vehicle: 0, speed: 4.5,
       seed: 1, path: [[47.6,41.8], [57.8,-27.5]],
       kerb: [-4.5, 4.5]}
termination:
  stop_on: [vehicle_collision, static_collision,
            off_scene]
  at_fault_only: true
\end{verbatim}
  \end{minipage}
  \caption{\textbf{Scenario file} of \cref{fig:supp_seq_ped} (abridged; the replay/reactive base file it extends is merged in). Paths are in the road frame (m); agent and output options are omitted.}
  \label{fig:supp_yaml}
\end{figure}

\subsection{Scene reconstruction}
\label{sec:supp_recon}
Each nuScenes~\cite{caesar2020nuscenes} log is reconstructed from its six cameras at $800{\times}450$ and its LiDAR with the OmniRe~\cite{chen2025omnire} decomposition into background, rigid vehicles and deformable humans, trained for 30k iterations.
The camera reconstruction is then continued jointly with LiDAR (\cref{sec:supp_lidar_impl}); all planners see the same resulting geometry.
For the LiDAR benchmark on PandaSet~\cite{xiao2021pandaset}, every second sweep from frame 2 to 78 (39 sweeps) is held out, and frame 0 is kept for appearance alignment.
Held-out LiDAR is removed before image projection, bounding-box computation and Gaussian initialization, so it never reaches the reconstruction.

\subsection{LiDAR rendering}
\label{sec:supp_lidar_impl}
\noindent\textbf{Rays.}
Each sensor uses its own elevation table with one row per physical laser, estimated from the recorded sweeps, and an azimuth resolution of \ang{0.2}; returns are rendered between \SI{0.5}{m} and \SI{120}{m}.
Gaussians are rasterized at their per-frame pose, and the plane offset is transported to the moving beam origin during the sweep.

\noindent\textbf{Range.}
\Cref{eq:range} is used where $|\bar{\mathbf{n}}^{\top}\mathbf{d}|>0.08$ and the planar range stays within 25\% of the blended center range; elsewhere, mostly at silhouettes, the blended center range is used.
A range-view convolutional residual (32 channels, zero-initialized) predicts a bounded multiplicative range correction.

\noindent\textbf{Intensity and ray drop.}
Each Gaussian carries a 12-dimensional feature.
$\Phi$ in \cref{eq:lidar_head} is an MLP with 64 hidden units that takes the composited feature, the incidence cosine, the range and an 8-dimensional laser embedding.
A learnable per-laser, per-azimuth ray-drop logit models sensor-fixed patterns such as self-occlusion, and the same convolutional residual refines intensity and drop.
The drop threshold is calibrated on training sweeps that are withheld from the LiDAR loss (every eighth one), never on held-out sweeps.

\noindent\textbf{Training.}
The joint phase optimizes the RGB, SSIM and perceptual (weight 0.2) losses of the camera reconstruction together with LiDAR terms scaled by 2: range (0.1), mean range (0.1), range tail (0.1), intensity (1.0), ray drop (0.2), line of sight (0.01) and normal consistency (0.01).
Geometry and appearance keep the learning rates of the camera reconstruction, and the LiDAR head uses $5{\times}10^{-3}$.
For the LiDAR benchmark the phase takes 4k iterations, about \SI{14}{min} on one RTX~5090; the closed-loop scenes use a shorter 2k-iteration phase.

\noindent\textbf{In closed loop.}
The jointly trained geometry, the per-Gaussian LiDAR features and the head are stored in a sidecar next to the reconstruction.
At every step the sweep is rendered from the same posed Gaussians as the cameras, at the ego's current LiDAR pose, so camera and LiDAR are consistent by construction, including moved and inserted actors.
Inserted vehicles, pedestrians and riders were never seen by a LiDAR; they receive the median LiDAR feature of the reconstructed foreground Gaussians.
Ray drop uses the threshold calibrated on training sweeps, and the points are returned in the rig's LiDAR frame.

\subsection{Novel-view restoration}
\label{sec:supp_nvr}
NVR keeps the architecture of the released Difix~\cite{wu2025difix3d}, a single-step diffusion model with a LoRA-adapted VAE, and is initialized from its weights.
Training pairs are renderings of under-converged reconstructions paired with the real images of the same views.
They come from twenty nuScenes scenes (2400 pairs over all six cameras), and another five scenes (630 pairs) are used only for model selection; the twenty-five scenes are excluded from every evaluation.
Fine-tuning uses Difix's own losses (L2, LPIPS and, after 100 steps, a VGG Gram loss, all with weight 1) at $448{\times}800$, timestep 199, batch size 1, learning rate $2{\times}10^{-6}$ and 200 steps.
On 30 validation pairs, LPIPS to the real image drops from 0.442 for the raw rendering to 0.349 with released Difix and 0.317 with NVR.
At test time, the six views of a step are restored as one batch in bfloat16.

\subsection{Fidelity flag}
\label{sec:supp_flag_impl}
The AP in \cref{eq:retention} is a two-class BEV detection AP (vehicles, including riders' two-wheelers, and pedestrians) with nuScenes center-distance matching at 0.5, 1, 2 and \SI{4}{m}, averaged over the thresholds and classes, within \SI{50}{m} of the ego.
The ground truth is the set of reconstructed actors in the ego frame, so it does not depend on the viewpoint, and it includes occluded actors, so absolute values are lower than official nuScenes mAP.
For the radius we compare each offset frame with the frame at the same step of the unshifted run (paired retention) and use $\tau{=}0.9$, at least 20 frames and 5 scenes per bin, and $\psi_{\max}{=}\ang{10}$.
UniAD is calibrated on closed-loop runs started 0 to \SI{3}{m} off the log on 20 scenes, giving $\delta^\star{=}\SI{3}{m}$ with NVR.
VAD's detection AP in these runs (0.03) is too low for a stable ratio, so its radius ($\SI{2}{m}$) is taken from the offset sweep of \cref{tab:supp_camera_full}.
TransFuser and GaussianFusion are not calibrated (\cref{sec:supp_planners}).
The flag is computed from the ego's lateral and heading deviation from the logged route, measured before each action.

\subsection{Ground and attitude}
\label{sec:supp_ground_impl}
The ground $z_0$ is a \SI{0.5}{m} height field fused from rendered depth, excluding the recording vehicle and low-opacity samples.
In \cref{eq:ground}, $\eta$ is 1 within \SI{2}{m} of the route and decays linearly to 0 at \SI{8}{m}, $g_\sigma$ has $\sigma{=}\SI{3}{m}$ of arc length, and the median of $\epsilon$ is removed so that constant offsets stay with the camera mount calibration.
The body plane is fitted over the wheelbase extended by \SI{1}{m} on the ground smoothed at \SI{3}{m}, and the suspension in \cref{eq:suspension} uses $\omega{=}\SI{9}{rad/s}$ (\SI{1.4}{Hz}).
Inserted and moved vehicles use the same footprint plane, including cross-fall.

\subsection{Reactive traffic}
\label{sec:supp_traffic_impl}
\Cref{alg:supp_traffic} gives one traffic step.

\noindent\textbf{Logged vehicles.}
In reactive mode, every vehicle of the log keeps its reconstructed appearance and follows its logged path with its logged speed profile.
Its path is the logged track of its box center, extended by \SI{40}{m} along its final heading so that it can keep driving after the log ends, and its reference arc length $s^{\mathrm{log}}(t)$ and speed $v^{\mathrm{log}}(t)$ are the projections of the logged positions onto this path, made monotone and smoothed over five frames.
A vehicle that moves less than \SI{1}{m} in the whole log is treated as parked and keeps its logged heading.
Without a leader, the vehicle tracks $s^{\mathrm{log}}(t)$ with a catch-up gain of $k{=}\SI{0.5}{s^{-1}}$: a vehicle that was held up returns to its logged timeline instead of staying behind it.

\noindent\textbf{Leader.}
A vehicle reacts when the ego or another vehicle is on its path within \SI{60}{m} ahead.
A candidate counts if its center projects onto the path ahead of the vehicle within half the sum of their widths plus \SI{0.4}{m} of it; the gap is the arc-length difference minus half the sum of their lengths, and the leader speed is the candidate's velocity projected on the vehicle's heading, so that a crossing car does not count as a fast leader.
Two logged vehicles that are both within \SI{1}{m} of their logged timeline ignore each other, because their interaction is already in the log and IDM would impose larger gaps than the recorded drivers kept.
The IDM~\cite{treiber2000idm} acceleration is
\begin{equation}
  a = a_{\max}\Big[1 - \Big(\frac{v}{v_0}\Big)^4 - \Big(\frac{s^*}{g}\Big)^2\Big],
  \label{eq:supp_idm}
\end{equation}
where $s^* = s_0 + \max\big(0,\ vT + v\,\Delta v/(2\sqrt{a_{\max} b})\big)$ is the desired gap,
$g$ is the gap and $\Delta v$ the approach rate, with time headway $T{=}\SI{1.2}{s}$, minimum gap $s_0{=}\SI{2.5}{m}$, $a_{\max}{=}\SI{2}{m/s^2}$ and comfortable deceleration $b{=}\SI{3}{m/s^2}$.
For logged vehicles it can only lower the acceleration requested by the log, and a gap below \SI{0.3}{m} triggers an emergency deceleration of \SI{9}{m/s^2}.
The ego is a leader like any other vehicle, but its own motion comes only from the planner.

\noindent\textbf{Added vehicles.}
Every \SI{2}{s}, if fewer than five moving vehicles are within \SI{40}{m} of the ego and fewer than six vehicles have been added, one IDM vehicle is spawned.
Candidate positions lie on HD-map lanes within \SI{9}{m} of the ego, 16 to \SI{40}{m} ahead of or 12 to \SI{30}{m} behind it along the lane, and their path continues through randomly chosen successor lanes.
A candidate is rejected within \SI{12}{m} of the ego or \SI{10}{m} of another vehicle, so vehicles do not appear in the ego's immediate view.
Desired speeds are drawn from \SIrange{6}{11}{m/s}; a vehicle on a lane in the ego's direction starts within \SI{1.5}{m/s} of the ego's speed, otherwise at its desired speed.
Added vehicles are rendered with 3DRealCar~\cite{du20243drealcar} assets placed on the ground plane of their footprint, and they are removed beyond \SI{90}{m} from the ego; logged vehicles are removed when they are more than \SI{60}{m} away and their path has ended or their log ended more than \SI{8}{s} earlier.
All random choices use the scene seed, so a run is reproducible.

\noindent\textbf{Rendering.}
Logged vehicles are rendered with their own reconstructed Gaussians at the simulated pose, which replaces the logged per-frame pose.
Before each render, the visibility of every logged vehicle is set to whether the traffic model currently drives it, so a vehicle appears only while it is simulated.

\noindent\textbf{Other modes.}
Recorded pedestrians are replayed as recorded in the closed-loop evaluation of \cref{sec:exp_loop}.
In log replay, all recorded actors follow their logged trajectories regardless of the ego.
The sparse-IDM variant in \cref{tab:ablation} removes the recorded vehicles and places three IDM vehicles with a desired speed of \SI{10}{m/s}.

\begin{algorithm}[t]
  \caption{\textbf{Reactive background traffic}, one step of length $\Delta t$ at time $t$. $s$ is the arc length of a vehicle along its own path, $k{=}\SI{0.5}{s^{-1}}$ the catch-up gain and $a_{\max}{=}\SI{2}{m/s^2}$.}
  \label{alg:supp_traffic}
  \footnotesize
  \begin{algorithmic}[1]
    \Require active vehicles $\mathcal{V}$, ego state, logged tracks $(s^{\mathrm{log}}(t), v^{\mathrm{log}}(t))$
    \State activate logged vehicles whose track starts at $t$
    \For{$i \in \mathcal{V}$} \Comment{all vehicles see the state at the start of the step}
      \State $L \gets \textsc{Leader}(i)$ \Comment{closest ego or vehicle on $i$'s path $\le\SI{60}{m}$ ahead}
      \If{$i$ is logged}
        \State $v^{\mathrm{ref}} \gets v^{\mathrm{log}}_i(t)$ (last logged speed after the log ends)
        \State $v^{\mathrm{tgt}} \gets \mathrm{clip}\big(v^{\mathrm{ref}} + k\,(s^{\mathrm{log}}_i(t) - s_i),\ 0,\ v^{\mathrm{ref}}{+}3\big)$
        \State $a_i \gets \mathrm{clip}\big(1.5\,(v^{\mathrm{tgt}} - v_i),\ -6,\ a_{\max}\big)$
        \If{$L \neq \emptyset$} $a_i \gets \min\big(a_i,\ \textsc{IDM}(v_i, \max(v^{\mathrm{tgt}},1), L)\big)$ \EndIf
      \Else
        \State $a_i \gets \textsc{IDM}(v_i, v_{0,i}, L)$
      \EndIf
      \If{gap to $L < \SI{0.3}{m}$} $a_i \gets -9$ \EndIf
      \State $v_i \gets \max(0, v_i + a_i \Delta t)$;\; $s_i \gets s_i + v_i \Delta t$;\; pose $\gets \textsc{Path}_i(s_i)$
      \State remove $i$ if it is far from the ego and its path or log has ended
    \EndFor
    \If{$\ge\SI{2}{s}$ since the last spawn \textbf{and} $<5$ moving vehicles within \SI{40}{m} \textbf{and} $<6$ added}
      \State spawn an IDM vehicle on a lane $16$--\SI{40}{m} ahead of or $12$--\SI{30}{m} behind the ego
    \EndIf
  \end{algorithmic}
\end{algorithm}

\subsection{Pedestrians and riders}
\label{sec:supp_vru_impl}
\noindent\textbf{Scenario input.}
A pedestrian or rider is specified by a path in the road frame, a preferred speed, an optional vehicle asset and a seed (\cref{fig:supp_yaml}).
The crossing interval $[s_{\mathrm{in}}, s_{\mathrm{out}}]$ is where the path enters and leaves the drivable area of the HD map, or is given as curb offsets.
Each agent owns a seeded random stream, so identical seeds and ego actions give identical runs, and a different ego action changes only what the agent perceives.

\noindent\textbf{Perception.}
Every substep, the agent stores the vehicles within its range (\SI{60}{m} for pedestrians, \SI{80}{m} for riders) and uses the snapshot from one reaction time earlier.
An inattentive agent refreshes its snapshot only at glances, $0.6$ to $1.4$ times a glance period apart.
The most critical vehicle is the one with the smallest time to arrival at the crossing among those whose heading line crosses the remaining crossing interval within \SI{1.5}{m}; its delayed state is extrapolated to the present at constant acceleration.
A vehicle is \emph{yielding} if it is slower than \SI{1}{m/s} within \SI{12}{m} of the conflict point, or if it brakes harder than \SI{0.8}{m/s^2} and would stop more than \SI{1}{m} before it.

\noindent\textbf{Behavior.}
\Cref{alg:supp_ped,alg:supp_rider} give the state machines.
\Cref{tab:supp_hazard} lists the hazard coefficients of \cref{eq:hazard}.
The pedestrian model is fitted by maximum likelihood on 7{,}147 decision samples with 1{,}138 road entries from nuScenes training scenes.
A decision sample is every \SI{2}{Hz} keyframe in the \SI{12}{s} before an entry while the pedestrian is within \SI{4}{m} of the drivable area.
The gap is the smallest time to arrival of any approaching vehicle at the entry point, capped at \SI{12}{s}; when no vehicle approaches, a constant replaces the gap term.
None of the 92 decision samples with a gap below \SI{2}{s} led to an entry, which we keep as a hard floor.
Riders have too few decisions in nuScenes and use the prior in \cref{tab:supp_hazard}.
Perception is delayed by \SI{0.6}{s} for pedestrians and \SI{0.7}{s} for riders, with ranges of 60 and \SI{80}{m}; delayed states are extrapolated at constant acceleration.
A crossing pedestrian retreats if threatened within \SI{1.2}{m} of the curb, and otherwise hurries (\SI{1.6}{\times} walking speed) or hesitates according to a \SI{1}{s} safety margin.

\begin{table}[t]
  \centering
  \caption{\textbf{Road-entry hazard coefficients} of \cref{eq:hazard} (per second). Waiting time is capped at \SI{10}{s}, curb distance at \SI{4}{m} and walking speed at \SI{2.5}{m/s}.}
  \label{tab:supp_hazard}
  \footnotesize
  \setlength{\tabcolsep}{4pt}
  \begin{tabular}{@{}lcc@{}}
    \toprule
    Term & Pedestrian (fitted) & Rider (prior) \\
    \midrule
    Intercept $\beta_0$          & $-2.089$ & $-2.6$ \\
    Gap $\beta_g$ (\si{s^{-1}})  & $0.197$ & $0.5$ \\
    No approaching vehicle       & $0.889$ & $3.4$ \\
    Yielding vehicle $\beta_y$   & $2.0^\dagger$ & $2.0^\dagger$ \\
    Waiting time $\beta_w$ (\si{s^{-1}}) & $0.093$ & $0.08$ \\
    Crosswalk $\beta_c$          & $0.621$ & -- \\
    Curb distance $\beta_d$ (\si{m^{-1}}) & $-4.474$ & -- \\
    Walking speed $\beta_v$ (\si{s/m}) & $2.024$ & -- \\
    \bottomrule
    \multicolumn{3}{@{}l@{}}{\scriptsize $^\dagger$Prior; yielding occurs in 1.2\% of the samples.}
  \end{tabular}
\end{table}

\begin{algorithm}[t]
  \caption{\textbf{Pedestrian behavior}, one substep $h$. The crossing occupies the arc interval $[s_{\mathrm{in}}, s_{\mathrm{out}}]$ of the pedestrian's path; $s$ is its arc length and $v$ its speed.}
  \label{alg:supp_ped}
  \footnotesize
  \begin{algorithmic}[1]
    \State $\mathcal{V}^{\mathrm{seen}} \gets$ vehicles within \SI{60}{m} as they were $\tau_r$ ago \Comment{refreshed only at glances if inattentive}
    \State $(\mathit{tta}, y, s_c) \gets \textsc{Threat}(\mathcal{V}^{\mathrm{seen}})$ \Comment{percept extrapolated to now}
    \Switch{state}
      \Case{\textsc{Walk}}
        \State walk at $v_p$ toward the curb
        \If{$s_{\mathrm{in}} - s < \SI{4}{m}$ \textbf{and} \Call{Accept}{}} \textsc{Cross} without stopping
        \ElsIf{$s_{\mathrm{in}} - s$ is within stopping distance} \textsc{Approach}
        \EndIf
      \EndCase
      \Case{\textsc{Approach}}
        \State decelerate at \SI{0.8}{m/s^2} to stop at the curb; \textsc{Wait} when stopped
      \EndCase
      \Case{\textsc{Wait}}
        \State $t_w \gets t_w + h$; look at the threat, or scan both sides
        \If{\Call{Accept}{}} \textsc{Start} \EndIf
      \EndCase
      \Case{\textsc{Start}}
        \State keep looking at the threat for $0.4$--\SI{1.2}{s}, then \textsc{Cross}
      \EndCase
      \Case{\textsc{Cross}, \textsc{Run}, \textsc{Hesitate}}
        \State $t_{\mathrm{me}} \gets (s_c - s + 1)/(1.08\,v_p)$;\; $t_{\mathrm{run}} \gets (s_c - s + 1.5)/(1.6\,v_p)$
        \If{$\mathit{tta} < \min(t_{\mathrm{me}} + 1, 5)$ \textbf{and not} $y$}
          \If{$s - s_{\mathrm{in}} < \SI{1.2}{m}$ \textbf{and} $\mathit{tta} < t_{\mathrm{run}} + 0.5$} \textsc{Retreat} to the curb
          \ElsIf{$\mathit{tta} > t_{\mathrm{run}} + 0.3$} \textsc{Run} at $1.6\,v_p$
          \Else{} \textsc{Hesitate}: stop \SI{1.6}{m} before the conflict point
          \EndIf
        \Else{} \textsc{Cross} at $1.08\,v_p$; \textsc{Done} once $s \ge s_{\mathrm{out}}$
        \EndIf
      \EndCase
      \Case{\textsc{Retreat}}
        \State walk back to \SI{0.6}{m} behind the curb, then \textsc{Wait}
      \EndCase
    \EndSwitch
    \State root: speed, heading and facing follow the command (\cref{sec:supp_vru_impl}); body: motion matching with contacts
    \Function{Accept}{}
      \If{$\mathit{tta} < \SI{2}{s}$ \textbf{and not} $y$} \Return false \EndIf
      \State $\lambda \gets$ \cref{eq:hazard} with $(g{=}\mathit{tta}, y, t_w, c, d, v)$;\; \Return $u < 1 - e^{-\lambda h}$, $u\sim\mathcal{U}(0,1)$
    \EndFunction
  \end{algorithmic}
\end{algorithm}

\begin{algorithm}[t]
  \caption{\textbf{Rider behavior}, one substep $h$, for bicycles, scooters and motorcycles. $s_{\mathrm{in}}$ is where the rider's path enters the conflicting road, $b_c{=}\SI{1.5}{m/s^2}$ the comfortable deceleration and $t_{\mathrm{me}}$ the rider's time to clear the conflict point.}
  \label{alg:supp_rider}
  \footnotesize
  \begin{algorithmic}[1]
    \State $(\mathit{tta}, y, s_c) \gets \textsc{Threat}(\text{delayed percept})$ \Comment{as in \cref{alg:supp_ped}, $\tau_r{=}\SI{0.7}{s}$, range \SI{80}{m}}
    \If{not committed}
      \If{\textsc{Ride} \textbf{and} $s_{\mathrm{in}} - s < 1.4\,(v^2/2b_c + 1) + 3$}
        \State \textbf{if} \Call{Accept}{} (rolling, over \SI{0.5}{s}): committed $\gets$ true \textbf{else} \textsc{Yield}
      \EndIf
      \State \textsc{Yield}: brake at $b_c$ to stop \SI{2.2}{m} before the road edge; \textsc{Stopped} (foot down) when $v < \SI{0.1}{m/s}$
      \State \textsc{Stopped}: $t_w \gets t_w + h$; \textbf{if} \Call{Accept}{}: look \SI{0.6}{s}, then \textsc{Ride}, committed $\gets$ true
    \ElsIf{$s_{\mathrm{in}} - 3 \le s < s_{\mathrm{out}}$ \textbf{and} $\mathit{tta} < \min(t_{\mathrm{me}} + 1.2, 4)$ \textbf{and not} $y$}
      \State \textbf{if} $v^2/(2 b_{\mathrm{hard}}) < s_c - s - 1$: \textsc{Brake} \Comment{$b_{\mathrm{hard}}{=}\SI{4.5}{m/s^2}$}
    \EndIf
    \State otherwise cruise at $v_r$ with acceleration $\le\SI{1}{m/s^2}$; pedal state $\in$ \{pedal, coast, brake, stopped\}
    \State bicycle root with lean $\arctan(v^2\kappa/g_0)$; crank from wheel speed and gear; rider pose from the rig
  \end{algorithmic}
\end{algorithm}

\noindent\textbf{Root.}
Speed follows its command with a time constant of \SI{1.3}{s} when speeding up and \SI{0.4}{s} when slowing down, bounded by \SI{1}{m/s^2}; the two start-up parameters are fitted on nuScenes training tracks by matching the distribution of start-up times.
Body facing is a separate, rate-limited and critically damped state that turns toward the commanded direction at up to \SI{45}{\degree/s} during gait initiation, so the torso turns before the path bends.
The preferred walking speed is \SI{1.3}{m/s}, and crossings are 8\% faster.

\noindent\textbf{Body.}
The motion database contains about 400k frames of AMASS~\cite{mahmood2019amass} CMU locomotion (walking at all speeds, standing, starts, stops, turns and side steps), from which crouching or kneeling frames are removed.
Each frame stores the pose in its facing frame, the skeleton and foot-contact labels.
The matching feature is the root trajectory \SIlist{0.33;0.67;1}{s} ahead (position and facing), the root velocity and both feet's positions and velocities.
The database is searched every \SI{0.33}{s} with a switching cost of 3, playback is time-warped by 0.7 to 1.4 to the simulated speed, and switches are inertialized with a half-life of \SI{0.1}{s}.
Planted feet are locked in the world and placed on the local ground, the legs are solved by two-bone inverse kinematics, and the pelvis drops only as much as the legs require.

\noindent\textbf{Riders.}
The bike rig (wheels, saddle, grips, crank or foot pegs) is estimated from the asset's points.
The bicycle model bounds speed, acceleration (\SI{1}{m/s^2}), braking (2.5 to \SI{4.5}{m/s^2}), steering angle and steering rate.
The crank follows wheel speed through a gear that keeps the cadence between 55 and \SI{95}{rpm}, with level pedals while coasting or braking.
The rider's pose is solved every frame from the rig, with one foot on the ground at a stop.

\subsection{Closed-loop simulation}
\label{sec:supp_sim}
\noindent\textbf{Ego vehicle.}
The ego is the vehicle that recorded the scene, so its box matches the camera rig: on nuScenes a Renault Zoe (length \SI{4.08}{m}, width \SI{1.73}{m}, wheelbase \SI{2.59}{m}, maximum steering angle \ang{35}, acceleration \SI{3}{m/s^2} and braking \SI{8}{m/s^2}).
It follows a kinematic bicycle model about the rear axle with acceleration and steering rate as inputs, integrated semi-implicitly.
The steering angle is additionally limited by a lateral-acceleration bound, $v^2\tan\delta/L \le a_{y,\max}$, because a kinematic model has no tire slip and could otherwise corner at several $g$.
The body attitude follows the ground through the suspension of \cref{eq:suspension}.

\noindent\textbf{Tracking.}
The tracker separates longitudinal and lateral control.
The longitudinal loop takes the plan's speed \SI{1}{s} ahead and commands the acceleration that reaches it in that time, within the vehicle limits.
The lateral loop is an arc-length-matched variant of pure pursuit: it chooses the constant steering angle whose circular arc best overlays the planned \emph{path} at the points the plan reaches every \SI{0.5}{s} over its horizon (up to \SI{3}{s}), weighting nearer points more, and the steering rate drives the wheel toward that angle within the rate limit.
Comparing positions at equal arc length rather than equal time means that a plan asking for a different speed never turns into a lateral error, and previewing the whole horizon lets the ego start turns that planners such as UniAD keep almost straight for the first second of every replan.

\noindent\textbf{Episodes.}
Episodes run for at most \SI{12}{s} (48 steps) on nuScenes.
Seed 0 starts from the logged state; seeds 1 and 2 add uniform perturbations of $\pm\SI{0.3}{m}$ lateral, $\pm\SI{0.03}{rad}$ yaw and $\pm\SI{1}{m/s}$ speed.
An episode ends at a vehicle or static collision, when the ego is more than \SI{5}{m} outside the reconstructed scene for four steps, or at the time limit.
A vehicle collision is a swept overlap of footprints between steps.
A static collision is more than 50 background Gaussians inside the ego box, counting only dense splats 0.3 to \SI{2.5}{m} above the ground outside the recorded ego path.
The same swept test covers pedestrians and riders, with a \SI{0.6}{m} footprint for a pedestrian and the bike's footprint for a rider.

\noindent\textbf{Runtime.}
On one RTX~5090, a step of UniAD with six $800{\times}450$ cameras takes \SI{1.97}{s} on average over five 48-step episodes with reactive traffic and inserted road users: \SI{0.32}{s} for rendering, \SI{0.93}{s} for NVR, \SI{0.65}{s} for the planner including data transfer, and \SI{0.08}{s} for ego, traffic and road-user updates and collision checks.
A \SI{12}{s} episode takes \SIrange{125}{132}{s} including scene loading; NVR is the largest share and can be disabled for runs that do not need it.
These runs also rendered and stored the LiDAR sweep and the visualization data of \cref{sec:supp_vis_loop} at every step, so the numbers are upper bounds.

\subsection{Planners}
\label{sec:supp_planners}
UniAD~\cite{hu2023uniad} and VAD~\cite{jiang2023vad} use their public nuScenes weights and receive the six cameras rendered at $800{\times}450$, upsampled to the $1600{\times}900$ they expect.
TransFuser is the Latent TransFuser (LTF) of NAVSIM~\cite{dauner2024navsim} with its public weights; its latent BEV replaces the LiDAR, so it uses only the three front cameras.
GaussianFusion~\cite{liu2025gaussianfusion} uses four cameras and the rendered LiDAR.
TransFuser and GaussianFusion are trained on nuPlan/OpenScene data and are therefore evaluated across datasets.
All planners output waypoints every \SI{0.5}{s}; \cref{tab:supp_interface} summarizes their inputs and outputs.
Our AP probe reads only the detections of UniAD and VAD, so the fidelity flag of TransFuser and GaussianFusion uses UniAD's radius and is marked as uncalibrated.

\section{Evaluation protocol}
\label{sec:supp_protocol}

\subsection{Scenes}
\label{sec:supp_scenes}
\Cref{tab:supp_scenes} lists the scenes of every experiment.
The closed-loop pool contains 58 reconstructed nuScenes scenes.
Two of them train NVR and one selects it; the remaining 55 form the held-out set of all closed-loop results.
Four TransFuser replay episodes failed for technical reasons, so TransFuser has 161 instead of 165 paired episodes.
A fixed subset of 20 held-out scenes is used for the more expensive experiments.

\begin{table*}[t]
  \centering
  \caption{\textbf{Scenes used in each experiment} (nuScenes scene numbers unless stated otherwise).}
  \label{tab:supp_scenes}
  \footnotesize
  \setlength{\tabcolsep}{4pt}
  \renewcommand{\arraystretch}{1.12}
  \begin{tabular}{@{}p{3.6cm}p{13.2cm}@{}}
    \toprule
    Experiment & Scenes \\
    \midrule
    Closed-loop pool (58) & 0002 0010 0026 0035 0043 0053 0056 0106 0155 0158 0163 0199 0210 0227 0240 0274 0331 0352 0354 0383 0391 0393 0402 0407 0410 0414 0417 0419 0420 0426 0429 0436 0511 0517 0523 0538 0585 0586 0599 0643 0651 0654 0679 0687 0711 0712 0790 0795 0853 0892 0960 0998 1019 1048 1068 1076 1084 1092 \\
    Held-out set (55) & closed-loop pool without the three NVR scenes (\cref{fig:closedloop}, \cref{tab:trust,tab:supp_paired}) \\
    Subset (20) & 0002 0035 0056 0158 0210 0274 0354 0393 0410 0419 0429 0517 0585 0643 0679 0712 0853 0998 1068 1092 (\cref{fig:sensor_quant}b--d, \cref{tab:ablation,tab:supp_open_loop}) \\
    Camera comparison (17) & 0013 0038 0041 0051 0062 0064 0071 0138 0166 0167 0254 0411 0418 0528 0661 0920 0930 (HUGSIM releases; \cref{tab:camera,tab:supp_camera_full}) \\
    LiDAR (PandaSet) & sequences 001, 011, 028 (\cref{tab:lidar}); 053, 106, 120 never used for design choices (\cref{tab:supp_lidar6}) \\
    Ground height (10) & 0002 0035 0056 0158 0210 0274 0391 0407 0585 0643 (\cref{fig:actors}a) \\
    Camera pose (8) & 0002 0035 0056 0158 0210 0274; PandaSet 002, 004 (\cref{sec:exp_ground}, \cref{tab:supp_pose}) \\
    Crossing tests & candidates 0043 0056 0163 0210 0391 0407 0585 0892; per planner, those it drives safely without crossers (\cref{fig:crossing}) \\
    Pedestrian tracks & 1{,}083 nuScenes validation tracks of at least \SI{4}{s} (\cref{tab:vru}) \\
    \bottomrule
  \end{tabular}
\end{table*}

\subsection{Metrics}
\label{sec:supp_metrics}
\noindent\textbf{Camera.}
In the camera comparison, every method renders the logged camera poses of 30 steps per scene (\SI{0.5}{s} apart) at lateral offsets of 0 to \SI{3}{m}.
FID~\cite{heusel2017gans} and KID are computed with torch-fidelity on the 3{,}060 front-and-surround images per offset against the 3{,}060 real images at the same steps.
Detection uses the two-class BEV AP of \cref{sec:supp_flag_impl} on the 510 frames, and plan consistency is the ADE and FDE between the plans from rendered and real images at the same step.
The restorer ablation (\cref{fig:sensor_quant}b) instead bins the images that UniAD actually received in closed loop by their lateral offset, with at most 1{,}000 images per bin.

\noindent\textbf{LiDAR.}
All methods are evaluated on the same 39 held-out sweeps, the same rays, the same valid-ray mask and returns between 0.5 and \SI{120}{m}.
CD is the sum of the mean nearest-neighbor distances in both directions, F is the F-score at \SI{5}{cm}, RMSE and MedSE are the root-mean-square and median squared range errors on real returns, Int.\ is the intensity RMSE on real returns, and Drop is the ray-drop accuracy over all valid rays.

\noindent\textbf{Closed loop.}
The score follows the PDM score of NAVSIM~\cite{dauner2024navsim}: $\text{NC}\cdot\text{DAC}\cdot(5\,\text{TTC}+2\,\text{C}+5\,\text{EP})/12$.
NC is 0 after an at-fault collision and 0.5 after a static collision; DAC is 0 if a corner leaves the drivable area by more than \SI{0.3}{m}; TTC is 0 if a constant-velocity projection predicts contact within \SI{0.9}{s}; C checks the nuPlan comfort bounds; and EP is the route progress relative to the logged ego over the same time.
A collision is at fault unless the ego is stationary (below \SI{0.1}{m/s}) or is hit from behind by a faster vehicle.
Confidence intervals bootstrap scenes and keep all seeds and both traffic modes of a scene together.

\noindent\textbf{Ground and pose.}
Ground height is compared with the accumulated LiDAR ground of the log within a radius of the route, split into the driven corridor and the rest.
Camera pose is evaluated by placing the ego at every logged position and heading and letting each ground model supply height, pitch and roll.
The resulting cameras are compared with the logged cameras by rotation and position error and by the frame-to-frame rotation error.

\noindent\textbf{Pedestrian and rider motion.}
Root models re-simulate each held-out track from its path, initial state, cruise speed, onset time and end point.
They are compared with the real keyframes by ADE, body-yaw error and the 1-D Wasserstein distances of acceleration and yaw rate.
Body metrics are computed on the posed SMPL mesh against the simulator's terrain at \SI{20}{Hz}:
\begin{itemize}
  \item skate: horizontal speed of the support foot;
  \item penetration: depth of the lowest sole vertex below the ground;
  \item jerk: RMS vertical jerk of the pelvis;
  \item jump: 99th percentile of per-frame joint-angle change.
\end{itemize}
Riders are tested in four scenarios (constant speed, S-curve, stop and restart, 5\% slope) with the same kinematic bicycle root and speed commands for every method.
Contact averages the distances of hands, feet and pelvis to the grips, pedals and saddle.
Pedal is the share of frames with a pedaling command, speed above \SI{1}{m/s} and no braking in which the crank turns; Brake is the share of braking frames in which it still turns; Stop is the height of the support foot above the ground at a stop.
Interaction tests place 8 crossings (pedestrians 30, 45 and \SI{60}{m} ahead from both sides; riders 40 and \SI{60}{m}) against five ego interventions (constant 8 or \SI{12}{m/s}, early yield, late brake, acceleration) and four negative controls (no ego, ego on a parallel road, ego driving away, ego parked), with 10 seeds per cell.

\subsection{Baselines}
\label{sec:supp_baselines}
\noindent\textbf{Camera.}
HUGSIM~\cite{zhou2026hugsim} uses the authors' released reconstructions.
Street Gaussians~\cite{yan2024streetgaussians} and PVG~\cite{chen2023pvg} are trained in DriveStudio with exactly the data, cameras, resolution and 30k-iteration budget of our OmniRe backbone.
NeuRAD~\cite{tonderski2024neurad}, which is the renderer of NeuroNCAP~\cite{ljungbergh2024neuroncap}, and the UniSim~\cite{yang2023unisim} re-implementation are trained in neurad-studio with their default settings on all six cameras and the top LiDAR.
OmniRe is our reconstruction rendered without NVR.
NeuRAD and UniSim could not be trained on two of the 19 HUGSIM scenes, whose raw LiDAR was unavailable, so all methods are compared on the remaining 17.

\noindent\textbf{LiDAR.}
SplatAD~\cite{hess2025splatad} is trained with its default 30k iterations on explicit training frames and exported on our rays.
NeuRAD~\cite{tonderski2024neurad} uses neurad-studio defaults; GS-LiDAR~\cite{jiang2025gslidar} is trained for 30k iterations followed by its refinement stage, and LiDAR-GS~\cite{chen2024lidargs} for 10k iterations with a maximum depth of \SI{120}{m}.
All baselines are initialized from training sweeps only.
LiDAR-GS diverged numerically on one sequence and is reported on the other two.

\noindent\textbf{Pedestrians and riders.}
\begin{itemize}
  \item Scripted: AETHER's previous walker, moving at constant speed along its path and triggered when the ego's time to collision drops below \SI{4}{s}.
  \item Social force~\cite{helbing1995social}: relaxation toward the path direction with a time constant of \SI{0.5}{s}.
  \item SF\,+\,ORCA~\cite{vandenberg2011orca}: social force with sampling-based velocity-obstacle avoidance.
  \item Clip playback: an AMASS walking clip indexed by travelled distance, with the lowest vertex placed on the ground in every frame.
  \item Motion matching: our body layer without contact enforcement.
  \item Rigid rider: one riding pose fixed to the bike, as in simulators that insert assets.
  \item Procedural IK: hands, feet and pelvis solved onto grips, pedals and saddle, with the crank turned by the distance traveled and no riding state.
\end{itemize}

\section{Additional results}
\label{sec:supp_results}

\subsection{Camera}
\label{sec:supp_camera}
\Cref{tab:supp_camera_full} extends \cref{tab:camera} to all four offsets.
VAD keeps 96\% of its on-log mAP at \SI{2}{m} and 88\% at \SI{3}{m}, which gives its fidelity radius $\delta^\star{=}\SI{2}{m}$ for $\tau{=}0.9$.
\Cref{tab:supp_camera_kid} adds KID, per-class AP and FDE.
The baselines lose most of their pedestrian AP, whereas AETHER keeps VAD's pedestrian AP above 0.1 up to \SI{3}{m}.

\noindent\textbf{Logged views.}
FID and KID measure how realistic the images are, but on the logged views a pixel-aligned reference exists, so \cref{tab:supp_onlog} also reports the usual reconstruction metrics.
Every method renders the exact logged camera poses and intrinsics of the dataset at $800{\times}450$, and the metrics are averaged over images.
On the 17 scenes, AETHER has the best LPIPS by a wide margin (.232 against .319 for the next method), the second-best SSIM, and a PSNR within \SI{0.06}{dB} of the best, so the restorer that makes off-log views realistic does not cost fidelity on the log.
The pixel metrics separate the strong reconstructions only slightly (the top four PSNRs lie within \SI{0.21}{dB}), whereas LPIPS, which follows perceived quality, does.
On the closed-loop pool, NVR lowers PSNR and SSIM of AETHER's own reconstruction, as expected for a generative restorer that synthesizes plausible high-frequency detail rather than the exact pixels, but it lowers LPIPS by 32\%.

\begin{table}[t]
  \centering
  \caption{\textbf{Reconstruction quality on the logged views.} PSNR, SSIM and LPIPS (AlexNet) against the real images at the logged camera poses, all six cameras. Top: the 17 scenes and 3{,}060 images of the camera comparison. \rankkey\ Bottom: AETHER's 58-scene closed-loop pool, eight frames per scene spread evenly over the log (2{,}784 images), with and without NVR; best in bold.}
  \label{tab:supp_onlog}
  \footnotesize
  \setlength{\tabcolsep}{4pt}
  \begin{tabular}{@{}lccc@{}}
    \toprule
    Method & PSNR $\uparrow$ & SSIM $\uparrow$ & LPIPS $\downarrow$ \\
    \midrule
    \multicolumn{4}{@{}l}{\textit{Camera comparison (17 scenes)}} \\
    UniSim$^\dagger$~\cite{yang2023unisim}   & 18.09 & .533 & .526 \\
    NeuroNCAP~\cite{ljungbergh2024neuroncap} & 21.55 & .655 & .401 \\
    HUGSIM~\cite{zhou2026hugsim}             & 24.22 & \first{.755} & \second{.319} \\
    PVG~\cite{chen2023pvg}                   & \first{24.43} & .725 & .430 \\
    StreetGS~\cite{yan2024streetgaussians}   & \third{24.29} & \third{.733} & \third{.340} \\
    OmniRe~\cite{chen2025omnire}             & 23.21 & .674 & .356 \\
    \ours                                    & \second{24.37} & \second{.746} & \first{.232} \\
    \midrule
    \multicolumn{4}{@{}l}{\textit{Closed-loop pool (58 scenes)}} \\
    OmniRe (no restorer)                     & \textbf{22.89} & \textbf{.691} & .441 \\
    \ours                                    & 22.09 & .630 & \textbf{.301} \\
    \bottomrule
  \end{tabular}\\[2pt]
  {\scriptsize $^\dagger$Three of the 17 scenes render almost blank; 20.53 / .647 / .425 without them.}
\end{table}

\begin{table*}[t]
  \centering
  \caption{\textbf{Camera fidelity at every offset} on 17 HUGSIM scenes, at 0 to \SI{3}{m} lateral offset. FID against real images, detection mAP of each planner's own head, and ADE (m) between plans from rendered and real images on the log. $^\dagger$Re-implementation. \rankkey}
  \label{tab:supp_camera_full}
  \footnotesize
  \setlength{\tabcolsep}{4.1pt}
  \renewcommand{\arraystretch}{1.08}
  \begin{tabular}{@{}l cccc cccc cccc cc@{}}
    \toprule
    & \multicolumn{4}{c}{FID $\downarrow$} & \multicolumn{4}{c}{UniAD mAP $\uparrow$} & \multicolumn{4}{c}{VAD mAP $\uparrow$} & \multicolumn{2}{c}{ADE $\downarrow$} \\
    \cmidrule(lr){2-5}\cmidrule(lr){6-9}\cmidrule(lr){10-13}\cmidrule(l){14-15}
    Method & 0\,m & 1\,m & 2\,m & 3\,m & 0\,m & 1\,m & 2\,m & 3\,m & 0\,m & 1\,m & 2\,m & 3\,m & UniAD & VAD \\
    \midrule
    \reff{Real images} & \reff{--} & \reff{--} & \reff{--} & \reff{--} & \reff{.380} & \reff{--} & \reff{--} & \reff{--} & \reff{.202} & \reff{--} & \reff{--} & \reff{--} & \reff{--} & \reff{--} \\
    \midrule
    UniSim$^\dagger$~\cite{yang2023unisim} & 136.4 & 144.3 & 152.6 & 162.1 & .107 & .100 & .096 & .091 & .043 & .039 & .035 & .034 & 1.451 & .928 \\
    NeuroNCAP~\cite{ljungbergh2024neuroncap} & \third{40.4} & \third{45.2} & \third{52.4} & \second{60.0} & .125 & .121 & .113 & .104 & .056 & .049 & .045 & .043 & .803 & .886 \\
    HUGSIM~\cite{zhou2026hugsim} & \second{37.4} & \second{42.3} & \second{50.0} & \third{60.7} & .241 & .230 & .215 & .197 & \second{.111} & .093 & .079 & .068 & \third{.523} & \second{.510} \\
    \midrule
    PVG~\cite{chen2023pvg} & 56.8 & 60.2 & 67.2 & 76.7 & .232 & .201 & .183 & .161 & .064 & .054 & .048 & .044 & .609 & .855 \\
    StreetGS~\cite{yan2024streetgaussians} & 42.2 & 46.0 & 53.4 & 63.0 & \third{.253} & \third{.251} & \third{.228} & \third{.215} & \third{.107} & \second{.101} & \second{.093} & \third{.080} & \second{.464} & \third{.699} \\
    OmniRe~\cite{chen2025omnire} & 47.2 & 50.6 & 56.7 & 65.3 & \second{.263} & \second{.252} & \second{.240} & \second{.235} & \third{.107} & \third{.099} & \third{.089} & \second{.081} & .563 & .845 \\
    \midrule
    \ours & \first{16.9} & \first{18.9} & \first{21.8} & \first{25.5} & \first{.265} & \first{.272} & \first{.255} & \first{.242} & \first{.170} & \first{.171} & \first{.163} & \first{.149} & \first{.411} & \first{.383} \\
    \bottomrule
  \end{tabular}
\end{table*}

\begin{table*}[t]
  \centering
  \caption{\textbf{Camera comparison, further metrics.} KID ($\times 10^3$) by lateral offset, per-class detection AP on the log and \SI{3}{m} off it, and FDE (m) between plans from rendered and real images on the log.}
  \label{tab:supp_camera_kid}
  \footnotesize
  \setlength{\tabcolsep}{3.4pt}
  \renewcommand{\arraystretch}{1.08}
  \begin{tabular}{@{}l cccc cccc cccc cc@{}}
    \toprule
    & \multicolumn{4}{c}{KID $\downarrow$} & \multicolumn{4}{c}{UniAD AP $\uparrow$ (veh.\ / ped.)} & \multicolumn{4}{c}{VAD AP $\uparrow$ (veh.\ / ped.)} & \multicolumn{2}{c}{FDE $\downarrow$} \\
    \cmidrule(lr){2-5}\cmidrule(lr){6-9}\cmidrule(lr){10-13}\cmidrule(l){14-15}
    Method & 0\,m & 1\,m & 2\,m & 3\,m & \multicolumn{2}{c}{0\,m} & \multicolumn{2}{c}{3\,m} & \multicolumn{2}{c}{0\,m} & \multicolumn{2}{c}{3\,m} & UniAD & VAD \\
    \midrule
    \reff{Real images} & \reff{--} & \reff{--} & \reff{--} & \reff{--} & \reff{.415} & \reff{.345} & \reff{--} & \reff{--} & \reff{.206} & \reff{.198} & \reff{--} & \reff{--} & \reff{--} & \reff{--} \\
    \midrule
    UniSim$^\dagger$~\cite{yang2023unisim}   & 67.4 & 75.0 & 82.1 & 90.1 & .207 & .008 & .179 & .003 & .085 & .002 & .068 & .000 & 2.78 & 1.68 \\
    NeuroNCAP~\cite{ljungbergh2024neuroncap} & 17.8 & 21.5 & 26.6 & 31.7 & .238 & .012 & .199 & .010 & .110 & .003 & .086 & .000 & 1.61 & 1.60 \\
    HUGSIM~\cite{zhou2026hugsim}             & 20.6 & 23.4 & 27.8 & 33.9 & .305 & \textbf{.176} & .287 & .106 & .138 & .083 & .102 & .035 & 1.03 & .91 \\
    PVG~\cite{chen2023pvg}                   & 34.6 & 36.6 & 42.0 & 49.8 & .337 & .126 & .276 & .047 & .124 & .004 & .089 & .000 & 1.16 & 1.53 \\
    StreetGS~\cite{yan2024streetgaussians}   & 23.0 & 25.9 & 31.9 & 39.9 & .371 & .136 & .339 & .090 & .176 & .037 & .139 & .022 & .90 & 1.21 \\
    OmniRe~\cite{chen2025omnire}             & 26.1 & 28.5 & 33.5 & 40.1 & .389 & .137 & \textbf{.374} & .096 & .169 & .045 & .147 & .015 & 1.11 & 1.46 \\
    \midrule
    \ours & \textbf{2.6} & \textbf{3.3} & \textbf{4.4} & \textbf{5.5} & \textbf{.392} & .137 & .370 & \textbf{.114} & \textbf{.218} & \textbf{.121} & \textbf{.191} & \textbf{.106} & \textbf{.84} & \textbf{.70} \\
    \bottomrule
  \end{tabular}
\end{table*}

\noindent\textbf{Restoration on the log.}
\Cref{tab:supp_restorer} gives the values of \cref{fig:sensor_quant}b.
Open loop on the logged poses of the 20-scene subset (\cref{tab:supp_open_loop}), NVR raises the detection mAP of both planners and brings the plans of UniAD, VAD and TransFuser closer to those on real images.
It also keeps the rate at which planned trajectories hit ground-truth boxes close to that on real images.

\begin{table}[t]
  \centering
  \caption{\textbf{FID of the restorers} by observed lateral offset in closed loop. $^\ddagger$Two scenes only.}
  \label{tab:supp_restorer}
  \footnotesize
  \setlength{\tabcolsep}{3pt}
  \begin{tabular}{@{}lccccc@{}}
    \toprule
    Restorer & 0--0.5\,m & 0.5--1\,m & 1--2\,m & 2--3\,m & $>$3\,m \\
    \midrule
    None           & 85.4 & 88.9 & 99.6 & 112.3 & 122.7 \\
    Released Difix & 48.3 & 51.1 & 51.3 & 69.0 & 129.8$^\ddagger$ \\
    NVR            & \textbf{46.4} & \textbf{46.9} & \textbf{48.8} & \textbf{55.1} & \textbf{79.2} \\
    \bottomrule
  \end{tabular}
\end{table}

\begin{table}[t]
  \centering
  \caption{\textbf{Open-loop replay on the logged poses} (20 scenes). Detection mAP, ADE (m) to the plan on real images, and rate of planned waypoints that hit a ground-truth box.}
  \label{tab:supp_open_loop}
  \footnotesize
  \setlength{\tabcolsep}{2.6pt}
  \begin{tabular}{@{}lccccccc@{}}
    \toprule
    & \multicolumn{3}{c}{mAP $\uparrow$} & \multicolumn{2}{c}{ADE $\downarrow$} & \multicolumn{2}{c}{Planned coll.} \\
    \cmidrule(lr){2-4}\cmidrule(lr){5-6}\cmidrule(l){7-8}
    Planner & Real & Raw & NVR & Raw & NVR & Real & NVR \\
    \midrule
    UniAD          & .415 & .199 & .221 & .71 & .46 & .078 & .085 \\
    VAD            & .183 & .075 & .129 & .88 & .40 & .097 & .089 \\
    TransFuser     & -- & -- & -- & .35 & .25 & .080 & .079 \\
    GaussianFusion & -- & -- & -- & .76 & .76 & .116 & .124 \\
    \bottomrule
  \end{tabular}
\end{table}

\noindent\textbf{Closed loop under start offsets.}
Started 0, 1, 2 and \SI{3}{m} off the log, UniAD scores 0.515, 0.432, 0.289 and 0.253 with NVR, 0.465, 0.353, 0.239 and 0.230 without a restorer, and 0.514 (\SI{0}{m}) and 0.296 (\SI{2}{m}) with released Difix.
Without NVR, VAD barely drives (route completion 0.10 vs.\ 0.42).

\subsection{Restoration and cross-modal alignment}
\label{sec:supp_align}
NVR changes the image but not the Gaussians, so the LiDAR is unaffected by construction; what can break the alignment between the two sensors is a restored image that moves or removes content.
We therefore compare the raw rendering and its restored version at the same pose on the 17 scenes of the camera comparison, using every third step and all six cameras.
An off-the-shelf Mask R-CNN~\cite{he2017maskrcnn} detects vehicles, pedestrians and riders on the raw rendering, and RAFT~\cite{teed2020raft} estimates the flow from the raw to the restored image.
Inside the actor masks, the median flow is \SI{0.05}{px} at every offset (0--\SI{3}{m}), the same as on the background (\SI{0.07}{px}).
The 95th-percentile flow of an actor stays below \SI{0.5}{px} for 99\% of the 11{,}436 actors.
As a sanity check, a rigid \SI{2}{px} shift of the restored image is recovered as \SI{2.1}{px}.
Restoration also makes more actors detectable (13{,}185 vs.\ 11{,}436 detections).
Matching detections between the two images by box IoU is less informative.
71\% of raw detections have a same-class match with IoU $\ge 0.5$, and most unmatched ones are low-confidence detections on blurred raw renderings, in particular pedestrians, whose flow is equally sub-pixel.

\subsection{LiDAR}
\label{sec:supp_lidar}
\Cref{tab:lidar_seq} reports the geometry metrics of \cref{tab:lidar} per held-out sequence.
\Cref{tab:supp_range} isolates the two LiDAR design choices with the final recipe, the same 4k iterations and the same rays.
Replacing the planar range of \cref{eq:range} by alpha-blended ranges of Gaussian centers degrades CD, F-score and MedSE on every sequence but leaves RMSE unchanged.
Freezing the geometry shared with the cameras, so that LiDAR supervision only trains the LiDAR head, degrades every metric, most strongly CD and MedSE.
Joint optimization is therefore the main source of the gain over SplatAD, and the planar range adds a smaller, consistent one.
Including three further sequences that were never used for design choices, AETHER keeps a lower CD, RMSE and MedSE than SplatAD, while F-score, intensity and ray drop are slightly behind (\cref{tab:supp_lidar6}).

\begin{table*}[t]
  \centering
  \caption{\textbf{LiDAR novel-view synthesis} on held-out PandaSet sweeps: mean over three sequences and geometry per sequence. CD and RMSE in m, MedSE in m$^2$. $^\ddagger$Diverges on one sequence; its mean covers the other two and is not ranked. \rankkey}
  \label{tab:lidar_seq}
  \footnotesize
  \setlength{\tabcolsep}{3.6pt}
  \renewcommand{\arraystretch}{1.08}
  \begin{tabular}{@{}l cccccc ccc ccc ccc@{}}
    \toprule
    & \multicolumn{6}{c}{Mean} & \multicolumn{3}{c}{Sequence 1} & \multicolumn{3}{c}{Sequence 2} & \multicolumn{3}{c}{Sequence 3} \\
    \cmidrule(lr){2-7}\cmidrule(lr){8-10}\cmidrule(lr){11-13}\cmidrule(l){14-16}
    Method & CD $\downarrow$ & F $\uparrow$ & RMSE $\downarrow$ & MedSE $\downarrow$ & Int. $\downarrow$ & Drop $\uparrow$ & CD $\downarrow$ & F $\uparrow$ & RMSE $\downarrow$ & CD $\downarrow$ & F $\uparrow$ & RMSE $\downarrow$ & CD $\downarrow$ & F $\uparrow$ & RMSE $\downarrow$ \\
    \midrule
    LiDAR-GS$^\ddagger$~\cite{chen2024lidargs} & \reff{.357} & \reff{.461} & \reff{18.34} & \reff{.4234} & \reff{.0957} & \reff{.772} & -- & -- & -- & .541 & \third{.241} & 25.32 & \first{.173} & \second{.680} & 11.36 \\
    GS-LiDAR~\cite{jiang2025gslidar} & .508 & \first{.495} & 9.44 & .0742 & .0940 & .825 & .509 & \second{.475} & 9.49 & .765 & \second{.299} & \third{12.22} & .251 & \first{.710} & 6.61 \\
    NeuRAD~\cite{tonderski2024neurad} & \third{.414} & .183 & \third{7.53} & \third{.0388} & \third{.0887} & \third{.957} & \third{.394} & .180 & \third{5.02} & \third{.496} & .132 & 13.25 & .353 & .238 & \third{4.30} \\
    SplatAD~\cite{hess2025splatad} & \second{.337} & \third{.470} & \second{3.75} & \second{.0082} & \first{.0588} & \first{.967} & \second{.316} & \third{.472} & \second{3.41} & \second{.453} & \first{.316} & \second{5.15} & \third{.242} & .622 & \second{2.69} \\
    \midrule
    \ours & \first{.314} & \second{.474} & \first{2.85} & \first{.0066} & \second{.0626} & \second{.966} & \first{.294} & \first{.477} & \first{2.71} & \first{.429} & \second{.299} & \first{4.11} & \second{.220} & \third{.647} & \first{1.74} \\
    \bottomrule
  \end{tabular}
\end{table*}

\begin{table}[t]
  \centering
  \caption{\textbf{LiDAR ablation}, mean over the three held-out sequences. CD in m, MedSE in m$^2$. Best in bold.}
  \label{tab:supp_range}
  \footnotesize
  \setlength{\tabcolsep}{2.4pt}
  \begin{tabular}{@{}lcccccc@{}}
    \toprule
    Variant & CD $\downarrow$ & F $\uparrow$ & RMSE $\downarrow$ & MedSE $\downarrow$ & Int. $\downarrow$ & Drop $\uparrow$ \\
    \midrule
    SplatAD~\cite{hess2025splatad} & .337 & .470 & 3.75 & .0082 & \textbf{.0588} & \textbf{.967} \\
    \midrule
    Center ranges                  & .319 & .466 & 2.85 & .0074 & .0627 & .965 \\
    Frozen shared geometry         & .427 & .363 & 3.54 & .0180 & .0725 & .960 \\
    Full                           & \textbf{.314} & \textbf{.474} & \textbf{2.85} & \textbf{.0066} & .0626 & .966 \\
    \bottomrule
  \end{tabular}
\end{table}

\begin{table}[t]
  \centering
  \caption{\textbf{AETHER vs.\ SplatAD on all six held-out sequences}, including three never used for design choices. Mean values; CD in m, MedSE in m$^2$.}
  \label{tab:supp_lidar6}
  \footnotesize
  \setlength{\tabcolsep}{2.4pt}
  \begin{tabular}{@{}lcccccc@{}}
    \toprule
    Method & CD $\downarrow$ & F $\uparrow$ & RMSE $\downarrow$ & MedSE $\downarrow$ & Int. $\downarrow$ & Drop $\uparrow$ \\
    \midrule
    SplatAD~\cite{hess2025splatad} & .350 & \textbf{.457} & 4.09 & .0074 & \textbf{.0554} & \textbf{.966} \\
    \ours                          & \textbf{.336} & .453 & \textbf{3.11} & \textbf{.0067} & .0590 & .963 \\
    \bottomrule
  \end{tabular}
\end{table}

\subsection{Ground and pose}
\label{sec:supp_pose}
\Cref{tab:supp_ground} gives the mean height errors of \cref{fig:actors}a, and \cref{tab:supp_pose} the camera-pose errors behind \cref{sec:exp_ground}.
Anchoring to the logged contacts removes most of the pose error of the depth-based surface without losing its height accuracy, but it does not beat a single plane on these short, mostly flat clips in position or jitter.

\begin{table}[t]
  \centering
  \caption{\textbf{Ground height error} (cm) against accumulated LiDAR ground, mean over 10 scenes.}
  \label{tab:supp_ground}
  \footnotesize
  \setlength{\tabcolsep}{4pt}
  \begin{tabular}{@{}lcc@{}}
    \toprule
    Surface & Driven corridor & Elsewhere \\
    \midrule
    Vertical columns (DecoupleGS~\cite{li2026decouplegs}) & 141.9 & 187.8 \\
    Lowest Gaussian layer       & 60.3 & 110.5 \\
    Gaussian centers            & 49.9 & 51.7 \\
    Top-down columns            & 26.7 & 31.0 \\
    \ours                       & \textbf{5.2} & \textbf{14.5} \\
    \bottomrule
  \end{tabular}
\end{table}

\begin{table}[t]
  \centering
  \caption{\textbf{Camera pose along the logged path} (8 scenes): rotation and position error against the logged cameras and frame-to-frame rotation error. Depth: surface from rendered depth.}
  \label{tab:supp_pose}
  \footnotesize
  \setlength{\tabcolsep}{3pt}
  \begin{tabular}{@{}lccc@{}}
    \toprule
    Ground & Rot.\ (\si{\degree})\,$\downarrow$ & Pos.\ (cm)\,$\downarrow$ & Rel.\ (\si{\degree}/fr.)\,$\downarrow$ \\
    \midrule
    Single plane                  & .639 & \textbf{2.02} & \textbf{.141} \\
    Gaussian centers              & .669 & 2.56 & .142 \\
    Depth, not anchored           & .748 & 6.95 & .161 \\
    Depth, anchored (ours)        & \textbf{.636} & 3.09 & .150 \\
    \bottomrule
  \end{tabular}
\end{table}

\subsection{Pedestrians and riders}
\label{sec:supp_vru}
\noindent\textbf{Root.}
The fitted start-up dynamics reproduce the distribution of start-up times on held-out tracks (1-D Wasserstein distance \SI{0.34}{s}, against \SI{1.42}{s} for an instantaneous start).
In 93\% of real starts from a misaligned facing, pedestrians start moving before they finish turning, which our coupled facing state reproduces and a prescribed path does not.
ADE differs little between root models (\cref{tab:vru}) because it is dominated by the \SI{2}{Hz} annotation noise; facing error and the acceleration and yaw-rate distributions separate them more clearly.

\noindent\textbf{Body.}
Besides the metrics in \cref{tab:vru}, our body has a support-foot height error of \SI{0.39}{cm} (mocap \SI{0.30}{cm}) and slides more than \SI{2.5}{cm} in 2.5\% of the frames (clip playback 18.9\%).
Plain motion matching has slightly lower pelvis jerk (83 vs.\ \SI{88}{m/s^3}) and pose jumps (\ang{14.1} vs.\ \ang{19.0}).
Contact locking and inverse kinematics trade these for 43\% less skating and no penetration.

\noindent\textbf{Riders.}
The hand, foot and pelvis gaps are 13.7, 28.4 and \SI{27.7}{cm} for clip playback, zero for the rigid rider, and 0, 0.8 and \SI{0}{cm} for procedural IK and ours.
Only ours leans into turns, with a lean error of \ang{0.06} against \ang{0.62} for both new baselines and \ang{0.51} for clip playback.

\noindent\textbf{Interaction.}
Our pedestrians collide only when the ego keeps \SI{8}{m/s} without reacting (5\% of the seeds of that cell), enter earlier in 92\% of the cases in which the ego yields and never later.
They show zero deviation in all negative controls and are deterministic given the seed.
Riders do not collide and enter earlier in 65\% of the cases in which the ego yields.
The scripted walker never enters earlier when yielded to and deviates by \SI{14.3}{m} in the parallel-road control.
Social force collides in 10 to 42\% of the cells, and SF\,+\,ORCA enters earlier in 17\% of the yielding cases.

\subsection{Fidelity flag in closed loop}
\label{sec:supp_flag}
The flag is evaluated on every step of the held-out episodes of \cref{sec:exp_loop}.
\Cref{tab:supp_psi} varies the heading bound.
Most untrusted steps exceed the heading bound (13.5\% of all steps) rather than the lateral one (2.0\%; 4.3\% both).

\begin{table}[t]
  \centering
  \caption{\textbf{Sensitivity to the heading bound} $\psi_{\max}$. Trusted steps, episodes that leave the trusted region, median time of the first exit, and share of at-fault collisions under reactive traffic that occur in a trusted step.}
  \label{tab:supp_psi}
  \footnotesize
  \setlength{\tabcolsep}{3pt}
  \begin{tabular}{@{}llcccc@{}}
    \toprule
    $\psi_{\max}$ & Planner & Trusted & Exit & First exit & At fault, trusted \\
    \midrule
    \ang{5}  & TransFuser     & 68\% & 53\% & \SI{1.0}{s} & 58\% \\
             & UniAD          & 72\% & 64\% & \SI{1.1}{s} & 72\% \\
             & GaussianFusion & 42\% & 89\% & \SI{1.5}{s} & 44\% \\
             & VAD            & 69\% & 55\% & \SI{0.8}{s} & 35\% \\
    \midrule
    \ang{10} & TransFuser     & 82\% & 38\% & \SI{2.0}{s} & 79\% \\
             & UniAD          & 87\% & 39\% & \SI{3.3}{s} & 76\% \\
             & GaussianFusion & 61\% & 68\% & \SI{2.0}{s} & 59\% \\
             & VAD            & 82\% & 35\% & \SI{2.0}{s} & 51\% \\
    \midrule
    \ang{15} & TransFuser     & 87\% & 28\% & \SI{2.6}{s} & 79\% \\
             & UniAD          & 93\% & 20\% & \SI{4.0}{s} & 86\% \\
             & GaussianFusion & 73\% & 47\% & \SI{2.5}{s} & 69\% \\
             & VAD            & 87\% & 28\% & \SI{3.3}{s} & 53\% \\
    \bottomrule
  \end{tabular}
\end{table}

\subsection{Collision attribution}
\label{sec:supp_collision}
Because log replay and reactive traffic start from the same scenes and seeds, episodes are paired by scene and seed.
\Cref{tab:supp_paired} gives the paired differences (reactive minus replay) with 95\% scene-bootstrap intervals.
Not-at-fault collisions drop for every planner; at-fault collisions rise slightly for every planner, but no interval excludes zero (pooled $+0.04$, $[-0.02, 0.10]$).
Of the 200 not-at-fault vehicle collisions under replay, 98\% are rear-end collisions, and the ego was stationary in 10\% of them.
In only 6\% did the ego brake harder than \SI{4}{m/s^2} within the preceding \SI{2}{s} (8\% for \SI{3}{m/s^2}), so these collisions are not produced by abrupt ego braking.

\begin{table}[t]
  \centering
  \caption{\textbf{Paired collision rates}, reactive minus replay, with 95\% scene-bootstrap intervals.}
  \label{tab:supp_paired}
  \footnotesize
  \setlength{\tabcolsep}{3pt}
  \begin{tabular}{@{}lccc@{}}
    \toprule
    Planner & Pairs & At fault & Not at fault \\
    \midrule
    TransFuser     & 161 & $+0.03\ [-0.05, 0.12]$ & $-0.30\ [-0.42, -0.19]$ \\
    UniAD          & 165 & $+0.02\ [-0.08, 0.12]$ & $-0.32\ [-0.44, -0.20]$ \\
    GaussianFusion & 165 & $+0.05\ [-0.01, 0.11]$ & $-0.30\ [-0.42, -0.19]$ \\
    VAD            & 165 & $+0.07\ [-0.01, 0.15]$ & $-0.21\ [-0.32, -0.12]$ \\
    \midrule
    All            & 656 & $+0.04\ [-0.02, 0.10]$ & $-0.28\ [-0.38, -0.20]$ \\
    \bottomrule
  \end{tabular}
\end{table}

\subsection{Crossing tests}
\label{sec:supp_crossing}
\begin{figure}[t]
  \centering
  \includegraphics[width=\columnwidth]{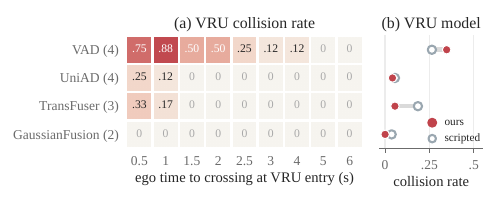}
  \caption{\textbf{Crossing tests.} (a) Rate of collisions with a pedestrian or rider that enters the ego lane when the ego is a given time from the crossing; number of scenes in brackets. (b) The same rate with scripted crossings.}
  \label{fig:crossing}
\end{figure}

Because pedestrians and riders are timed against the ego, any log can be turned into a controlled crossing test.
A pedestrian (\SI{1.4}{m/s}) or rider (\SI{4}{m/s}) enters the ego lane when the ego is 0.5 to \SI{6}{s} from the crossing (9 values).
The crossing is placed on the planner's own crossing-free rollout, so that it is reached in every member, and only on scenes where the planner drives safely without it; each value is run once per scene and road-user type.
This gives 72 episodes per VRU model for UniAD and VAD (4 scenes), 54 for TransFuser (3 scenes) and 27 for GaussianFusion (2 scenes for pedestrians, 1 for riders).
Outcomes are not monotone in the margin, because a crosser occupies the lane for several seconds, so we report the collision rate over the fixed grid instead of a boundary.
UniAD, TransFuser and GaussianFusion collide only at the two shortest margins, whereas VAD still hits half of the crossers \SI{2}{s} ahead and some up to \SI{4}{s} ahead (\cref{fig:crossing}).
Scripted crossings give similar rates (scripted vs.\ ours: pedestrians 0.16 vs.\ 0.20, riders 0.14 vs.\ 0.07), so the gap between planners does not stem from our VRU model.

\section{Visualizations}
\label{sec:supp_vis}

\subsection{Closed-loop sequences}
\label{sec:supp_vis_loop}
\begin{figure*}[t]
  \centering
  \setlength{\tabcolsep}{0pt}
  \renewcommand{\arraystretch}{0}
  \begin{tabular}{@{}c@{\hspace{3pt}}c@{}}
    & \makebox[0.1905\textwidth]{\scriptsize \SI{2.0}{s}, walk}\hspace{0.0019\textwidth} \makebox[0.1905\textwidth]{\scriptsize \SI{4.75}{s}, wait}\hspace{0.0019\textwidth} \makebox[0.1905\textwidth]{\scriptsize \SI{6.25}{s}, step off}\hspace{0.0019\textwidth} \makebox[0.1905\textwidth]{\scriptsize \SI{8.75}{s}, cross}\hspace{0.0019\textwidth} \makebox[0.1905\textwidth]{\scriptsize \SI{11.25}{s}, cross} \\[3pt]
    \seqlabel{Camera} & \includegraphics[width=0.960\textwidth]{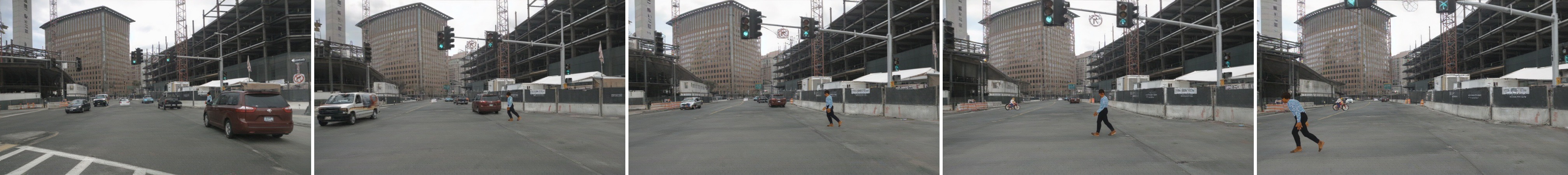} \\[2pt]
    \seqlabel{LiDAR} & \includegraphics[width=0.960\textwidth]{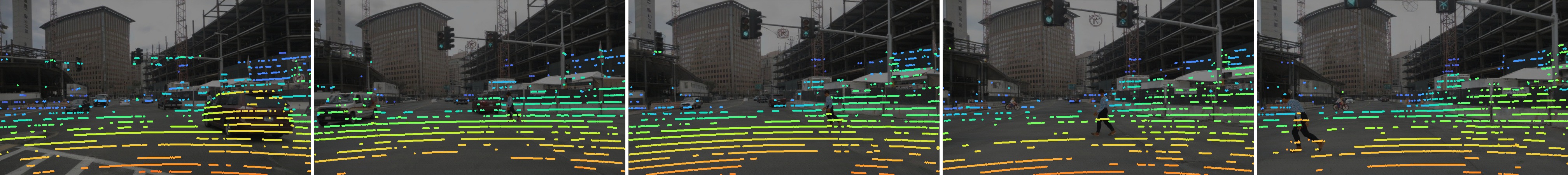} \\[2pt]
    \seqlabel{BEV} & \includegraphics[width=0.960\textwidth]{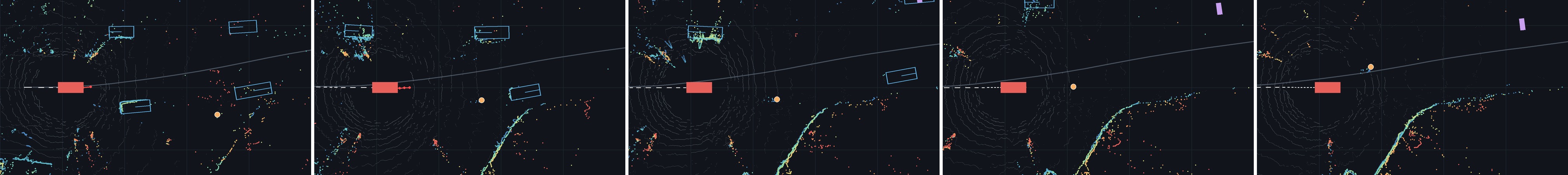} \\
  \end{tabular}
  \caption{\textbf{Closed loop with a crossing pedestrian} (UniAD, reactive traffic). The inserted pedestrian walks to the curb, waits, accepts the gap left by the slowing ego and crosses in front of it; a rider yields at the far side (purple). Rows: front camera after NVR, the LiDAR sweep of the same step projected into it (colored by range), and BEV. In the BEV ($x$ forward), LiDAR points are colored by height, vehicles are blue boxes, pedestrians orange dots and riders purple boxes; the ego is red, its plan a red line and its driven path a dashed white line.}
  \label{fig:supp_seq_ped}
\end{figure*}

\begin{figure*}[t]
  \centering
  \setlength{\tabcolsep}{0pt}
  \renewcommand{\arraystretch}{0}
  \begin{tabular}{@{}c@{\hspace{3pt}}c@{}}
    & \makebox[0.1905\textwidth]{\scriptsize \SI{2.0}{s}, ride}\hspace{0.0019\textwidth} \makebox[0.1905\textwidth]{\scriptsize \SI{3.5}{s}, ride}\hspace{0.0019\textwidth} \makebox[0.1905\textwidth]{\scriptsize \SI{5.0}{s}, yield}\hspace{0.0019\textwidth} \makebox[0.1905\textwidth]{\scriptsize \SI{6.5}{s}, yield}\hspace{0.0019\textwidth} \makebox[0.1905\textwidth]{\scriptsize \SI{8.0}{s}, braking} \\[3pt]
    \seqlabel{Camera} & \includegraphics[width=0.960\textwidth]{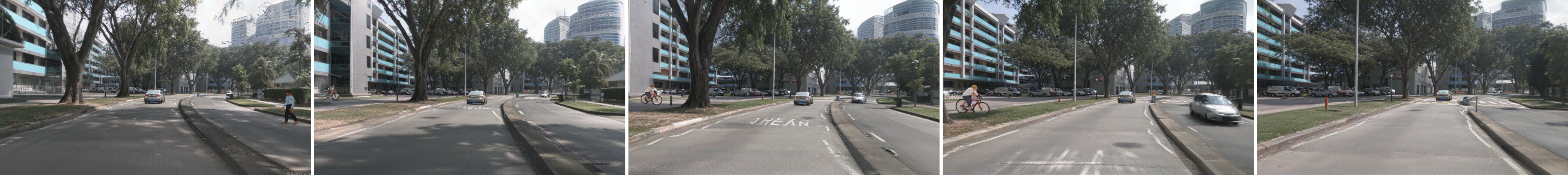} \\[2pt]
    \seqlabel{LiDAR} & \includegraphics[width=0.960\textwidth]{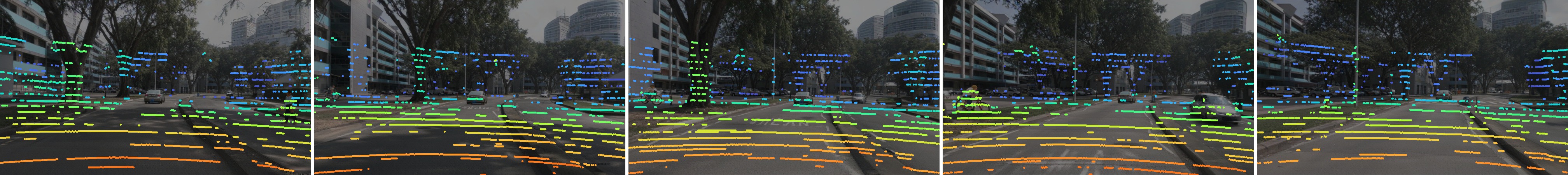} \\[2pt]
    \seqlabel{BEV} & \includegraphics[width=0.960\textwidth]{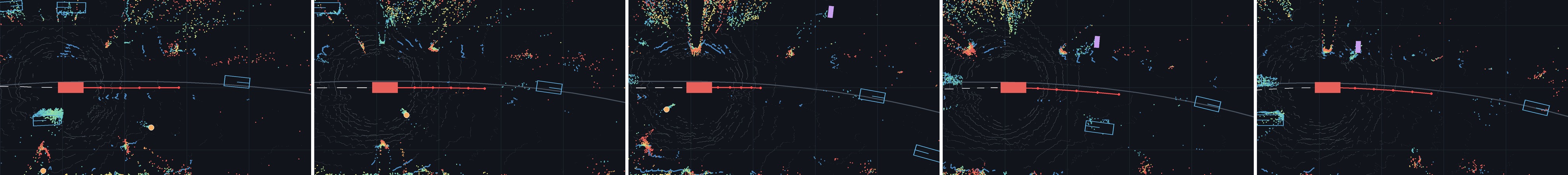} \\
  \end{tabular}
  \caption{\textbf{Closed loop with a crossing cyclist} (UniAD, reactive traffic). The cyclist approaches from the left, does not accept the gap to the ego, brakes and stops before the road edge while the ego passes. A pedestrian walks along the right sidewalk. Rows and BEV legend as in \cref{fig:supp_seq_ped}.}
  \label{fig:supp_seq_cyc}
\end{figure*}

\begin{figure*}[t]
  \centering
  \setlength{\tabcolsep}{0pt}
  \renewcommand{\arraystretch}{0}
  \begin{tabular}{@{}c@{\hspace{3pt}}c@{}}
    & \makebox[0.1905\textwidth]{\scriptsize \SI{0}{s}}\hspace{0.0019\textwidth} \makebox[0.1905\textwidth]{\scriptsize \SI{2}{s}}\hspace{0.0019\textwidth} \makebox[0.1905\textwidth]{\scriptsize \SI{4}{s}}\hspace{0.0019\textwidth} \makebox[0.1905\textwidth]{\scriptsize \SI{6}{s}}\hspace{0.0019\textwidth} \makebox[0.1905\textwidth]{\scriptsize \SI{8}{s}} \\[3pt]
    \seqlabel{Camera} & \includegraphics[width=0.960\textwidth]{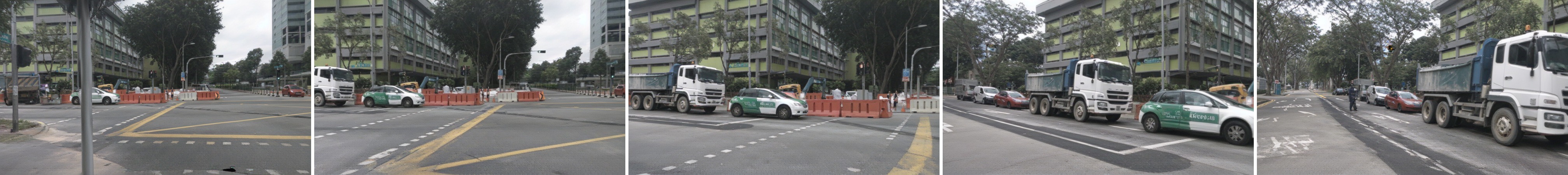} \\[2pt]
    \seqlabel{LiDAR} & \includegraphics[width=0.960\textwidth]{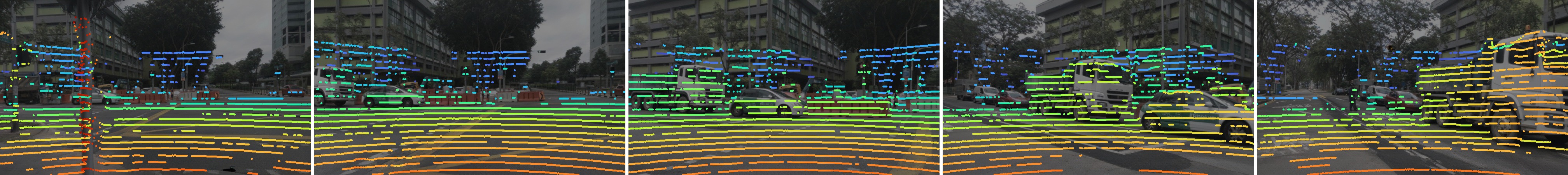} \\[2pt]
    \seqlabel{BEV} & \includegraphics[width=0.960\textwidth]{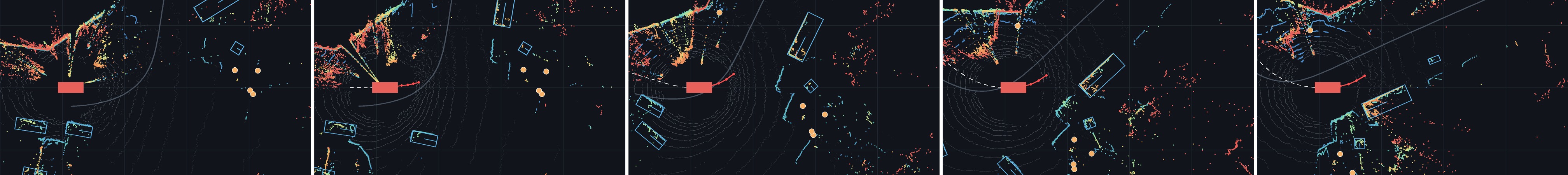} \\
  \end{tabular}
  \caption{\textbf{Closed loop started \SI{3}{m} off the log} (UniAD). Camera and LiDAR are rendered from a viewpoint that was never recorded. UniAD steers back toward the route, then takes the turn wide and leaves the drivable area. Rows and BEV legend as in \cref{fig:supp_seq_ped}.}
  \label{fig:supp_seq_off}
\end{figure*}

\begin{figure*}[t]
  \centering
  \setlength{\tabcolsep}{0pt}
  \renewcommand{\arraystretch}{0}
  \begin{tabular}{@{}c@{\hspace{3pt}}c@{}}
    & \makebox[0.1905\textwidth]{\scriptsize \SI{0.75}{s}}\hspace{0.0019\textwidth} \makebox[0.1905\textwidth]{\scriptsize \SI{1.5}{s}}\hspace{0.0019\textwidth} \makebox[0.1905\textwidth]{\scriptsize \SI{2.25}{s}}\hspace{0.0019\textwidth} \makebox[0.1905\textwidth]{\scriptsize \SI{3.0}{s}}\hspace{0.0019\textwidth} \makebox[0.1905\textwidth]{\scriptsize \SI{4.0}{s}} \\[3pt]
    \seqlabel[0.0505\textwidth]{Camera} & \includegraphics[width=0.960\textwidth]{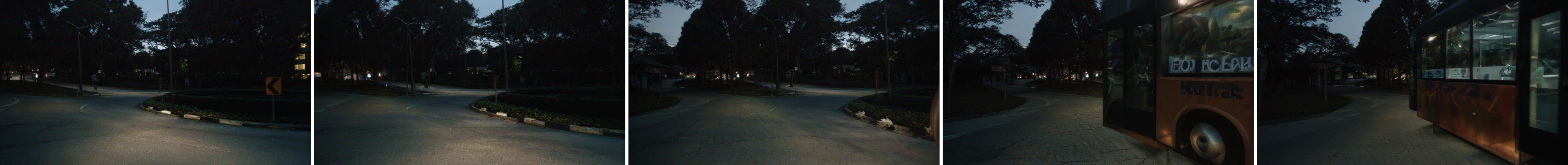} \\[2pt]
    \seqlabel[0.0505\textwidth]{LiDAR} & \includegraphics[width=0.960\textwidth]{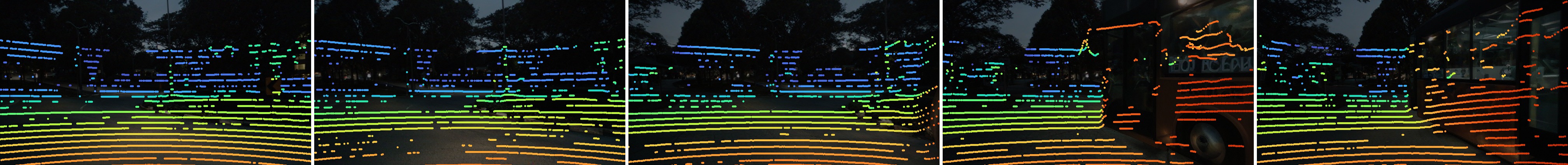} \\[2pt]
    \seqlabel[0.0505\textwidth]{BEV} & \includegraphics[width=0.960\textwidth]{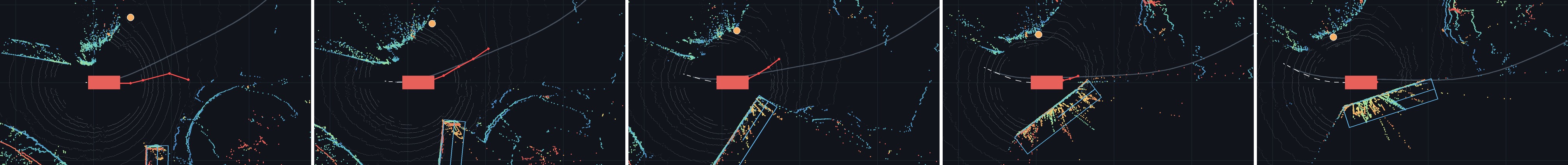} \\
  \end{tabular}
  \caption{\textbf{An at-fault collision at night} (UniAD, reactive traffic). Entering the junction, UniAD does not stop for a bus approaching from the right and the episode ends with a collision at \SI{4.25}{s}. Rows and BEV legend as in \cref{fig:supp_seq_ped}.}
  \label{fig:supp_seq_night}
\end{figure*}

\begin{figure*}[t]
  \centering
  \setlength{\tabcolsep}{0pt}
  \renewcommand{\arraystretch}{0}
  \begin{tabular}{@{}c@{\hspace{3pt}}c@{}}
    & \makebox[0.1905\textwidth]{\scriptsize \SI{1.0}{s}}\hspace{0.0019\textwidth} \makebox[0.1905\textwidth]{\scriptsize \SI{2.5}{s}}\hspace{0.0019\textwidth} \makebox[0.1905\textwidth]{\scriptsize \SI{4.0}{s}}\hspace{0.0019\textwidth} \makebox[0.1905\textwidth]{\scriptsize \SI{5.0}{s}}\hspace{0.0019\textwidth} \makebox[0.1905\textwidth]{\scriptsize \SI{5.5}{s}} \\[3pt]
    \seqlabel{Replay} & \includegraphics[width=0.960\textwidth]{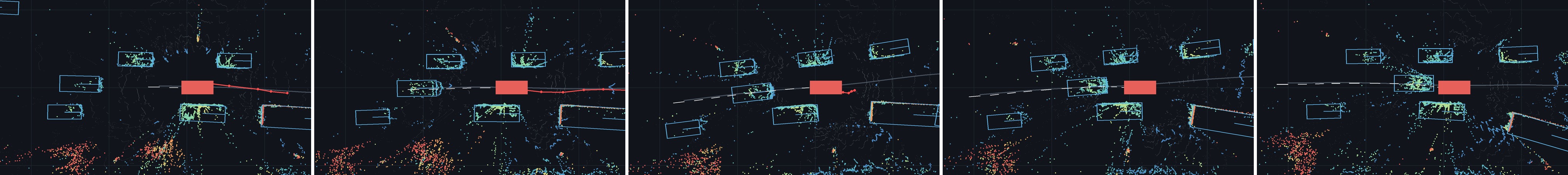} \\[2pt]
    \seqlabel{Reactive} & \includegraphics[width=0.960\textwidth]{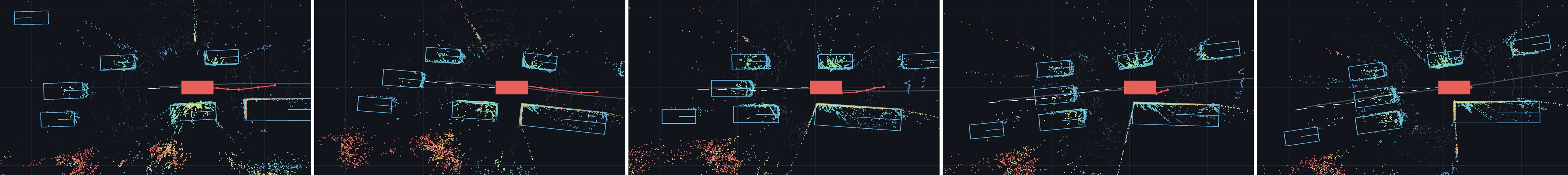} \\
  \end{tabular}
  \caption{\textbf{Log replay vs.\ reactive traffic} on the same scene and seed (UniAD, BEV). Under replay, the logged follower behind the ego keeps its recorded speed while the ego drives slower than the recording and rear-ends it at \SI{5.75}{s} (not at fault). Under reactive traffic the same vehicle brakes behind the ego and the episode runs to the time limit. Rows and BEV legend as in \cref{fig:supp_seq_ped}.}
  \label{fig:supp_seq_rear}
\end{figure*}

\Cref{fig:supp_seq_ped,fig:supp_seq_cyc,fig:supp_seq_off,fig:supp_seq_night,fig:supp_seq_rear} show closed-loop episodes of UniAD as the planner saw them, every few steps.
Each column is one simulation step: the restored front camera that UniAD received, the LiDAR sweep rendered at the same step and projected into that camera, and a bird's-eye view of the sweep with the simulated road users, the current plan and the path driven so far.
No frame is re-rendered for the figure, and the LiDAR is drawn from the same posed Gaussians as the image, so the points stay on the moving pedestrian, rider and vehicles.
The first two episodes show interaction: a pedestrian who waits for and then uses the gap that the slowing ego leaves (\cref{fig:supp_seq_ped}), and a cyclist who yields to the ego (\cref{fig:supp_seq_cyc}).
The next two show failures that AETHER exposes: an episode started \SI{3}{m} off the log, where UniAD regains the route but then leaves the drivable area in the turn (\cref{fig:supp_seq_off}), and a night episode with an at-fault collision with a bus (\cref{fig:supp_seq_night}).
\Cref{fig:supp_seq_rear} contrasts log replay and reactive traffic on the same scene and seed: the rear-end collision under replay, typical of the not-at-fault collisions in \cref{sec:supp_collision}, disappears when the follower reacts to the ego.

\subsection{Pedestrian and rider motion}
\label{sec:supp_vis_vru_dyn}
\begin{figure*}[t]
  \centering
  \setlength{\tabcolsep}{0pt}
  \renewcommand{\arraystretch}{0}
  \begin{tabular}{@{}c@{}}
    \makebox[0.1176\textwidth]{\scriptsize \SI{5.0}{s} wait}\hspace{0.0027\textwidth} \makebox[0.1176\textwidth]{\scriptsize \SI{5.5}{s} wait}\hspace{0.0027\textwidth} \makebox[0.1176\textwidth]{\scriptsize \SI{6.0}{s} look}\hspace{0.0027\textwidth} \makebox[0.1176\textwidth]{\scriptsize \SI{6.5}{s} step off}\hspace{0.0027\textwidth} \makebox[0.1176\textwidth]{\scriptsize \SI{7.0}{s} cross}\hspace{0.0027\textwidth} \makebox[0.1176\textwidth]{\scriptsize \SI{7.5}{s} cross}\hspace{0.0027\textwidth} \makebox[0.1176\textwidth]{\scriptsize \SI{8.0}{s} cross}\hspace{0.0027\textwidth} \makebox[0.1176\textwidth]{\scriptsize \SI{8.5}{s} cross} \\[3pt]
    \includegraphics[width=0.96\textwidth]{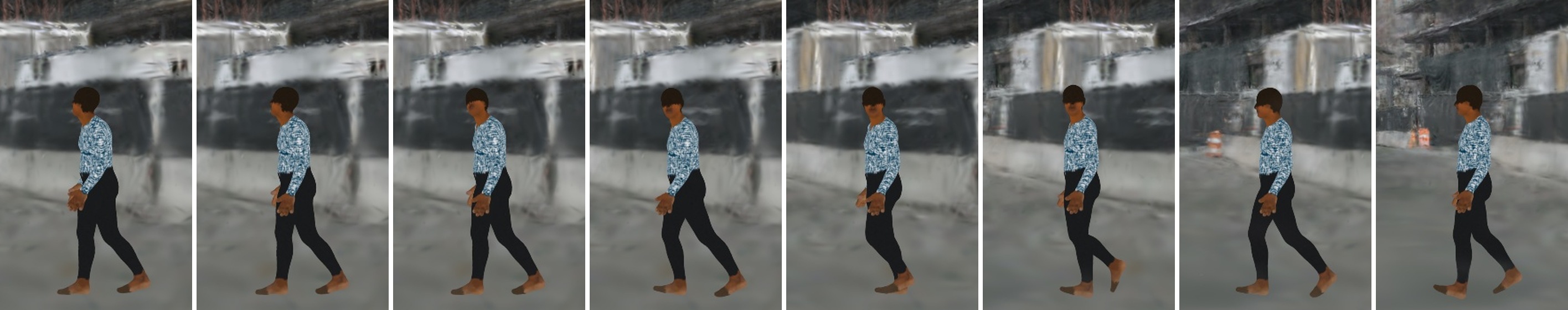} \\[4pt]
    {\small (a) Pedestrian, seen from the ego's front camera while the ego waits about \SI{8}{m} before the crossing} \\[8pt]
    \makebox[0.1055\textwidth]{\scriptsize \SI{7.0}{s} brake}\hspace{0.0013\textwidth} \makebox[0.1055\textwidth]{\scriptsize \SI{7.5}{s} brake}\hspace{0.0013\textwidth} \makebox[0.1055\textwidth]{\scriptsize \SI{8.0}{s} brake}\hspace{0.0013\textwidth} \makebox[0.1055\textwidth]{\scriptsize \SI{8.5}{s} stop}\hspace{0.0013\textwidth} \makebox[0.1055\textwidth]{\scriptsize \SI{9.0}{s} foot down}\hspace{0.0013\textwidth} \makebox[0.1055\textwidth]{\scriptsize \SI{11.5}{s} foot down}\hspace{0.0013\textwidth} \makebox[0.1055\textwidth]{\scriptsize \SI{12.5}{s} look}\hspace{0.0013\textwidth} \makebox[0.1055\textwidth]{\scriptsize \SI{13.0}{s} pedal}\hspace{0.0013\textwidth} \makebox[0.1055\textwidth]{\scriptsize \SI{13.5}{s} pedal} \\[3pt]
    \includegraphics[width=0.96\textwidth]{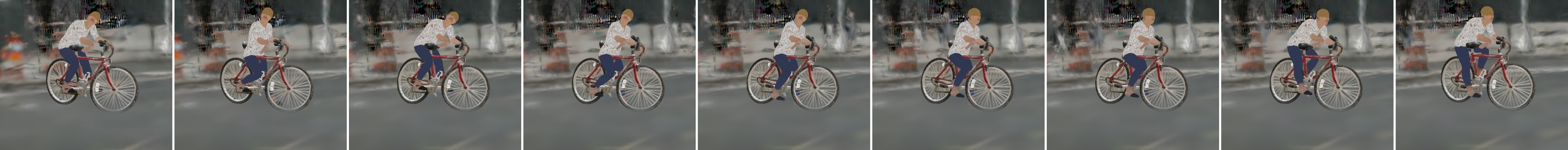} \\[4pt]
    {\small (b) Cyclist, seen from a fixed camera at the road edge while the ego passes in front of it} \\
  \end{tabular}
  \caption{\textbf{Pedestrian and rider motion over time}, crops of raw renderings (no restorer) labeled with time and the behavior state of \cref{alg:supp_ped,alg:supp_rider}. (a) The pedestrian stops at the curb, turns its head toward the stopped ego, accepts the gap and steps off with a gait that speeds up over the first steps. (b) The cyclist brakes, stops with one foot on the ground, looks before restarting and pedals away. Both are driven by the same models as in closed loop, here with a scripted ego.}
  \label{fig:supp_vru_dyn}
\end{figure*}

\Cref{fig:supp_vru_dyn} follows one pedestrian and one rider over several seconds at a finer interval than the closed-loop figures, cropped around the road user.
The pedestrian's head turns toward the stopped ego before it steps off, and its speed builds up over the first steps instead of jumping to the walking speed; the cyclist brakes to a stop, puts a foot down, looks and pedals away.
These transitions come from the state machines of \cref{alg:supp_ped,alg:supp_rider} acting on the root and body layers, not from scripted clips.

\subsection{Sensors and assets}
\label{sec:supp_vis_assets}
\begin{figure*}[p]
  \centering
  \small
  \includegraphics[width=\textwidth]{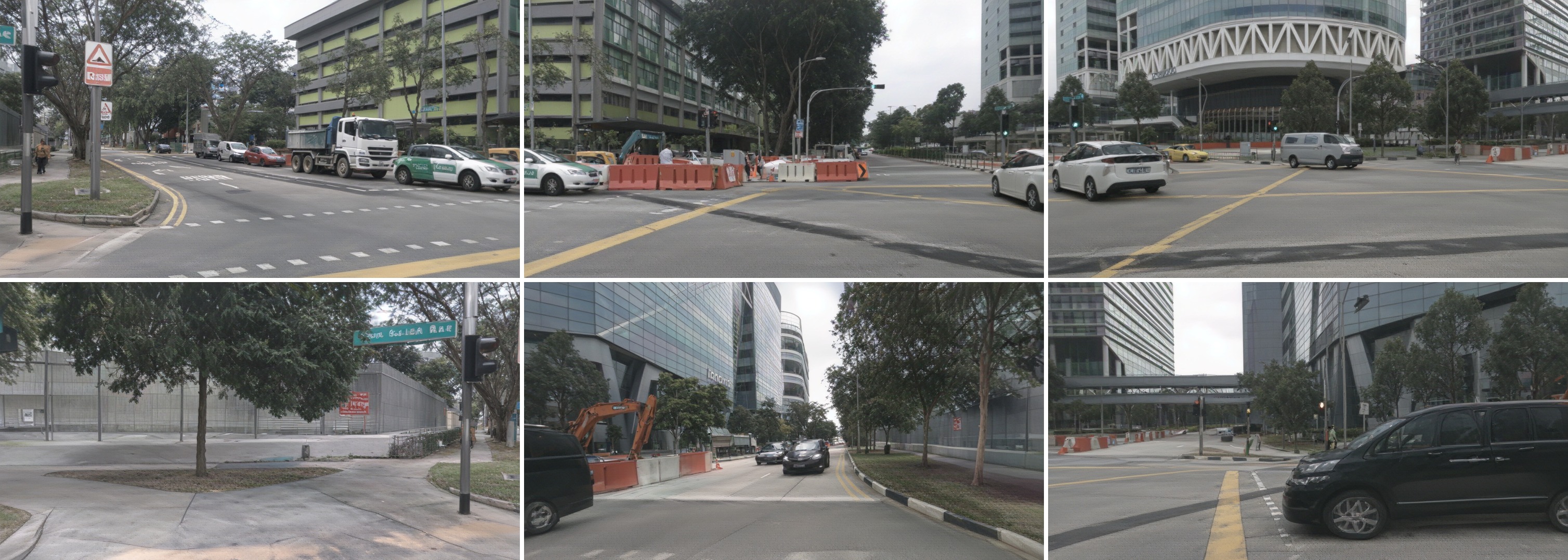}\\[1pt]
  (a) Daytime, ego on its logged path\\[5pt]
  \includegraphics[width=\textwidth]{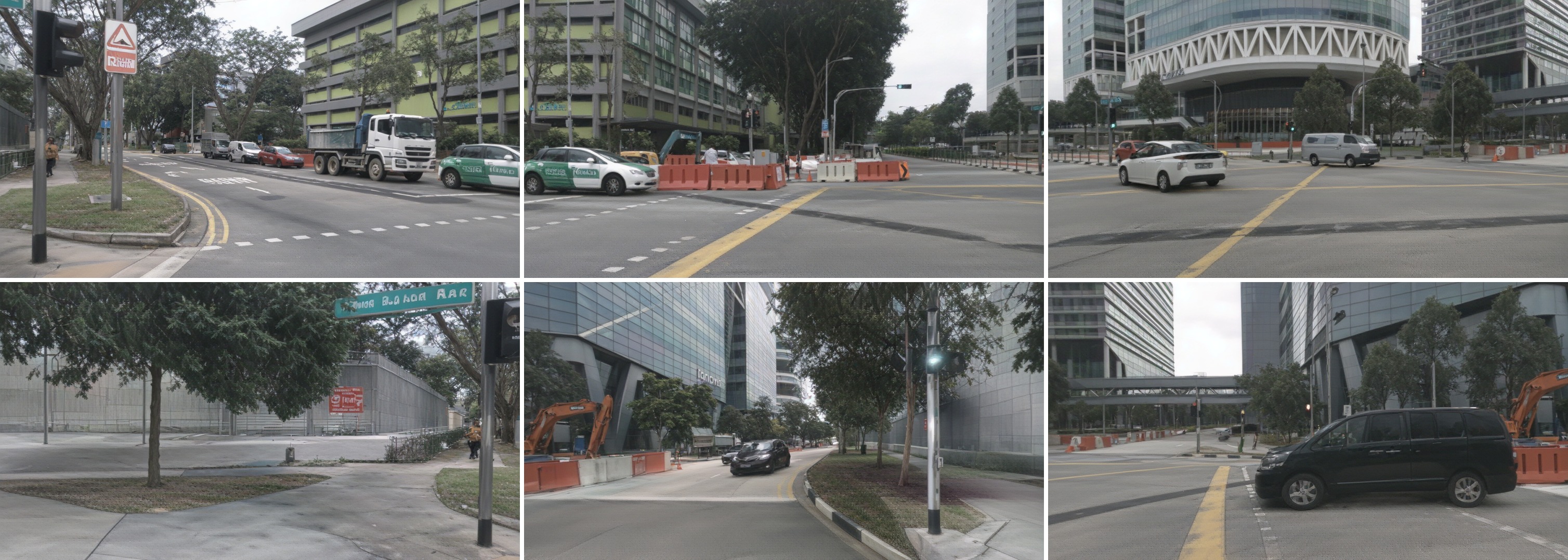}\\[1pt]
  (b) Same instant, ego moved \SI{2}{m} to the left of its logged path\\[5pt]
  \includegraphics[width=\textwidth]{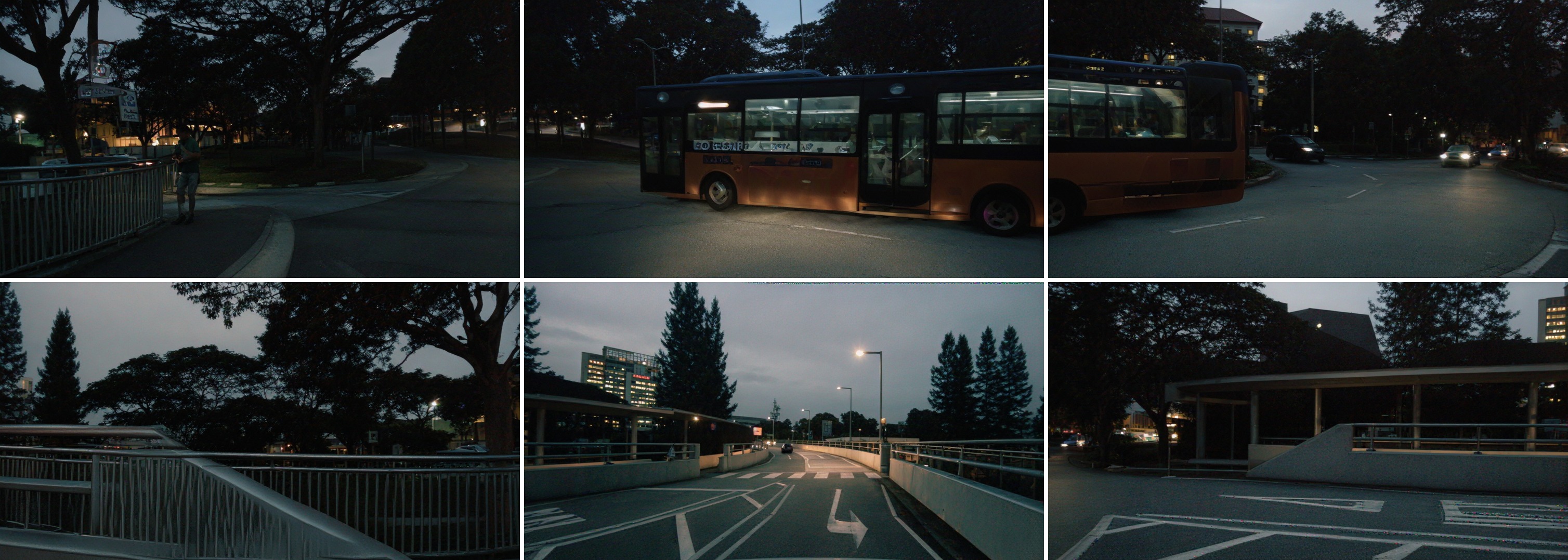}\\[1pt]
  (c) Night, ego on its logged path
  \caption{\textbf{All six cameras rendered by AETHER} at $1600{\times}900$ after restoration. In each panel, the top row shows the front-left, front and front-right cameras and the bottom row the back-left, back and back-right cameras.}
  \label{fig:supp_surround}
\end{figure*}

\begin{figure*}[p]
  \centering
  \small
  \setlength{\tabcolsep}{1.5pt}
  \begin{tabular}{cc}
    Recorded sweep & AETHER \\[1pt]
    \multicolumn{2}{c}{(a) Point cloud, oblique view} \\[1pt]
    \includegraphics[width=0.49\textwidth]{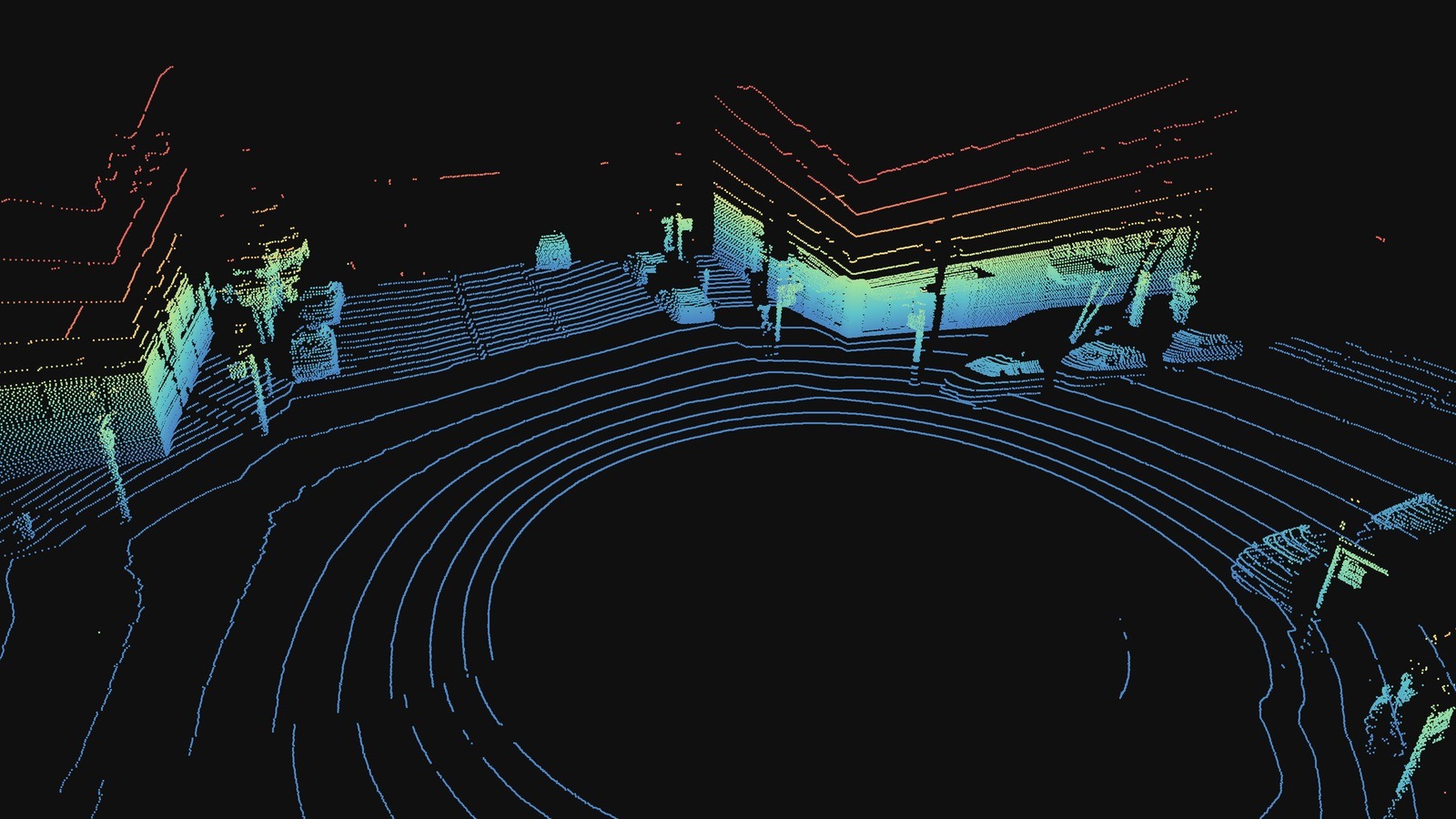} &
    \includegraphics[width=0.49\textwidth]{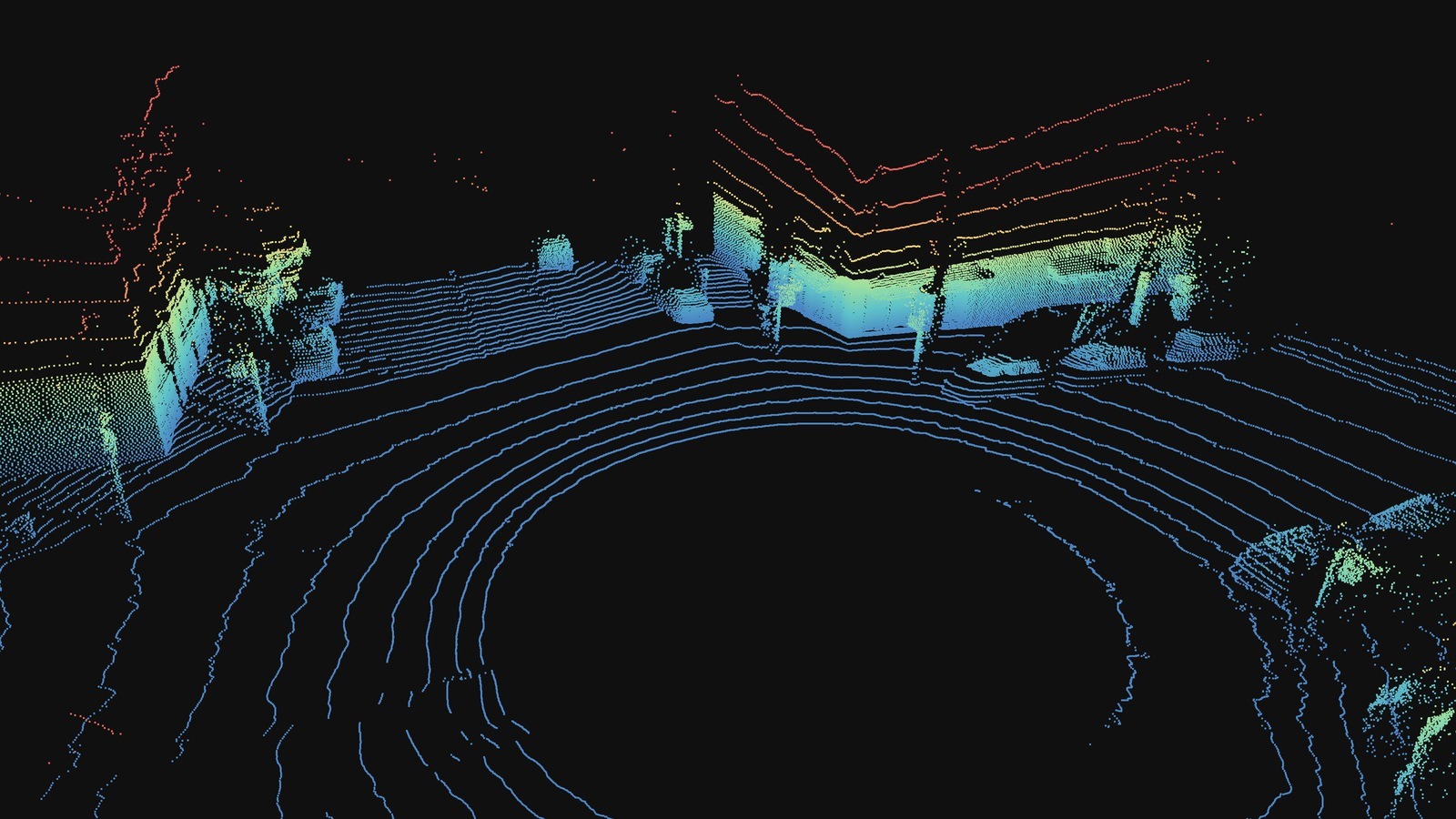} \\[3pt]
    \multicolumn{2}{c}{(b) Point cloud, top view} \\[1pt]
    \includegraphics[width=0.40\textwidth]{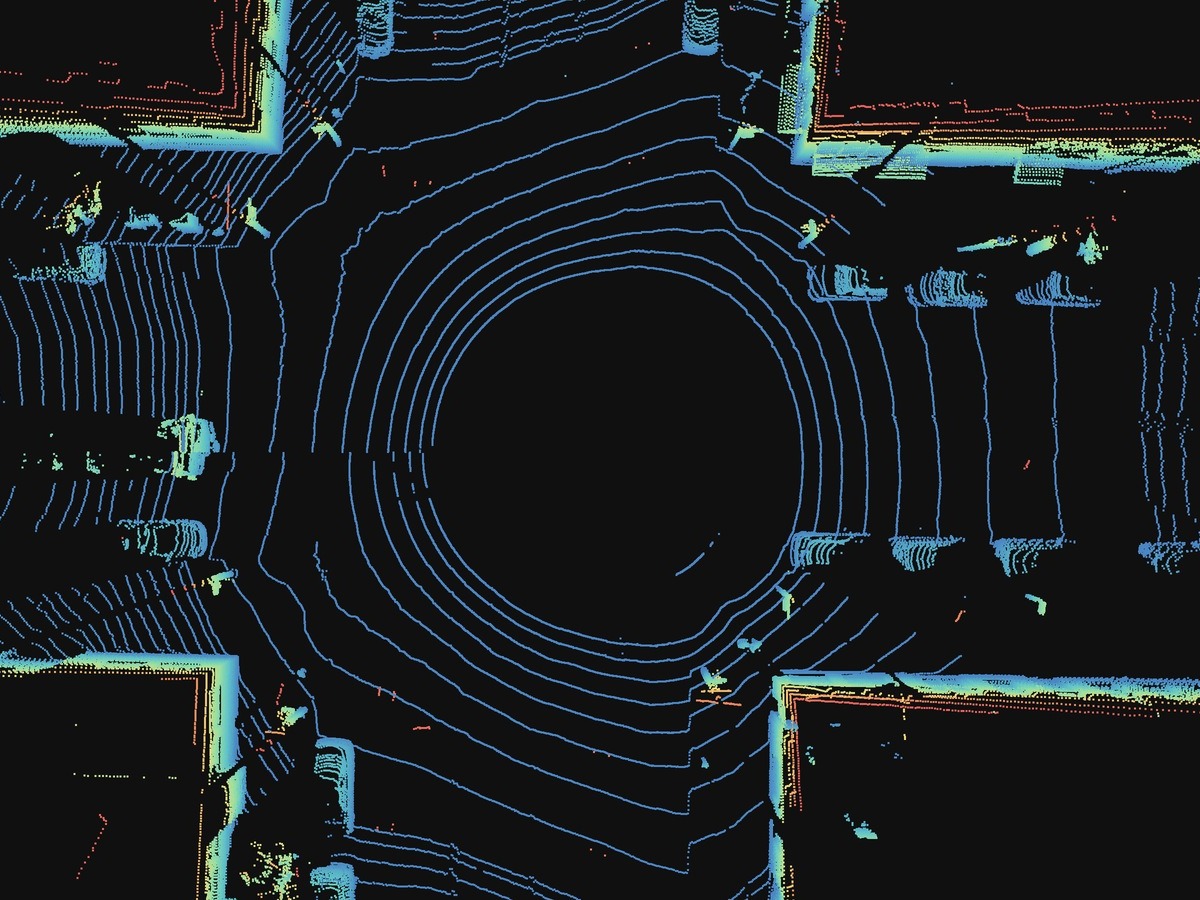} &
    \includegraphics[width=0.40\textwidth]{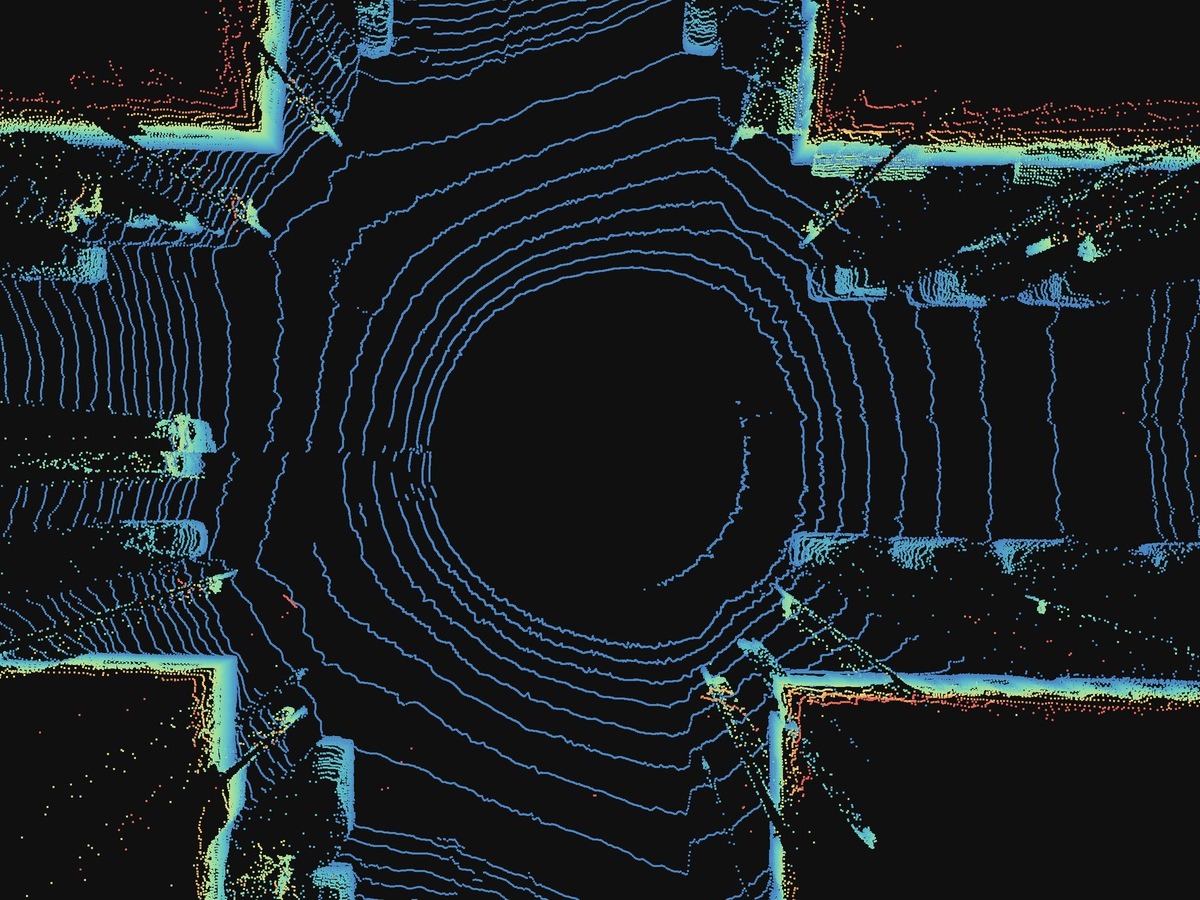} \\[3pt]
    \multicolumn{2}{c}{(c) Range image, front half of the sweep} \\[1pt]
    \includegraphics[width=0.49\textwidth]{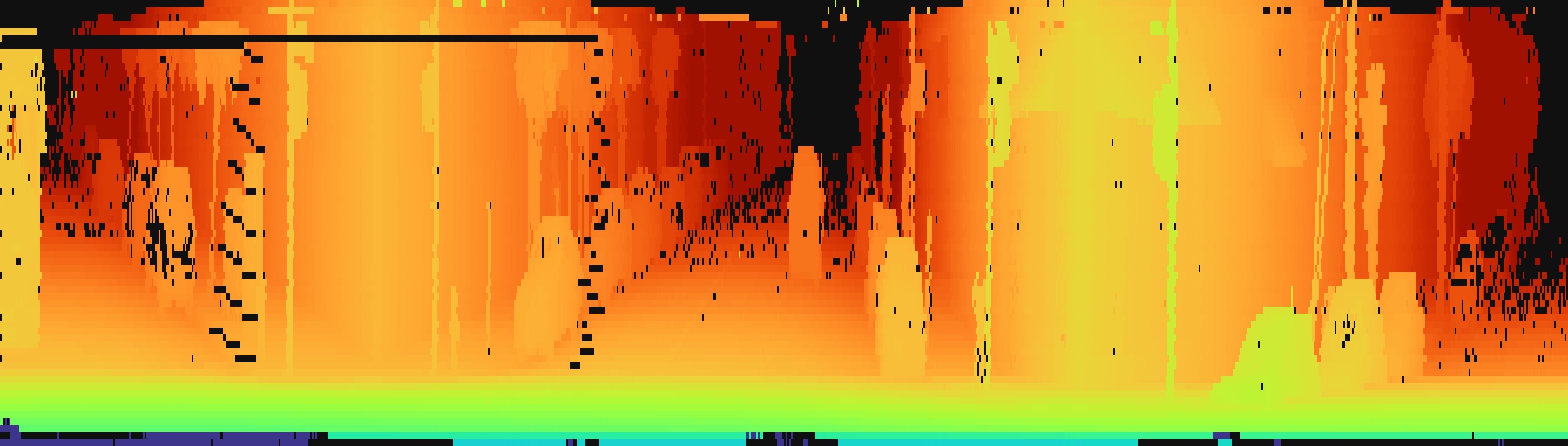} &
    \includegraphics[width=0.49\textwidth]{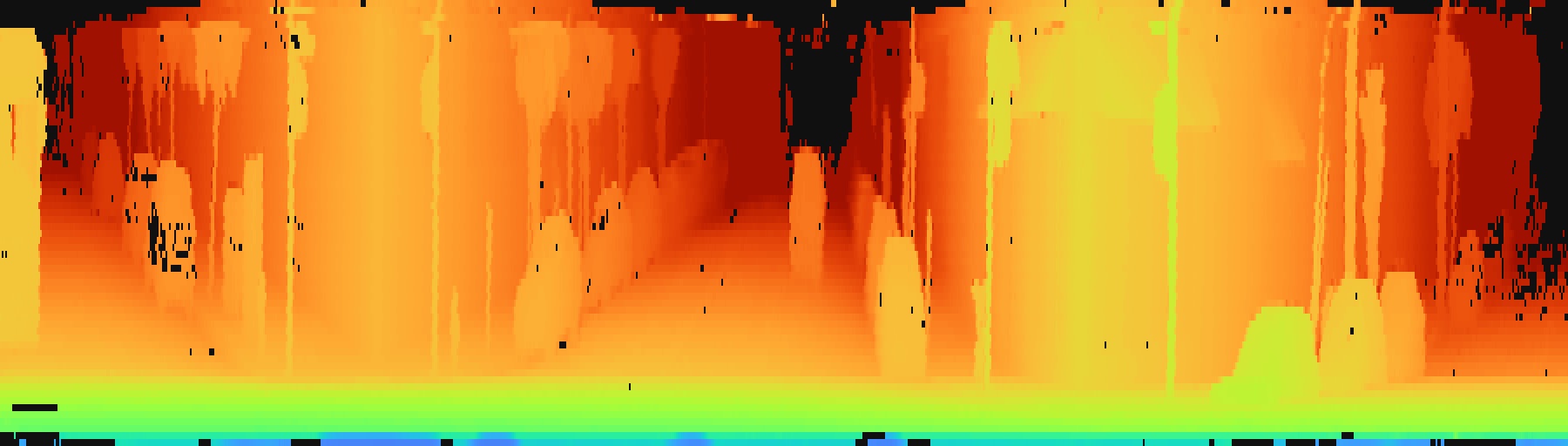} \\[3pt]
    \multicolumn{2}{c}{(d) Intensity image, front half of the sweep} \\[1pt]
    \includegraphics[width=0.49\textwidth]{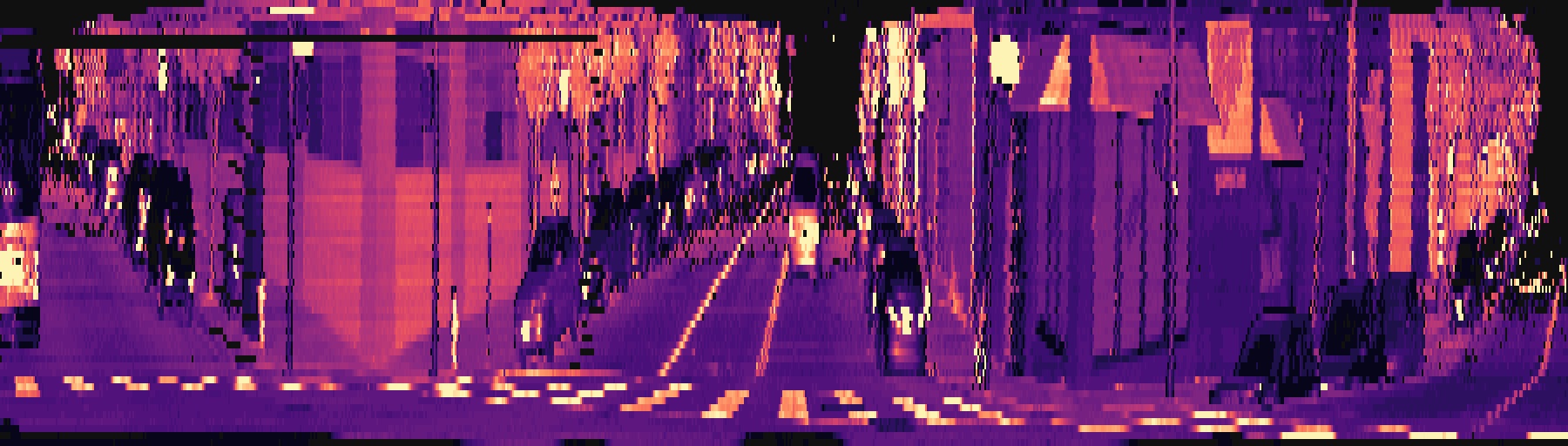} &
    \includegraphics[width=0.49\textwidth]{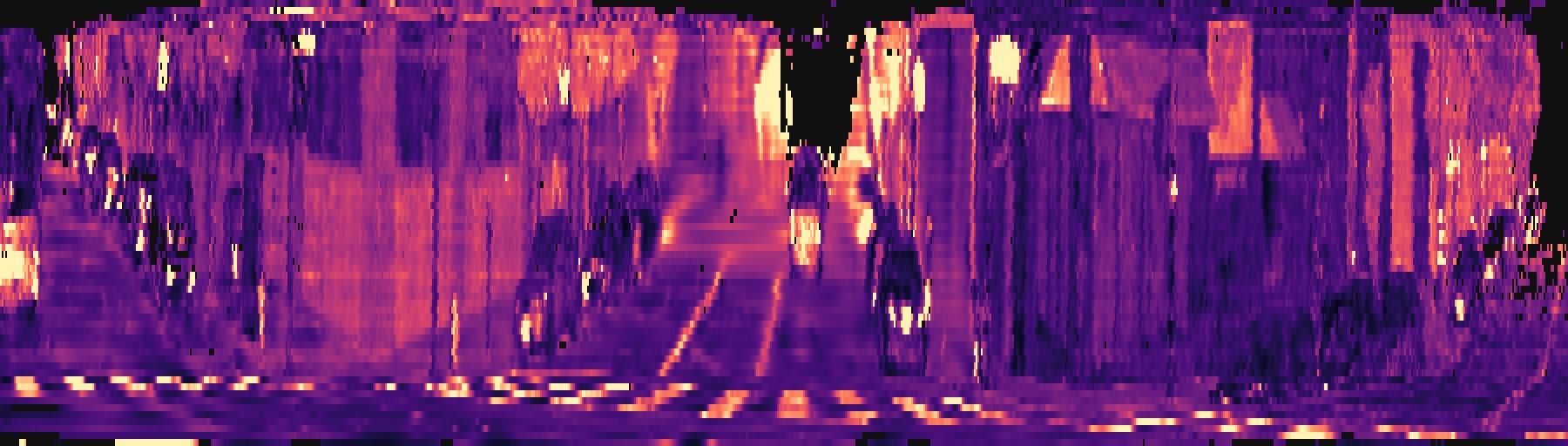}
  \end{tabular}
  \caption{\textbf{A held-out PandaSet sweep and the AETHER rendering of it.} (a, b) Points between \SI{7}{m} and \SI{32}{m} from the sensor, colored by height. (c) Range on a log scale from \SI{2}{m} (blue) to \SI{60}{m} (red). (d) Intensity. Black marks rays without a return; image rows are stretched four times.}
  \label{fig:supp_lidar_vis}
\end{figure*}

\begin{figure*}[p]
  \centering
  \small
  \includegraphics[width=\textwidth]{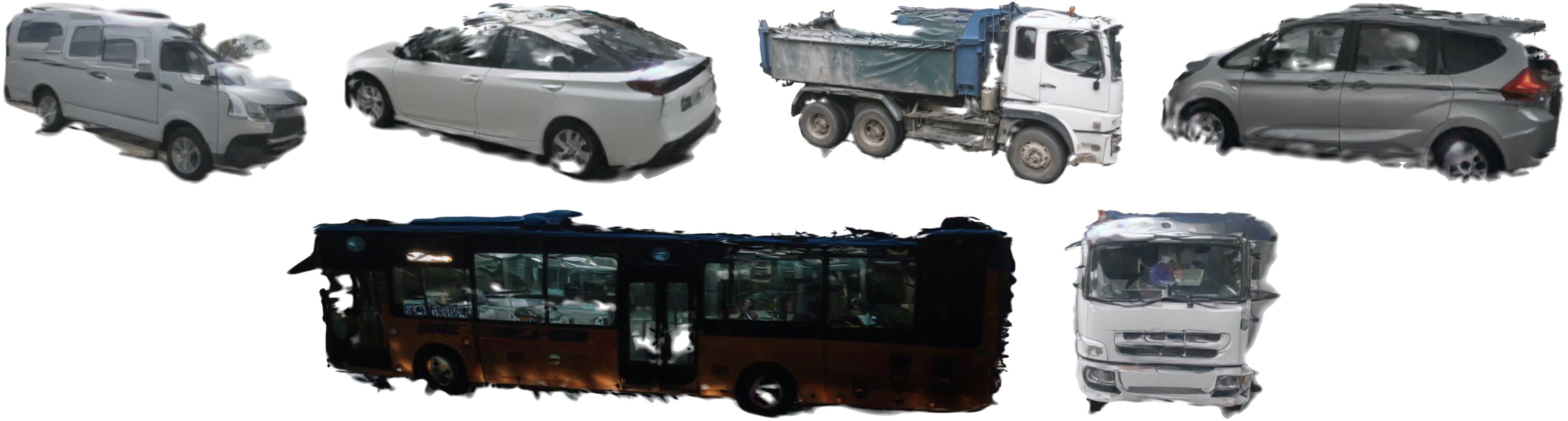}\\[2pt]
  (a) Vehicles reconstructed from the logs, rendered alone\\[8pt]
  \includegraphics[width=\textwidth]{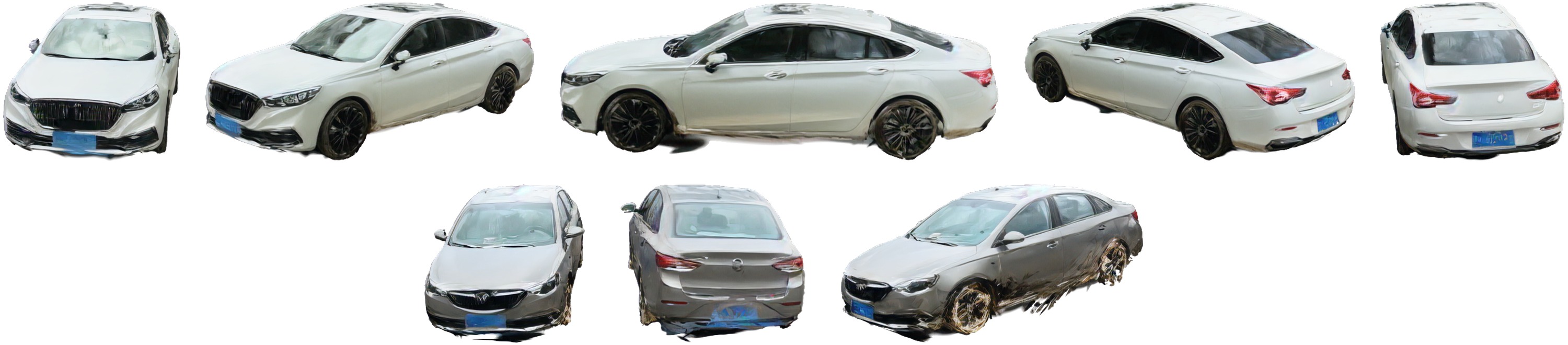}\\[2pt]
  (b) Inserted 3DRealCar~\cite{du20243drealcar} vehicles, seen from several directions
  \caption{\textbf{Vehicles in AETHER}, rendered at $1600{\times}900$ and restored. Logged vehicles are seen from within \ang{35} of the direction from which the ego observed them.}
  \label{fig:supp_vehicles}
\end{figure*}

\begin{figure*}[p]
  \centering
  \small
  \includegraphics[width=\textwidth]{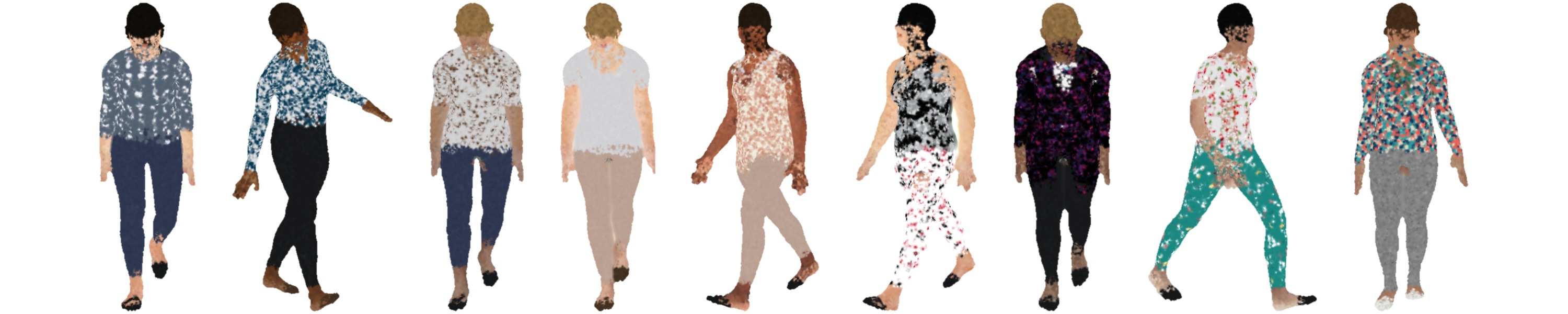}\\[2pt]
  (a) Pedestrian avatars from BEDLAM~\cite{black2023bedlam}, posed by our motion model\\[8pt]
  \includegraphics[width=\textwidth]{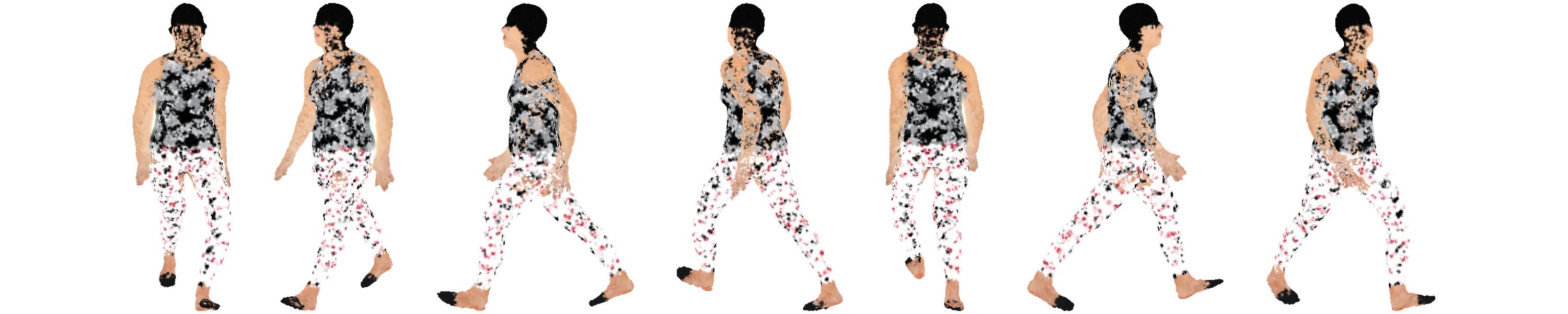}\\[2pt]
  (b) One walking pedestrian seen from seven directions\\[8pt]
  \includegraphics[width=\textwidth]{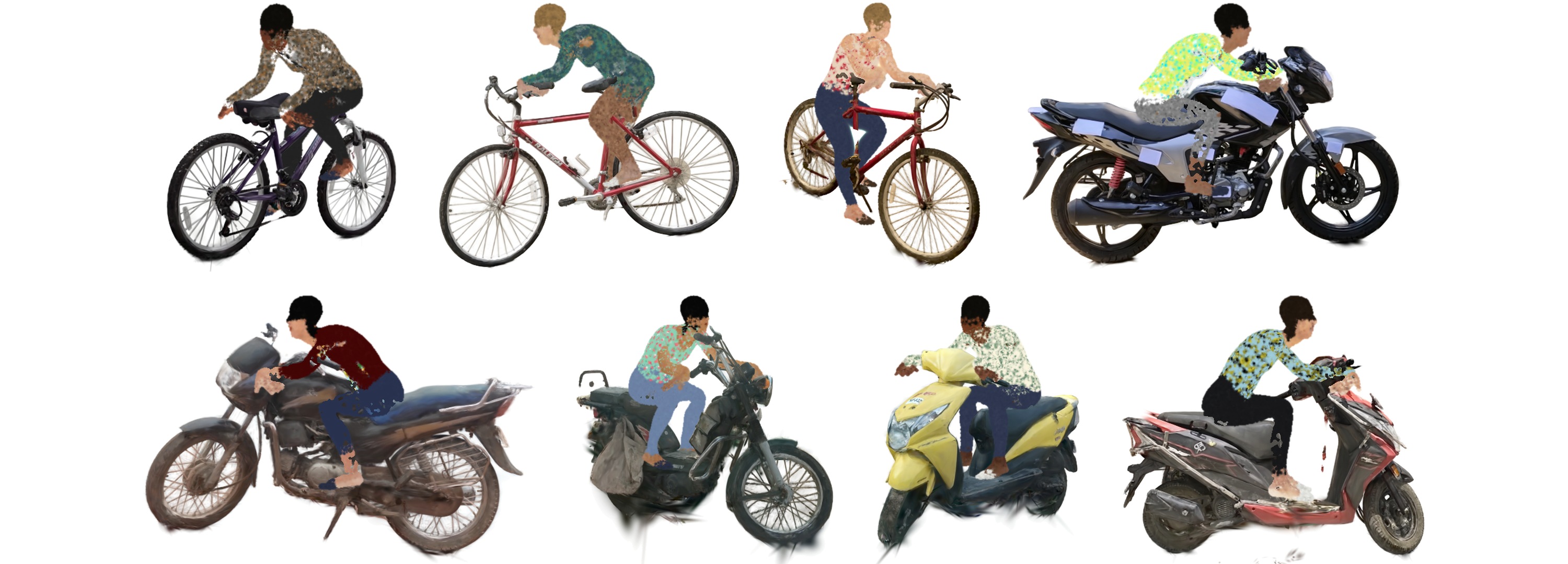}\\[2pt]
  (c) Riders on bicycles, motorcycles and scooters from uCO3D~\cite{liu2025uco3d}, posed from the bike rig\\[8pt]
  \includegraphics[width=\textwidth]{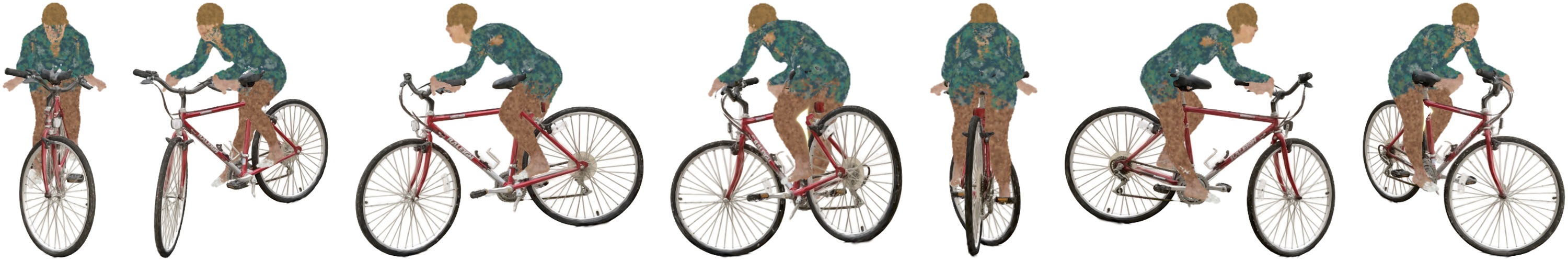}\\[2pt]
  (d) One cyclist seen from seven directions
  \caption{\textbf{Pedestrians and riders inserted by AETHER}, rendered alone at $1600{\times}900$.}
  \label{fig:supp_vru}
\end{figure*}

\Cref{fig:supp_surround} shows all six cameras for a daytime scene, on the logged path and \SI{2}{m} off it, and for a night scene.
\Cref{fig:supp_lidar_vis} compares a held-out LiDAR sweep with the AETHER rendering of it as point clouds and as range and intensity images.
\Cref{fig:supp_vehicles} shows reconstructed and inserted vehicles, and \cref{fig:supp_vru} the pedestrians and riders that AETHER inserts, each seen from several directions.

\section{Limitations}
\label{sec:supp_limits}
The fidelity flag measures the retention of the planner's own detection relative to rendering on the log.
It does not certify real-world fidelity, does not cover other perception outputs such as maps or motion forecasts, and is uncalibrated for planners without a readable detection head.
Reactive traffic follows the log until it must yield and adds IDM vehicles only where traffic thins out; it does not change lanes, and its realism against real driving is not evaluated.
Pedestrian behavior is fitted on nuScenes, conditions on eventual crossing, and treats yielding as a prior; riders rely on priors throughout.
The ground is accurate along the driven corridor but less so away from it, and on these short clips it does not beat a single plane in camera position.
Finally, LiDAR intensity and ray drop remain slightly behind SplatAD.

\end{document}